\documentclass[msom,nonblindrev]{informs3arxiv}

\OneAndAHalfSpacedXI

\usepackage{natbib}
 \bibpunct[, ]{(}{)}{,}{a}{}{,}%
 \def\bibfont{\small}%

\usepackage{booktabs}
\usepackage{multicol}
\usepackage{algorithm}
\usepackage{algorithmic}
\usepackage[skip=2pt,font=normal]{subcaption}
\usepackage{xcolor}
\colorlet{blue}{black}                                   %
\newcommand{\rev}[1]{{\color{black}#1}}                 %
\newenvironment{revblock}{\color{black}}{}
\usepackage{url}

\usepackage{breakurl}
\usepackage[breaklinks]{hyperref}
\usepackage{multibib}
\usepackage{threeparttable}
\usepackage{tabularx}                                   %
\usepackage{placeins}
\usepackage{longtable}

\newcites{ec}{Electronic Companion References}

\TheoremsNumberedThrough     %
\ECRepeatTheorems

\EquationsNumberedThrough    %

\MANUSCRIPTNO{}

\begin{document}

\RUNTITLE{Algorithmic Insurance}

\TITLE{Algorithmic Insurance}

\ARTICLEAUTHORS{%
\AUTHOR{Dimitris Bertsimas}
\AFF{Sloan School of Management, Massachusetts Institute of Technology, MA, USA \EMAIL{dbertsim@mit.edu}} %
\AUTHOR{Agni Orfanoudaki}
\AFF{Saïd Business School, Oxford University, United Kingdom, \EMAIL{agni.orfanoudaki@sbs.ox.ac.uk}}
} %

\ABSTRACT{%
\textbf{Problem definition:} When artificial intelligence (AI) systems make errors in high-stakes domains like medical diagnosis or autonomous vehicles, a single algorithmic flaw across varying operational contexts can generate highly heterogeneous losses that challenge traditional insurance assumptions. Algorithmic insurance constitutes a novel form of financial coverage for AI-induced damages, representing an emerging market that addresses algorithm-driven liability. However, insurers currently struggle to price these risks, while AI developers lack rigorous frameworks connecting system design with financial liability exposure. 
\textbf{Methodology/results:} We analyze the connection between operational choices of binary classification performance—specifically, how practitioners set decision thresholds for their models—to tail risk exposure. Using conditional value-at-risk (CVaR) to capture extreme losses, we prove that established approaches like maximizing accuracy can significantly increase worst-case losses compared to tail risk optimization, with penalties growing \rev{at least} quadratically as thresholds deviate from optimal. We propose a liability insurance contract structure that mandates risk-aware classification thresholds and characterize the conditions under which it creates value for AI providers. Our analysis extends to degrading model performance and human oversight scenarios. \rev{We conduct a mammography case study, demonstrating that accuracy-maximizing thresholds carry up to four times the tail risk of CVaR-optimal thresholds. Insurance contracts create 7--8\% gains for well-calibrated firms through risk transfer and regulatory capital relief, and up to 105\% for poorly calibrated firms through mandatory recalibration. Partially aligned firms can lose from a strict mandate, calling for threshold flexibility in contract design.} 
\textbf{Managerial implications:} Unlike traditional insurance that merely transfers risk, algorithmic insurance can function as both a financial instrument and an operational governance mechanism, simultaneously enabling efficient risk transfer while improving AI safety. AI providers should calibrate their models using tail risk metrics rather than accuracy or expected loss when errors carry asymmetric costs with liability exposure.} 


\maketitle
\vspace{-2em}

\section{Introduction}\label{sec:introduction}

\rev{Failure to diagnose breast cancer generates malpractice payments averaging close to \$1 million per litigated case, with delayed diagnosis representing the most common allegation in breast-related litigation \citep{lee2020breast}.} With multiple artificial intelligence (AI) systems now \rev{cleared by the FDA} for mammography screening and their deployment accelerating, organizations face new liability challenges stemming from data-driven models \citep{stern2022ai}. While AI systems can match or exceed human radiologists in average accuracy, they are typically calibrated to maximize overall performance. \rev{Yet a missed cancer diagnosis can generate a liability payment of that size, while a false positive costs about \$1{,}200 in follow-up care, in 2025 dollars \citep{ong2015}.} This fundamental mismatch between how we optimize AI systems and how liability accumulates extends across all high-stakes applications \citep{amodei2016concrete}, from autonomous vehicles to algorithmic trading, where standardized decision rules can generate catastrophic losses in rare but critical scenarios \citep{danielsson2022artificial}. The absence of risk management frameworks tailored to AI failure modes often forces organizations to choose between forgoing the demonstrated benefits of data-driven systems or accepting unquantified risk exposure \citep{dai2021artificial}.

\textit{Algorithmic insurance} emerges as a market-based solution to this liability challenge, complementing regulatory oversight by creating financial incentives for safer AI deployment \citep{pld2024, ukava2024}. These products provide coverage for damages arising from AI errors, protecting ML providers and users when algorithms make costly mistakes, such as missed diagnoses or false fraud alerts \citep{kumar2020case}. As organizations integrate machine learning (ML) models into critical business processes, they might face multiple forms of risk exposure: liability from erroneous decisions, operational losses from system failures, intellectual property disputes over training data, regulatory penalties for biased outcomes, and reputational damage \citep{NIST2023}. Algorithmic insurance can address these varied risks through specialized financial coverage that accounts for the unique characteristics of AI systems \citep{lior2021insuring}. 

The emergence of algorithmic insurance as a distinct market segment reflects the insurance industry's response to the unique risk profile of AI-driven decision systems. Major insurers have begun developing specialized products to address AI risk, with Munich Re launching in 2018 the first major offering, followed by dedicated AI insurance providers including Armilla AI, Vouch, and Relm Insurance \citep{wtwco}. Industry projections suggest this nascent market could generate approximately \rev{\$4.7 billion} in annual global premiums by 2032, with a compound annual growth rate of around 80\% \citep{Deloitte_2024_AI_Insurance}. 

This rapid market development occurs despite fundamental uncertainties. Industry research indicates that both AI developers and insurers lack the analytical tools to accurately assess algorithmic liability exposure, while financial institutions are still developing appropriate underwriting frameworks in the absence of historical loss data \citep{Zurich2023}. Meanwhile, ML providers typically optimize systems using standard performance metrics without frameworks to translate these operational choices into liability exposure \citep{bastani2020online}. This disconnect between operational calibration and liability quantification creates inefficient risk allocation. Providers deploy systems without analyzing their tail risk exposure, while insurers price coverage without established models linking algorithmic parameters to loss distributions \citep{SwissRe2024}.

We analyze the design and pricing of liability insurance for binary classification systems, focusing on the risk management challenges that arise when algorithms make high-stakes accept or reject decisions in domains such as medical diagnosis, loan underwriting, and fraud prevention. Unlike operational risks from system downtime or cybersecurity breaches, algorithmic liability arises from the core decision-making function itself, specifically when AI classifications directly cause financial or physical harm to affected parties. Such decision-based liability presents distinct insurability challenges that require new frameworks beyond traditional professional liability models. Algorithmic failures challenge established actuarial models \citep{mcneil2015quantitative}, since a fixed decision threshold generates predictable but heterogeneous losses across different contexts. Contrary to human decisions, classification models possess a controllable decision boundary that directly determines error rates. These unique features enable algorithmic insurance to act simultaneously as a financial risk transfer vehicle and enforce ML calibration decisions that reduce the underlying model hazard.

\subsection{Organization and Summary of Contributions}\label{sec:contributions}

We develop a framework that links operational ML decisions, specifically threshold calibration, to financial liability exposure and insurance contract design. Our model captures the fundamental trade-off in binary classification: adjusting the decision threshold to reduce false negatives necessarily increases false positives, with each error type incurring distinct financial consequences. By formulating this as a tail risk minimization problem using conditional value-at-risk (CVaR) \citep{rockafellar2002conditional}, we derive closed-form conditions for optimal threshold selection and establish how insurance contracts create value by transferring risk, reducing capital requirements, and mandating risk-optimal thresholds. Our theoretical results apply to any binary classification system \rev{with a calibration map satisfying the conditions of Remark~\ref{rem:scope}}, where false positive and false negative errors generate different losses across scenarios. We \rev{illustrate} the framework through a case study of AI-assisted mammography. Our contributions are summarized as follows:

\begin{itemize}
    \item We formulate threshold selection as an optimization problem that minimizes the risk measured by the CVaR of economic losses from classification errors. Using a non-linear model that captures how false positive and false negative rates vary with the decision threshold, \rev{we derive closed-form characterizations of risk-optimal thresholds as a function of model quality, class imbalance, and cost ratios. We also give an algorithm that computes them in near-linear time and a continuity guarantee in the cost parameters.} By comparing these risk-aware thresholds to accuracy maximization and expected loss minimization, we establish that using these standard approaches instead of CVaR optimization incurs a tail risk penalty that grows \rev{at least} quadratically with threshold deviation. 
    \item We analyze the design of insurance contracts with per-occurrence and aggregate limits, similar to professional liability coverage, that mandate risk-optimal thresholds for binary classification models. We prove that insurance value decomposes into three components: risk transfer (the difference between expected claim payments and risk-based premiums), capital efficiency (the reduction in required regulatory capital for insured versus uninsured firms), and operational improvement (the CVaR reduction from switching suboptimal thresholds to risk-optimal ones). We characterize when insurance creates positive value and discuss the trade-offs in coverage limit selection. Thus, we demonstrate that algorithmic insurance fundamentally differs from traditional coverage, enabling risk reduction alongside risk transfer.
    \item We extend our framework to cases where ML model performance degrades over time and where firms can invest in human oversight to reduce error rates. \rev{We characterize optimal contract duration under model drift as the point where the renewal cost equals the excess risk that renewing avoids, and show that re-optimizing the threshold on a degraded model cannot restore its tail risk, which only retraining achieves.} Moreover, we analyze how interpretability investments reduce insurance premiums through improved human supervision, deriving the first-order conditions for optimal transparency levels. Our results demonstrate that algorithmic liability is endogenous to ML operational choices, fundamentally linking insurance design with operational decisions at deployment.
    \item We illustrate how our framework translates into practice through a case study in AI-assisted mammography across screening and diagnostic settings. Using realistic cost structures and claim emergence patterns, we demonstrate how to translate model characteristics into liability exposure and insurance premiums. \rev{Our results suggest that accuracy-maximizing firms face 77\% higher CVaR in screening and 297\% higher CVaR in diagnostic settings compared to risk-aware calibration. Insurance value decomposition shows that well-calibrated firms gain 7--8\% of the tail risk under the mandated threshold through base insurance value, and that poorly calibrated firms capture up to 105\% in the diagnostic setting through capital relief and mandatory recalibration. Partially aligned firms can lose from a strict mandate. With inputs grounded in documented sources and results tested over their ranges, we establish how theoretical risk measures translate into implementable insurance contracts.}
\end{itemize}

The remainder of the paper is organized as follows. Section~\ref{sec:study_setting} reviews the related literature and establishes our institutional setting and modeling assumptions. Section~\ref{sec:baselineriskmanagement} introduces our baseline model formulation and uses it to derive optimal classification thresholds under tail risk minimization. In Section \ref{sec:theorybenchmarks}, we compare CVaR-optimal thresholds to accuracy and expected loss benchmarks, quantifying the resulting tail risk penalties. Section~\ref{sec:insurance_value} formalizes the proposed insurance contract design and characterizes the conditions for positive value creation. Section~\ref{sec:extensions} extends our model to performance degradation and investments in human oversight. Section~\ref{sec:casestudy} \rev{illustrates} our findings with empirical evidence from a case study in AI-assisted mammography. We conclude in Section~\ref{sec:discussion}.

\section{Problem Setting}\label{sec:study_setting}

This section provides the background upon which we ground our analysis. We first review relevant literature on AI risk management and insurance economics (Section \ref{sec:litreview}). Subsequently, we describe the market structure linking ML providers, clients, and insurers (Section \ref{sec:marketstructure}) and formalize the dual-limit contract  and key modeling assumptions of the proposed insurance design (Section \ref{sec:contractstructure}). 

\subsection{Literature Review}\label{sec:litreview}

The algorithm risk management literature has primarily focused on adoption barriers, particularly algorithm aversion to automated decision-making despite superior performance \citep{dietvorst2018overcoming}. While \citet{burton2020systematic} and \citet{geistfeld2017roadmap} identify mistrust and accountability fears as key operational challenges in high-stakes domains, such as healthcare and autonomous driving, they largely treat algorithmic systems as fixed entities rather than examining how their operational parameters affect risk exposure. Recent work on human-AI collaboration examines how to optimally combine human judgment with algorithmic predictions \citep{ibrahim2021eliciting,grand2024best,krakowski2025human}. While these papers optimize for operational efficiency and effectiveness, they do not consider how human-AI configurations affect liability exposure.

A related stream of operations research has examined operational parameters of ML model deployment, recognizing that ex-post decisions critically impact system performance. \citet{kallus2022assessing} demonstrate how classification thresholds directly affect operational metrics and allocation decisions, showing that ML calibration extends beyond accuracy to affect fairness and resource distribution. In healthcare operations, \citet{feizi2023vertical} demonstrate how threshold-based patient streaming policies in emergency departments can lead to increased efficiency and effectiveness. \citet{ban2019big} show that cost-asymmetric operational decisions require threshold calibration of ML predictions, though the outlined newsvendor framework addresses expected profitability rather than the tail risk management central to liability contexts. Existing work does not address how threshold calibration affects tail risk when algorithmic errors lead to liability exposure.

This liability dimension becomes critical when we consider how insurance economics addresses emerging technologies. Traditional insurance models reveal fundamental inadequacies when applied to technological innovations \citep{eling2016we}. Conventional actuarial approaches rely on historical claims data and assume predictable loss distributions, yet fail to capture how a single algorithmic calibration choice creates heterogeneous risk exposure across different contexts \citep{embrechts2002correlation}. While researchers recognize the need for new risk assessment methodologies \citep{shavell2020redesign}, past studies have not developed operational frameworks that quantify how specific ML calibration decisions translate into tail risk exposure and insurance value.

\rev{The regulatory environment is responding to algorithmic harm, though liability for AI errors still rests mainly on national tort and product law. The revised EU Product Liability Directive extends no-fault liability to software and AI systems \citep{pld2024}. The UK Automated Vehicles Act 2024 assigns responsibility for the driving task to an authorized self-driving entity, while compensation is still paid through the vehicle's insurer \citep{ukava2024}.} \citet{vcerka2015liability} suggest market-based instruments like specialized insurance could help manage this emerging class of liability, but provide limited guidance on how such instruments should be designed.

The design of liability insurance products is determined by extreme loss events rather than expected values, requiring specialized tail risk management frameworks \citep{mcneil2015quantitative}. CVaR, capturing the expected losses exceeding a specified quantile of the loss distribution, provides a coherent risk measure particularly suited for catastrophic outcomes \citep{rockafellar2002conditional}.  While CVaR has been extensively applied in finance and supply chain management \citep{krokhmal2002portfolio,choi2011multiproduct}, its application to algorithmic risk remains an open question.

Our contribution bridges these streams of research by developing an integrated framework that connects ML operational decisions to tail risk exposure and designs insurance contracts that incentivize risk-aware deployment. Unlike prior work that optimizes for average performance or treats algorithmic risk as exogenous, we show how operational choices fundamentally determine liability exposure and how insurance can serve as both a risk transfer mechanism and a governance tool. This integration of operational design with financial risk management provides a foundation for the emerging algorithmic insurance market.

\subsection{Market Structure}\label{sec:marketstructure}

\begin{figure}
    \centering
\includegraphics{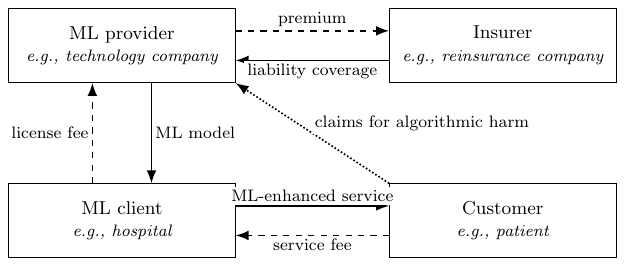}
    \caption{The Algorithmic Insurance Business Model. \rev{Solid: model and service; dashed: payments; dotted: claims.}}
    \label{fig:alginsurancebusinessmodel}
\end{figure}

In Figure \ref{fig:alginsurancebusinessmodel}, we illustrate the proposed business model and provide an example in the context of healthcare. The ML provider (e.g., a technology company) receives a license fee from its client (e.g., a hospital) in exchange for the deployment of an ML model. Respectively, the ML client (e.g., a hospital) provides a service to its customers (e.g., a patient) that is enhanced by the ML model and receives, in return, a service fee. When algorithmic errors cause harm, affected parties may pursue claims against the ML provider \rev{ under product liability and tort law \citep{pld2024}}. The algorithmic insurance contract is offered by an insurance or a reinsurance company in exchange for a premium to cover the liability exposure of the ML \rev{provider} to \rev{the client's} customers.

Unlike traditional insurance that relies on historical claims from heterogeneous decision-makers, algorithmic insurance can leverage the specific characteristics of the deployed model and its operating environment. Insurers can verify and mandate specific operational parameters such as classification thresholds as coverage conditions, leveraging the programmable nature of algorithmic systems. The standardized deployment model, where fixed decision rules apply across diverse implementation contexts, creates concentration in tail risk exposure. This configuration aligns risk management incentives with technical capabilities. ML providers control model architecture and calibration decisions, while insurers possess actuarial expertise to price tail risks. The resulting market structure enables both risk transfer and operational improvement through contractually mandated risk management practices.

\subsection{Contract Scope and Assumptions}\label{sec:contractstructure}

We consider settings where organizations deploy binary classification systems in high-stakes environments with asymmetric costs. False positive and false negative predictions generate different liability exposures with high heterogeneity across deployment settings. The premium is determined as a function of the assumed risk, varying according to both the likelihood of occurrence (frequency) and the magnitude of consequences (severity). 

During a single contract period, we assume the underlying data distribution and cost structures remain stable. Model performance may degrade predictably (addressed in Section~\ref{sec:drift}), but the fundamental decision environment does not shift discontinuously. Contract renewal points provide opportunities to recalibrate for environmental changes. Moreover, we assume access to scenario-based cost estimates. While historical algorithmic claims may be limited, domain expertise (e.g., medical malpractice patterns) provides guidance on expected levels of claim severity. The framework explicitly models heterogeneous costs across scenarios, departing from traditional actuarial approaches that rely on historical claim averages. \rev{We further assume that the insurer can observe and enforce the mandated threshold and that the severity distributions of the scenarios are taken as given. Section~\ref{sec:limitations} discusses the consequences of relaxing both.} \rev{Finally, we price coverage with the cost-of-capital method of Solvency II and standard actuarial practice \citep{solvency2015, mohr2011}. The premium is a loading on expected payouts plus a charge, at the insurer's cost of capital, on the tail risk of the insured payments beyond their expectation.} This approach aligns \rev{the incentives of the insurer and the ML provider} for tail risk management \rev{and limits the scope for} adverse selection by high-risk operators.

\section{Risk-Aware Optimal Threshold Design for ML Systems}\label{sec:baselineriskmanagement}

In this section, we develop the theoretical foundation of our framework. We focus on the classification threshold $\tau$, the primary operational decision that determines which instances are classified as positive versus negative, and characterize its optimal value under tail risk from liability exposure. Section \ref{sec:trigonomodel} constructs a stylized non-linear model of binary classification performance, deriving closed-form expressions for how false positive and false negative rates depend on the threshold choice. Section \ref{sec:cvaropt} formulates and solves the threshold optimization problem under the CVaR measure, establishing conditions for optimality and characterizing the solution structure. Finally, Section \ref{sec:sensivititybaseline} demonstrates how optimal thresholds vary with cost asymmetry and risk aversion. 

\subsection{A Stylized Non-Linear Model for Binary Classification Performance}\label{sec:trigonomodel}

The threshold optimization problem requires explicit characterization of how false positive and false negative rates depend on the threshold choice. We consider binary classification where class 0 represents the negative outcome (e.g., healthy patient, legitimate transaction) and class 1 represents the positive outcome requiring action (e.g., sick patient, fraudulent transaction). We develop a parametric family of models where error rates can be expressed analytically as functions of the threshold, permitting closed-form characterization of the risk-optimal threshold. Our model employs two parameters: $\alpha \in (0,1)$, representing the proportion of class 0 instances in the population, and $k \in (0,\pi/2]$, which governs the classifier's discriminative power. The parameters $\alpha$ and $k$ allow us to analyze the threshold selection problem across varying class imbalances and model discrimination levels.

Let $\overline{\gamma} \in [0,1]$ denote the ML model's score for an instance, which approximates the unobserved true likelihood $\gamma$ of belonging to class 1. Classification proceeds by applying a decision threshold $\tau \in [0,1]$: an instance is assigned to class 1 if $\overline{\gamma} \geq \tau$, and to class 0 otherwise. We define $\mathbb{P}(Y = 1 \mid \overline{\gamma})$ as the conditional probability that an instance with score $\overline{\gamma}$ truly belongs to class 1. We assume that the marginal distribution of the model score $\overline{\gamma}$ is uniform over $[0,1]$, \rev{as for a rank-transformed score}. Table~\ref{tab:notation} \rev{in Section~\ref{tab:notationsummary}} provides a summary of the notation introduced across the manuscript.

To enable analytical tractability while capturing essential features of classification problems, we employ a trigonometric functional form for the conditional probability $\mathbb{P}(Y=1|\overline{\gamma})$: 
\begin{equation}\label{eq:trig-model}
\mathbb{P}(Y=1 \mid \overline{\gamma}) =
\begin{cases}
\frac{1-\alpha}{\sin k} \left[\sin\left(\frac{k}{\alpha} \cdot \overline{\gamma} - k \right) + \sin k \right], & \text{if } \overline{\gamma} \leq \alpha \\[0.5em]
(1 - \alpha) + \frac{\alpha}{k} \left[ \arcsin\left( \frac{(\overline{\gamma} - 1)\sin k}{1 - \alpha} \right) + k \right], & \text{if } \overline{\gamma} > \alpha
\end{cases}
\end{equation}

\begin{figure}[htbp]
    \centering
    \begin{subfigure}[t]{0.4\textwidth}
        \centering
        \includegraphics[width=\linewidth]{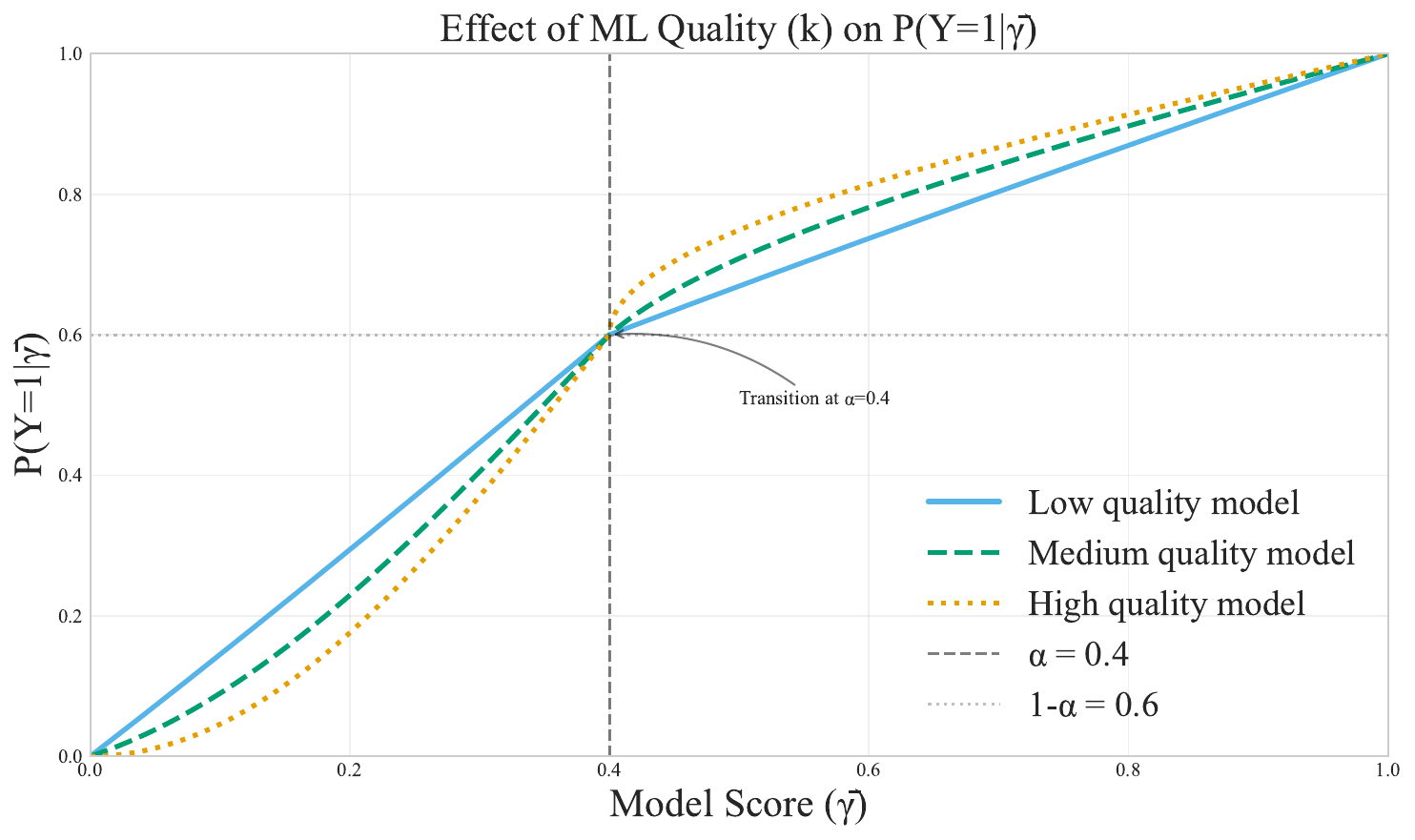}
        \vspace{4pt}
        \caption{Varying model quality ($k$)}
    \end{subfigure}
    \hfill
    \begin{subfigure}[t]{0.4\textwidth}
        \centering
        \includegraphics[width=\linewidth]{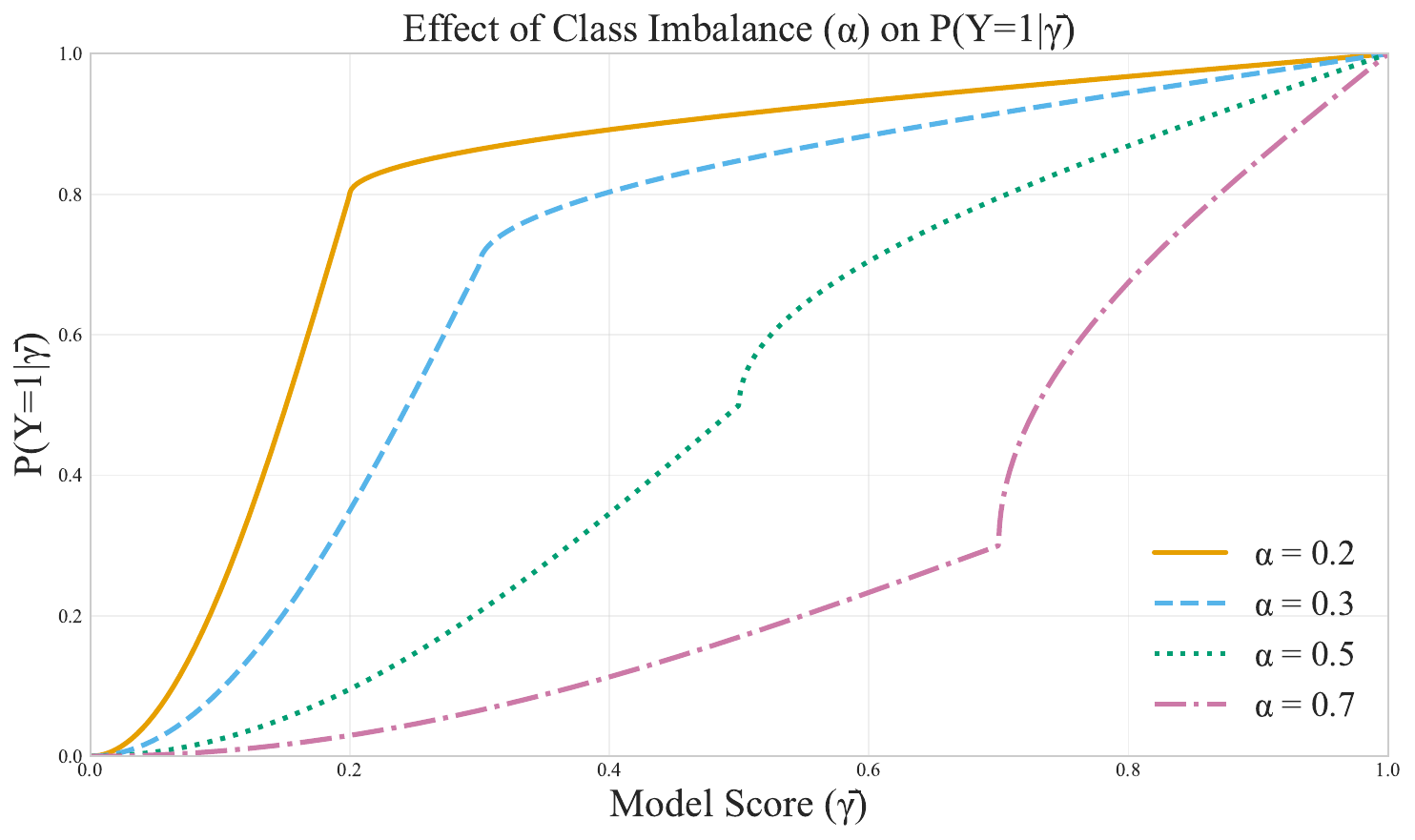}
        \vspace{4pt}
        \caption{Varying class imbalance ($\alpha$)}
    \end{subfigure}
    \caption{$P(Y{=}1 \mid \overline{\gamma})$ under varying model quality (a) and class imbalance (b) in the trigonometric model.}
    \label{fig:trig-side-by-side}
\end{figure}

The choice of trigonometric functions is motivated by their ability to capture the S-shaped calibration curves commonly observed in ML classifiers while maintaining 
analytical tractability. Figure \ref{fig:trig-side-by-side} in Section~\ref{sec:suppfigures} illustrates the $\mathbb{P}(Y=1 \mid \overline{\gamma})$ for varying model configurations of discrimination quality ($k$) and class imbalance ($\alpha$).
This specification is motivated by three considerations: (i) it satisfies the calibration constraints $\mathbb{P}(Y=1|0) = 0$, $\mathbb{P}(Y=1|1) = 1$, and preserves the population proportion $\int_0^1 \mathbb{P}(Y=1|\overline{\gamma})d\overline{\gamma} = 1-\alpha$ (Lemmas \ref{lem:boundary}-\ref{lem:conservation}); (ii) it ensures monotonicity with $\frac{d\mathbb{P}(Y=1|\overline{\gamma})}{d\overline{\gamma}} > 0$, reflecting that higher model scores indicate greater likelihood of class 1 membership (Lemma \ref{lem:monotonicity}), while parameter $k$ controls how sharply this probability transitions from 0 to 1; and (iii) the trigonometric functions admit closed-form integration when computing the false positive and negative rates, enabling analytical optimization (Lemmas \ref{lem:fn_case}--\ref{lem:fp_case}). 

The mapping $\mathbb{P}(Y=1|\overline{\gamma})$ exhibits an S-shaped pattern \rev{common} to classification algorithms, from logistic regression to neural networks, where extreme scores indicate high confidence and intermediate values reflect uncertainty \rev{\citep{niculescu2005predicting}}. Our trigonometric specification captures this empirical regularity. \rev{Moreover, a single parameter $k$ controls model quality, with higher values indicating better discrimination, and it can be mapped from a real-world classifier's AUC. The family spans AUCs from about 0.83 to 0.91, which corresponds to the discrimination reported for AI systems in mammography (Section~\ref{sec:casesetup}).\footnote{\rev{As $k$ approaches $\pi/2$ the calibration map becomes sharpest and the AUC largest (0.89--0.91, depending on $\alpha$); as $k$ approaches 0 the map becomes piecewise linear and the AUC tends to $5/6$. The ROC curve is proper (concave, by Lemma~\ref{lem:monotonicity}), with a vertical tangent at the origin and a discontinuity in curvature at the $\alpha$-quantile, where the slope equals one and the two branches of the map meet.}}} Thus, among single-parameter families that capture classification behavior, the proposed form enables analytical characterization of tail risk through closed-form integration while maintaining sufficient flexibility to represent diverse ML models. This analytical tractability is essential for deriving actionable insights about insurance design that we present in the following sections.

\rev{We validate the functional form in Section~\ref{app:trigonometric_validation} (Tables~\ref{tab:dataset_configs} and \ref{tab:trigonometric_validation}, Figure~\ref{fig:validation_anchors}) by fitting it to out-of-fold, rank-transformed calibration curves of six classifiers on four clinical and 44 synthetic datasets. Within its range of discrimination and at moderate prevalence the fit is close (held-out $R^2$ of 0.97 on the clinical dataset closest to our case study). In that range, setting the threshold with the trigonometric model when the true map is nonparametric raises tail risk by at most 11\% at the median. We also test the robustness of our results to other families of calibration maps, such as the equal-variance binormal model of Section~\ref{sec:binormal}. Analytically, Remark~\ref{rem:scope} shows that every result except the closed forms holds for any calibration map satisfying the conditions stated there. The case study uses both models (Section~\ref{sec:casesetup}).}

\begin{figure}[htbp]
    \centering
    \begin{subfigure}[t]{0.49\textwidth}
        \centering
        \includegraphics[width=\linewidth]{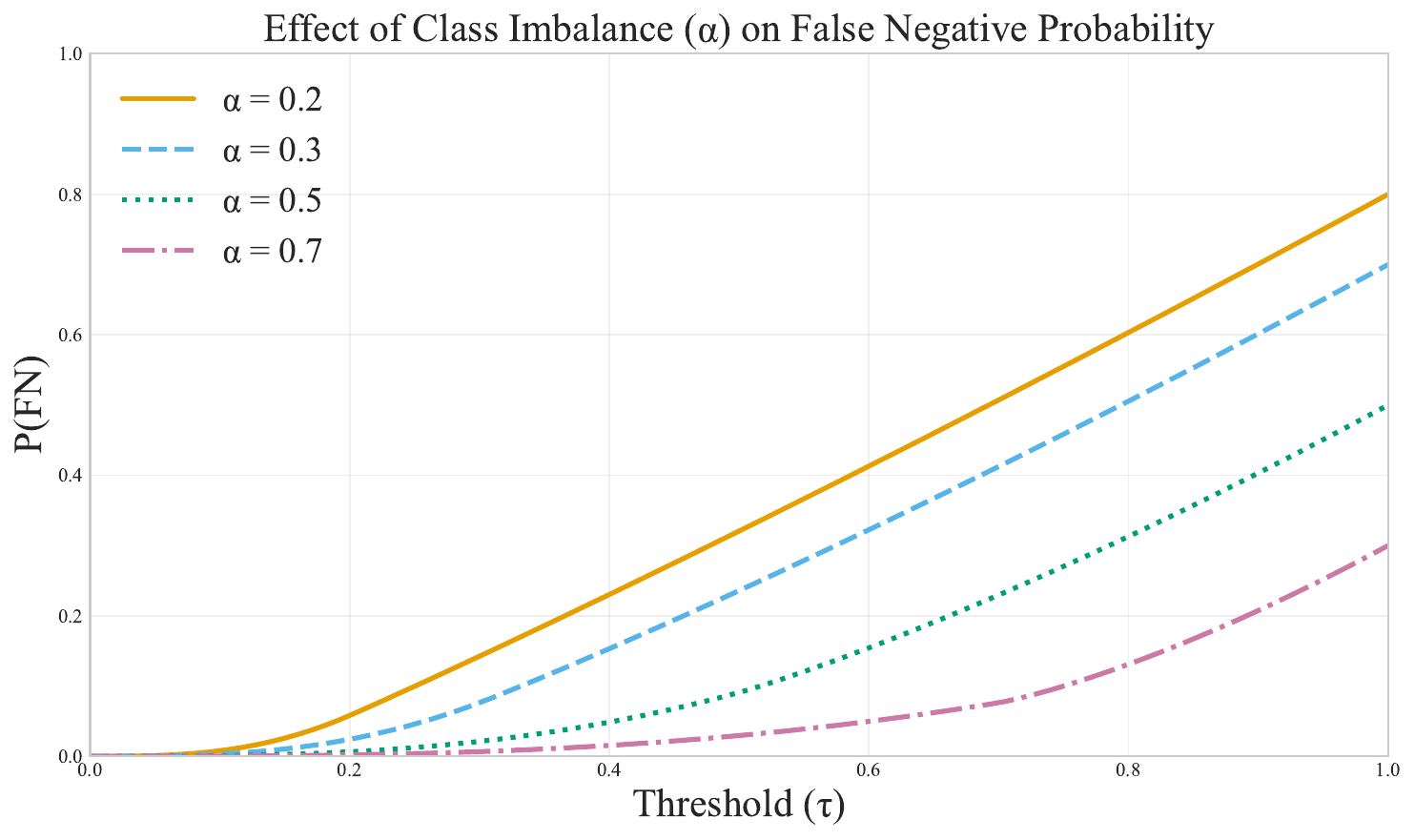}
        \vspace{4pt}
        \caption{$\mathbb{P}_{\mathrm{FN}}(\tau)$}
    \end{subfigure}
    \hfill
    \begin{subfigure}[t]{0.49\textwidth}
        \centering
        \includegraphics[width=\linewidth]{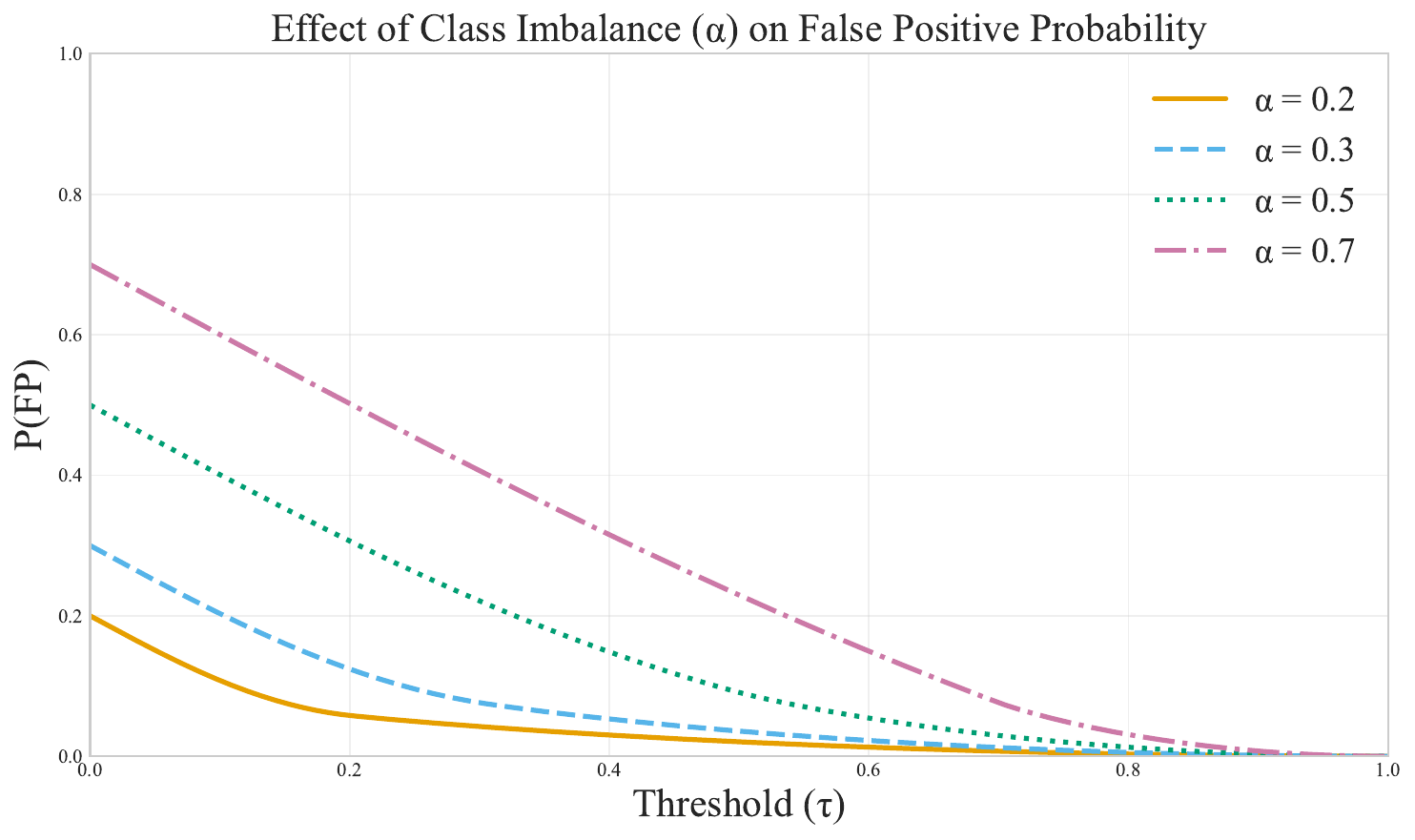}
        \vspace{4pt}
        \caption{$\mathbb{P}_{\mathrm{FP}}(\tau)$}
    \end{subfigure}
    \caption{Illustration of error probabilities $\mathbb{P}_{\mathrm{FN}}(\tau)$ and $\mathbb{P}_{\mathrm{FP}}(\tau)$ under varying class imbalance $\alpha$.}
    \label{fig:fp_fn}
\end{figure}

\rev{Under the trigonometric model}, we derive closed-form expressions for the false positive rate $\mathbb{P}_{\mathrm{FP}}(\tau)$ and false negative rate $\mathbb{P}_{\mathrm{FN}}(\tau)$ as functions of the threshold choice $\tau$ (see Lemmas \ref{lem:fn_case}--\ref{lem:fp_case}). These expressions reveal that error rates vary non-linearly with $\tau$ depending on both model quality $k$ and class proportion $\alpha$ (see Figure \ref{fig:fp_fn} for a visualization). The derived error rates exhibit a fundamental asymmetry: the marginal impact of threshold changes on false positives versus false negatives depends critically on whether $\tau < \alpha$ or $\tau > \alpha$. The final $\mathbb{P}_{\mathrm{FP}}(\tau)$ and $\mathbb{P}_{\mathrm{FN}}(\tau)$ expressions form the basis of our structural analysis, enabling tractable evaluation of risk under different threshold choices and model parameters and allowing us to compare the value of algorithmic insurance against alternative benchmarks. The formal statements and proofs are included in Section \ref{proofstrigonoproperties}.

\subsection{Threshold Optimization under Conditional Value-at-Risk}\label{sec:cvaropt}

We now analyze the optimal classification threshold when the decision maker seeks to minimize the tail risk of downstream losses due to algorithmic errors using the CVaR measure. This risk measure, widely adopted in finance and operations, captures the expected losses conditional on exceeding the $\beta$-quantile, providing a coherent framework for risk-aware ML model calibration \citep{rockafellar2000optimization}.

Formally, for a loss variable $Z$ \rev{with a continuous distribution}, CVaR at confidence level $\beta$ equals $\mathbb{E}[Z | Z \geq \text{VaR}_\beta(Z)]$, where $\text{VaR}_\beta(Z) = F_Z^{-1}(\beta)$ is the $\beta$-quantile. The key insight of \citet{rockafellar2000optimization} reformulates this as:
\begin{equation}\label{eq:cvar_representation}
\text{CVaR}_\beta(Z) = \min_{\nu \in \mathbb{R}} \left\{\nu + \frac{1}{1-\beta}\mathbb{E}[(Z-\nu)^+]\right\},
\end{equation}
where $(z)^+ = \max\{z, 0\}$ and the minimum is attained at $\nu^\ast = \text{VaR}_\beta(Z)$, transforming CVaR minimization into a convex optimization problem, crucial for computational tractability.

In our classification setting, we evaluate $N$ instances across $J$ risk scenarios representing different operational conditions (e.g., peak vs. off-peak periods, varying cost structures, or routine vs. emergency cases,), where instance $i$ in scenario $j$ incurs cost $K_{ji} \rev{> 0}$ for false positive errors and $L_{ji} \rev{> 0}$ for false negative errors. The loss in scenario $j$ becomes:
\begin{equation}\label{eq:loss_scenario}
S_j(\tau) = \sum_{i=1}^{N} \left[\mathbb{P}_{\mathrm{FP}}(\tau) K_{ji} + \mathbb{P}_{\mathrm{FN}}(\tau) L_{ji}\right],
\end{equation}

where $\mathbb{P}_{\mathrm{FP}}(\tau)$ and $\mathbb{P}_{\mathrm{FN}}(\tau)$ are the closed-form error rates derived from our trigonometric model (see Lemmas \ref{lem:fn_case}--\ref{lem:fp_case}\rev{, with the geometry of the integration regions in Figure~\ref{fig:trigonometricregions}}).\footnote{\rev{They are the joint probabilities $\mathbb{P}(\overline{\gamma} \geq \tau, Y = 0)$ and $\mathbb{P}(\overline{\gamma} < \tau, Y = 1)$, so that $N\mathbb{P}_{\mathrm{FP}}(\tau)$ and $N\mathbb{P}_{\mathrm{FN}}(\tau)$ are the expected numbers of errors; we refer to them as rates for brevity.}} The risk-aware threshold optimization problem becomes $\min_{\tau \in [0,1]} \text{CVaR}_\beta(S(\tau))$, which, using the empirical approximation from \citet{rockafellar2002conditional}, yields:
\begin{equation}\label{eq:empirical_cvar}
\min_{\tau \in [0,1], \nu \in \mathbb{R}} \left\{\nu + \frac{1}{(1-\beta)J} \sum_{j=1}^{J} (S_j(\tau) - \nu)^+\right\}
\end{equation}
This formulation captures a fundamental trade-off: lowering $\tau$ reduces false negatives but increases false positives. The CVaR objective ensures we optimize this trade-off not for the average case, but for protecting against worst-case cost realizations, corresponding to the scenarios that carry liability risk in high-stakes algorithmic decisions. The convexity of the CVaR objective (Lemma~\ref{lem:convexity}), combined with our trigonometric model's closed-form error rates, enables us to derive analytical characterizations of the optimal threshold $\tau$. \begin{revblock}
Since the tail scenario set depends on $\tau$, we work with the dual representation of CVaR \citep{rockafellar2002conditional}: with $m := (1-\beta)J$ and $\mathcal{Q}_\beta := \{q \in \mathbb{R}^J : 0 \leq q_j \leq 1/m,\ \sum_j q_j = 1\}$,
\begin{equation}\label{eq:envelope}
\text{CVaR}_\beta(S(\tau)) = \max_{q \in \mathcal{Q}_\beta} \sum_{j=1}^{J} q_j S_j(\tau),
\end{equation}
where the maximizing weights place mass $1/m$ on the scenarios above $\text{VaR}_\beta(S(\tau))$, no mass on those below it, and split the remaining mass among scenarios tied at $\text{VaR}_\beta$. Lemma~\ref{lem:cvar_technical} gives the resulting value of $\text{CVaR}_\beta(S(\tau))$ in closed form, together with its one-sided derivatives. Writing $K_j = \sum_i K_{ji}$ and $L_j = \sum_i L_{ji}$ for the aggregate costs of scenario $j$, the optimal threshold is characterized as follows.

\begin{theorem}[Threshold Characterization]\label{thm:optimal_threshold}
The minimizer $\tau^{\ast}$ of $\text{CVaR}_\beta(S(\tau))$ on $[0,1]$ is unique and lies in $(0,1)$. There exist worst-case weights $q^\ast \in \mathcal{Q}_\beta$ such that $(\tau^{\ast}, q^\ast)$ is a saddle point of $\sum_j q_j S_j(\tau)$ on $[0,1] \times \mathcal{Q}_\beta$, and with the tail-weighted costs $K_{q^\ast} := \sum_j q^\ast_j K_j$ and $L_{q^\ast} := \sum_j q^\ast_j L_j$,
\begin{equation}\label{eq:foc}
\mathbb{P}(Y=1 \mid \tau^{\ast}) = \frac{K_{q^\ast}}{K_{q^\ast} + L_{q^\ast}} =: p^\ast .
\end{equation}
Under the trigonometric model, $\tau^{\ast}$ takes one of two forms:
\begin{enumerate}
\item If $p^\ast \leq 1-\alpha$: $\tau^{\ast} = \alpha + \frac{\alpha}{k}\arcsin\left(\left(\frac{p^\ast}{1-\alpha} - 1\right)\sin k\right) \leq \alpha$;
\item If $p^\ast > 1-\alpha$: $\tau^{\ast} = 1 - (1-\alpha)\frac{\sin\left(\frac{k}{\alpha}(1-p^\ast)\right)}{\sin k} > \alpha$.
\end{enumerate}
\end{theorem}

When $m$ is an integer and the $m$-th and $(m+1)$-th largest losses at $\tau^{\ast}$ differ, $q^\ast$ is uniform on the tail set $\mathcal{J}_\beta$ of the $m$ largest losses at $\tau^{\ast}$ and $p^\ast = K_{\mathcal{J}_\beta}/(K_{\mathcal{J}_\beta}+L_{\mathcal{J}_\beta})$ is the share of false positive costs in the tail. When several scenarios are tied at $\text{VaR}_\beta$, $q^\ast$ splits the marginal mass among them (an adversarial tie-breaking) and the threshold remains unique. \end{revblock}

Theorem~\ref{thm:optimal_threshold} characterizes how classification thresholds depend on the interplay between model quality, cost asymmetry, and class proportions. \rev{The two regimes are separated by the class proportion. The threshold satisfies $\tau^{\ast} \leq \alpha$ exactly when the tail-weighted share of false positive costs is at most the positive-class share $1-\alpha$, so that at least a share $1-\alpha$ of instances is flagged. For a given cost share $p^\ast$, a higher $k$ moves $\tau^{\ast}$ toward $\alpha$ in both regimes, so a sharper model flags a share of instances closer to the positive-class share whatever the costs.} \rev{Algorithm~\ref{alg:bisection} in Section~\ref{sec:cvarECresults} locates the optimal threshold to a tolerance $\varepsilon$ by bisection on a subgradient of Equation~\eqref{eq:envelope}, in $O(J\log J\,\log(1/\varepsilon))$ operations (Proposition~\ref{prop:algorithm}). The threshold $\tau^{\ast}$ is H\"older continuous of order $1/2$ in the cost parameters for $k < \pi/2$ (Lemma~\ref{lem:continuity}), so nearly tied scenario losses do not destabilize it.} All supplementary technical results and formal proofs are included in Sections \ref{sec:suptheory}-\ref{sec:proofs}.

\subsection{Sensitivity Analyses}\label{sec:sensivititybaseline}

Having characterized \rev{the optimal threshold}, we now examine how \rev{the risk-aware threshold responds} to changes in key problem parameters. By understanding when optimal thresholds are most sensitive to cost structures and risk preferences, we can identify deployment contexts where risk-aware calibration provides the greatest value over standard approaches. \rev{In Proposition~\ref{prop:cost_monotonicity} we hold the worst-case weights $q^\ast$ of Theorem~\ref{thm:optimal_threshold} fixed, so that the comparative statics describe the response of $\tau^{\ast}$ for a given tail weighting. Proposition~\ref{prop:monotonicity_beta} lets the weights vary with $\beta$.}

We first establish how thresholds respond to cost asymmetry.

\begin{revblock}
\begin{proposition}[Threshold Sensitivity to Cost Asymmetry]\label{prop:cost_monotonicity}
Under the trigonometric model constraints $0 < k \leq \pi/2$ and $0 < \alpha < 1$, the optimal threshold $\tau^{\ast}(\bar{r}_{\text{tail}})$ is continuous and strictly increasing in the tail cost ratio $\bar{r}_{\text{tail}} := K_{q^\ast}/L_{q^\ast} > 0$. The two regimes join continuously at $\bar{r}_{\text{tail}} = (1-\alpha)/\alpha$, where $\tau^{\ast} = \alpha$.
\end{proposition}
\end{revblock}

Our analysis demonstrates that optimal classification thresholds must \rev{strictly} increase with the relative cost of false positives\rev{, holding the tail weighting fixed}, establishing a fundamental monotonic relationship between cost structures and decision boundaries. When false positive costs rise due to regulatory penalties, resource constraints, or reputational damage, decision-makers should require stronger evidence before classifying instances as positive. \rev{The threshold responds to the cost asymmetry over its entire range, so cost-blind approaches such as accuracy maximization can become increasingly suboptimal as the asymmetry between error types grows.}

While cost asymmetry provides the primary economic driver of threshold selection, the level of risk aversion fundamentally shapes how decision-makers weight tail scenarios relative to average outcomes. We now examine how optimal thresholds respond to changes in the confidence level $\beta$:

\begin{proposition}[Threshold Sensitivity to Risk Aversion]\label{prop:monotonicity_beta}
Define the sensitivity of scenario $j$ at threshold $\tau$,
\begin{equation}
\rho_j(\tau) := \frac{dS_j(\tau)}{d\tau} = \sum_{i=1}^N \left[K_{ji}(\mathbb{P}(Y=1|\tau) - 1) + L_{ji}\mathbb{P}(Y=1|\tau)\right].
\end{equation}
\rev{Let $\beta$ be such that $m = (1-\beta)J$ is not an integer, let $n = \lfloor m \rfloor$, and suppose that at $\tau^{\ast} = \tau^{\ast}(\beta)$ the $(n+1)$-th largest loss is attained by a single scenario $j^{\ast}$ (the marginal scenario, which carries the fractional weight $(m-n)/m$), strictly separated from the $n$-th and $(n+2)$-th largest losses. Then $\tau^{\ast}(\beta)$ is continuously differentiable near $\beta$ (one-sided if $\tau^{\ast} = \alpha$) and:}
\begin{enumerate}
\item If $\rho_{j^{\ast}}(\tau^{\ast}) > 0$ (marginal scenario is FN-dominated), then $\frac{d\tau^{\ast}}{d\beta} > 0$.
\item If $\rho_{j^{\ast}}(\tau^{\ast}) < 0$ (marginal scenario is FP-dominated), then $\frac{d\tau^{\ast}}{d\beta} < 0$.
\item If $\rho_{j^{\ast}}(\tau^{\ast}) = 0$ (marginal scenario is balanced), then $\frac{d\tau^{\ast}}{d\beta} = 0$.
\end{enumerate}
\rev{If $m$ is an integer, the same rule applies to the one-sided derivatives, with $j^{\ast}$ the $m$-th (right derivative) or $(m+1)$-th (left derivative) largest scenario, provided that these losses are strictly separated from their neighbors.}
\end{proposition}

Proposition~\ref{prop:monotonicity_beta} reveals that increasing risk aversion does not uniformly lead to more conservative thresholds, contrary to conventional intuition. The direction of threshold adjustment depends on whether false negatives or false positives dominate the marginal tail scenario. When the marginal scenario involves costly false negatives (missing critical defects, failed diagnoses), greater risk aversion leads to increases in the optimal threshold. \rev{A higher $\beta$ narrows the tail and removes the weight of the marginal scenario. When that scenario is FN-dominated, its weight favors a lower threshold, so its removal raises $\tau^{\ast}$.} The sensitivity parameter $\rho_{j^{\ast}}(\tau^{\ast})$ captures this effect, measuring how losses change with the threshold at the risk boundary rather than on average, explaining why systems with identical accuracy metrics may require different thresholds under tail risk constraints.

\begin{remark}[Cost Structure and Threshold Behavior]\label{rem:cost_structure}
The relationship between risk aversion and optimal thresholds in Proposition~\ref{prop:monotonicity_beta} depends critically on the heterogeneity of cost structures across scenarios. Two important special cases illustrate this dependence:

\textbf{(i) Proportional costs:} When \rev{$K_{ji} = c_j K_i$ and $L_{ji} = c_j L_i$ for some $c_j > 0$}, all scenarios have the same relative cost trade-offs, differing only in scale. In this case, $\tau^{\ast}(\beta)$ remains constant as $\beta$ varies—risk aversion does not affect the optimal threshold because all scenarios prefer the same decision boundary.

\textbf{(ii) Heterogeneous cost ratios:} When scenarios have different relative costs for false positives versus false negatives, $\tau^{\ast}(\beta)$ can exhibit non-monotonic behavior. As $\beta$ increases, the marginal scenario may shift from one that is FP-dominated to one that is FN-dominated (or vice versa), causing the optimal threshold to change direction.

For example, consider a medical AI system operating under different scenarios: normal operations (balanced costs), resource scarcity (high FP costs due to limited supplies), and epidemic conditions (high FN costs due to contagion risk). As the hospital becomes more risk-averse, the optimal diagnostic threshold might initially increase (to conserve resources) but then decrease sharply (to prevent outbreak spread) \rev{as the tail narrows to the epidemic scenarios}. This non-monotonicity highlights how risk-aware AI calibration must account not just for the magnitude of potential losses, but for the qualitative nature of different risk scenarios.
\end{remark}

\section{The Cost of Ignoring Tail Risk in Threshold Selection}\label{sec:theorybenchmarks}

We now compare our CVaR-based threshold selection presented in Section \ref{sec:baselineriskmanagement} with standard calibration practices, quantifying the hidden costs of ignoring tail risk in algorithmic decision-making. Most practitioners set classification thresholds by maximizing aggregate performance metrics such as accuracy \rev{or} F1-score, implicitly treating all errors as equally costly. This approach fundamentally misaligns with liability reality, where a small fraction of errors can generate severe losses that dominate the risk profile. By contrasting risk-aware thresholds with accuracy maximization—the most prevalent calibration method—we establish when and why traditional approaches fail \rev{ (Section~\ref{sec:acccomparison} proves the supporting lemmas)}. We also briefly consider expected loss minimization as an intermediate benchmark, though we defer detailed analysis to Section \ref{sec:explossmaximizer}.

We begin by analyzing the prevalent calibration approach in practice of setting thresholds to maximize classification accuracy\footnote{For a binary classifier with threshold $\tau$ and uniformly distributed model scores, accuracy is defined as $\text{Acc}(\tau) = \mathbb{P}(\hat{Y} = Y) = \int_\tau^1 \mathbb{P}(Y=1|\overline{\gamma}) d\overline{\gamma} + \int_0^\tau [1 - \mathbb{P}(Y=1|\overline{\gamma})] d\overline{\gamma}$. Thus, under the trigonometric model, accuracy can be expressed as $\text{Acc}(\tau) = \tau + (1-\alpha) - 2\mathbb{P}_{\mathrm{FN}}(\tau)$ (see Lemma \ref{lem:accuracy_form}).}:
\begin{lemma}[Accuracy-Optimal Threshold]\label{lem:accuracy_threshold}
The threshold $\tau_{\mathrm{acc}}$ that maximizes classification accuracy satisfies $\mathbb{P}(Y=1|\tau_{\mathrm{acc}}) = \frac{1}{2}$. Under the trigonometric model with parameters $\alpha \in (0,1)$ and $k \in (0,\pi/2]$:
\begin{enumerate}
\item If $\alpha < \frac{1}{2}$: $\tau_{\mathrm{acc}} = \alpha + \frac{\alpha}{k}\arcsin\left(\sin k \cdot \frac{2\alpha - 1}{2(1-\alpha)}\right)$;
\item If $\alpha = \frac{1}{2}$: $\tau_{\mathrm{acc}} = \alpha$;
\item If $\alpha > \frac{1}{2}$: $\tau_{\mathrm{acc}} = 1 - \frac{(1-\alpha)\sin\left(\frac{k}{2\alpha}\right)}{\sin k}$.
\end{enumerate}
\end{lemma}

Lemma \ref{lem:accuracy_threshold} demonstrates that $\tau_{\mathrm{acc}}$ is uniquely determined by solving $\mathbb{P}(Y=1|\tau) = \frac{1}{2}$, independent of cost parameters. This cost-invariance property implies that accuracy maximization cannot distinguish between scenarios where false negative costs exceed false positive costs by orders of magnitude. This represents a critical limitation in liability sensitive deployments where tail scenarios often exhibit extreme cost asymmetries. To quantify the operational and financial implications of cost-blind threshold selection, we introduce two performance metrics comparing risk-aware and accuracy-based thresholds.

\begin{definition}[Threshold Gap Metrics]
Let $\tau_{\mathrm{acc}}$ denote the accuracy-maximizing threshold and $\tau^{\ast}$ the CVaR-minimizing threshold \rev{ of Theorem~\ref{thm:optimal_threshold}}. We define:
\begin{enumerate}
\item \textit{Efficiency Gap}: $\Delta_\tau = |\tau^{\ast} - \tau_{\mathrm{acc}}|$, measuring the absolute deviation between thresholds;
\item \textit{Risk Penalty}: $\Delta_{\text{CVaR}} = \frac{\text{CVaR}_\beta(S(\tau_{\mathrm{acc}})) - \text{CVaR}_\beta(S(\tau^{\ast}))}{\text{CVaR}_\beta(S(\tau^{\ast}))}$, capturing the relative increase in tail risk.
\end{enumerate}
\end{definition}

The efficiency gap quantifies the magnitude of operational adjustment required to transition from standard practice to risk-aware calibration. The tail risk penalty measures the proportional increase in conditional tail expectation resulting from threshold differences, providing a dimensionless metric for cross-context comparison.\footnote{\rev{Throughout this section, sums over the tail set $\mathcal{J}_\beta$ are weighted by the worst-case weights $q^\ast$ of Theorem~\ref{thm:optimal_threshold}, so that $\sum_{j \in \mathcal{J}_\beta} K_{j}$ stands for $m K_{q^\ast}$. The two coincide whenever $m$ is an integer and the $m$-th and $(m+1)$-th largest losses at $\tau^{\ast}$ differ.}}

\begin{proposition}[Efficiency–gap Characterization]\label{prop:efficiency_gap}
Fix $\alpha \in (0,1)$ and $k\in(0,\pi/2)$.  
Let $\tau^{\ast}$ be the \textnormal{CVaR}–optimal threshold and  
$\tau_{\mathrm{acc}}$ the accuracy–optimal threshold.

\medskip\noindent
\textit{(i) Proportional costs:}  
If $K_{ji}=c_jK_i$ and $L_{ji}=c_jL_i$ for all $i,j$, then
\[
\Delta_\tau=
\begin{cases}
0 & \text{if }\sum_i K_i=\sum_i L_i,\\
>0 & \text{otherwise.}
\end{cases}
\]

\smallskip\noindent
\textit{(ii) Heterogeneous costs:}  
Assume $\tau^{\ast},\tau_{\mathrm{acc}}\in(0,\alpha)$ or  
$\tau^{\ast},\tau_{\mathrm{acc}}\in(\alpha,1)$.  Define the tail-weighted cost ratio
\[
\bar r_{\mathrm{tail}}
:=\frac{\sum_{j\in\mathcal J_\beta}\sum_i K_{ji}}
       {\sum_{j\in\mathcal J_\beta}\sum_i L_{ji}},
\]
and the branch-specific derivative bounds
\[
\rev{p_{\min}}:=\min\!\Bigl\{\tfrac{(1-\alpha)k\cos k}{\alpha\sin k},
                 \tfrac{\alpha\sin k}{k(1-\alpha)}\Bigr\},\qquad
\rev{p_{\max}}:=\max\!\Bigl\{\tfrac{(1-\alpha)k}{\alpha\sin k},
                 \tfrac{\alpha\sin k}{k(1-\alpha)\cos k}\Bigr\}.
\]
With $C_1:=1/\rev{p_{\max}}$ and $C_2:=1/\rev{p_{\min}}$,
\[
C_1\Bigl\lvert\tfrac{\bar r_{\mathrm{tail}}}{1+\bar r_{\mathrm{tail}}}-\tfrac12\Bigr\rvert
\;\le\;
\Delta_\tau
\;\le\;
C_2\Bigl\lvert\tfrac{\bar r_{\mathrm{tail}}}{1+\bar r_{\mathrm{tail}}}-\tfrac12\Bigr\rvert.
\]
\end{proposition}

Proposition \ref{prop:efficiency_gap} establishes that the efficiency gap \rev{is of the order of the deviation of the tail cost share from one half}. The bounds reveal that threshold divergence is inevitable whenever $\bar{r}_{\text{tail}} \neq 1$, with the thresholds coinciding if and only if tail scenarios exhibit perfectly balanced costs between false positives and false negatives. When tail scenarios exhibit substantial cost asymmetry (e.g., $\bar{r}_{\text{tail}} \gg 1$ or $\bar{r}_{\text{tail}} \ll 1$), the operational adjustment required for risk-aware deployment becomes correspondingly large. 

For instance, in a medical diagnostic system where tail scenarios involve malpractice claims \rev{with $\bar{r}_{\text{tail}} = 1/3$ (false negatives cost three times as much as false positives in litigation), the threshold gap satisfies $0.25\,C_1 \leq \Delta_\tau \leq 0.25\,C_2$.} \rev{The ratio $C_2/C_1$ grows as $k$ approaches $\pi/2$, where the calibration map is steepest near $\alpha$ and flattest near the endpoints, so the bounds are most informative for models of moderate quality.} From a deployment perspective, organizations should audit their tail cost ratios before implementing AI systems. A $\bar{r}_{\text{tail}}$ significantly different from 1 signals that standard ML practices will yield substantially suboptimal operational policies, with the exact deviation quantified by our bounds.

While the efficiency gap characterizes the operational adjustment required for risk-aware calibration, the financial consequences of maintaining accuracy-based thresholds prove far more severe. The following result establishes that tail risk exposure grows \rev{at least} quadratically with threshold misalignment\rev{.}

\begin{proposition}[Risk-penalty bounds]\label{prop:risk_penalty}
Fix $\alpha \in (0,1)$, $k\in(0,\pi/2)$, and set $\tau^{\ast}$ (CVaR-optimal) and $\tau_{acc}$ (accuracy-optimal). 

\noindent \textit{(i)  Proportional costs.}  
If $K_{ji}=c_jK_i$ and $L_{ji}=c_jL_i$ (all $i,j$) then
\[
\Delta_{CVaR}=
\begin{cases}
0, & \displaystyle\sum_i K_i=\sum_i L_i,\\[4pt]
\dfrac{S_0(\tau_{\mathrm{acc}})-S_0(\tau^{\ast})}{S_0(\tau^{\ast})},
& \text{otherwise},
\end{cases}
\]
where $S_0(\tau)=\sum_i K_i\mathbb{P}_{\mathrm{FP}}(\tau)+\sum_i L_i\mathbb{P}_{\mathrm{FN}}(\tau)$.

\smallskip\noindent
\rev{\textit{(ii)  Heterogeneous costs.}
Let $q^\ast$ be the worst-case weights at $\tau^{\ast}$ (Theorem~\ref{thm:optimal_threshold}) and $q_{\mathrm{acc}}$ any maximizing weights in \eqref{eq:envelope} at $\tau_{\mathrm{acc}}$, and let $I$ be the closed interval between $\tau^{\ast}$ and $\tau_{\mathrm{acc}}$. Let
\[
\tilde{m}:=\min_{j:\,q^\ast_j>0}\;
      \operatorname*{ess\,inf}_{\tau\in I}
      \partial_{\tau\tau}S_j(\tau),\qquad
\tilde{M}:=\max_{j:\,q_{\mathrm{acc},j}>0}\;
      \operatorname*{ess\,sup}_{\tau\in I}
      \partial_{\tau\tau}S_j(\tau),
\]
both strictly positive and finite by Lemma~\ref{lem:differentiability} and $k < \pi/2$. Then
\[
\frac{\tilde{m}}{2\,\text{CVaR}_\beta\!\bigl(S(\tau^{\ast})\bigr)}\,
(\Delta_\tau)^2
\;\;\le\;\;
\Delta_{CVaR}
\;\;\le\;\;
\frac{\tilde{M}\,(\Delta_\tau)^2/2 \;+\; \Delta_\tau\,\bigl|\sum_j q_{\mathrm{acc},j}\,\rho_j(\tau^{\ast})\bigr|}{\text{CVaR}_\beta\!\bigl(S(\tau^{\ast})\bigr)} .
\]}
\end{proposition}

Proposition~\ref{prop:risk_penalty} quantifies the hidden cost of deploying AI systems based on traditional accuracy metrics rather than risk-aware optimization. The quadratic \rev{lower bound} reveals that \rev{the penalty grows rapidly as the gap widens}. For instance, a 10\% threshold gap ($\Delta_\tau = 0.1$) \rev{yields a risk penalty of at least 20\% whenever $\tilde{m}/\text{CVaR}_\beta(S(\tau^{\ast})) > 40$, which occurs when the calibration map is steep around $\tau^{\ast}$. When the maximizing weights at $\tau_{\mathrm{acc}}$ can be taken equal to $q^\ast$, in particular when the tail set is the same at both thresholds and no losses are tied, the first-order term in the upper bound vanishes and the upper bound becomes $\tilde{M}(\Delta_\tau)^2/(2\,\text{CVaR}_\beta(S(\tau^{\ast})))$.}

The proportional costs case (part i) demonstrates that when all scenarios scale uniformly, risk-aware and accuracy-based deployments diverge only when false positive and false negative costs are imbalanced. This occurs precisely when traditional ML metrics fail to capture the asymmetric nature of real-world consequences. The heterogeneous case (part ii) is more striking, as the bounds depend on the curvature $\partial_{\tau\tau}S_j$ in tail scenarios, which captures how rapidly losses accelerate as the threshold moves away from optimum. In medical AI applications, where tail scenarios might represent preventable patient harm, these second derivatives can be significantly larger than in typical scenarios, making accuracy-based deployment hazardous in high-stakes liability settings.

\begin{figure}[t]
    \centering
    \includegraphics[width=\linewidth]{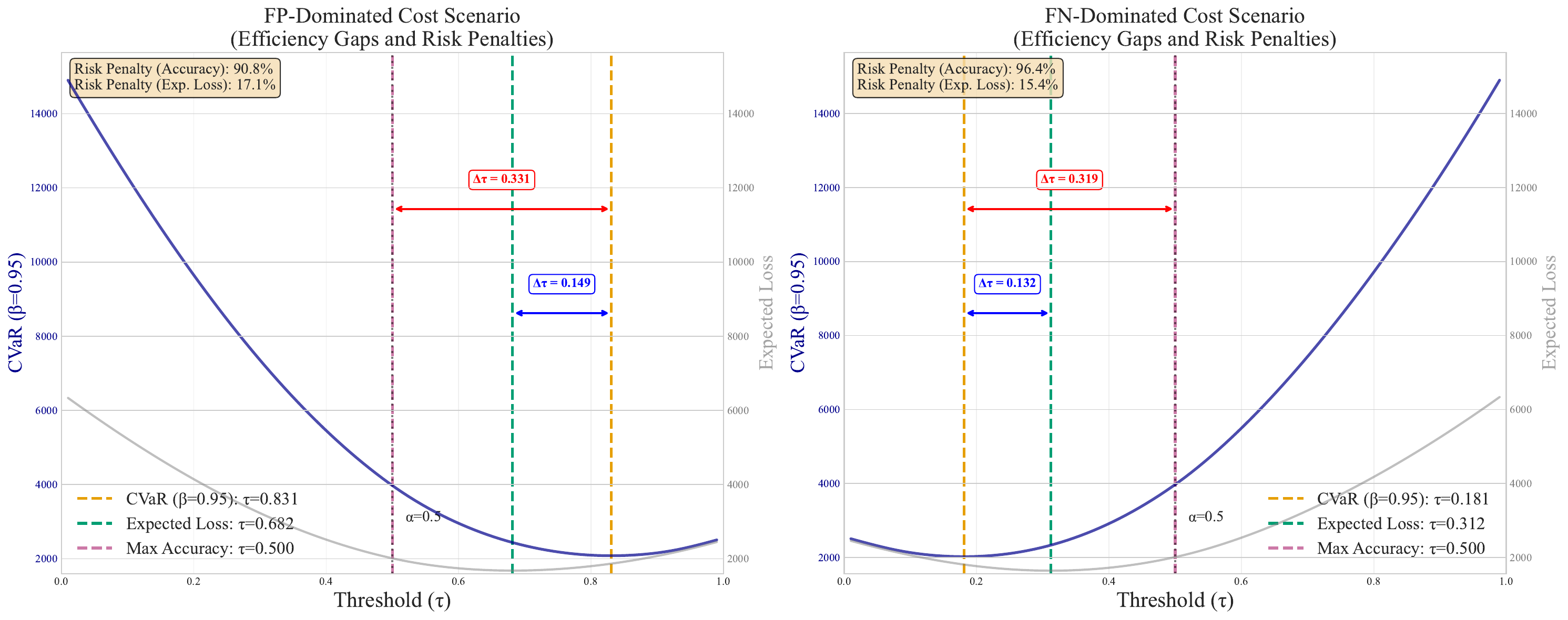}
    \caption{\rev{Accuracy-maximizing and risk-aware thresholds under false-positive-dominated (left) and false-negative-dominated (right) costs, with efficiency gaps (arrows) and CVaR penalties (text boxes).}}
    \label{fig:optimalitygaps}
\end{figure}

Figure \ref{fig:optimalitygaps} illustrates these theoretical insights across representative cost structures. \rev{The left panel depicts false-positive-dominated scenarios, where accuracy-based thresholds prove insufficiently selective. The right panel shows false-negative-dominated settings, where accuracy maximization yields excessive conservatism. In both panels the penalty grows with the threshold gap, approaching 100\% in the scenarios shown.}

While accuracy maximization represents the most common baseline, practitioners might consider minimizing expected loss as a more sophisticated alternative. This approach incorporates cost information through average values rather than tail-weighted metrics. As we demonstrate in Section~\ref{sec:explossmaximizer} \rev{(Proposition~\ref{prop:expected_loss_threshold})}, expected loss minimization yields thresholds that \rev{lie between accuracy and CVaR optimization whenever the average cost share of false positives lies between the tail share and one half, as in our case study}. However, this intermediate positioning provides limited benefit when tail and average cost structures diverge substantially. The analysis reveals that expected loss optimization still incurs risk penalties growing \rev{at least} quadratically with the gap between average and tail cost ratios, confirming that any calibration method neglecting tail heterogeneity may remain inadequate for liability-sensitive deployments.

\section{The Value of Algorithmic Insurance}\label{sec:insurance_value}

Having established that standard calibration approaches could lead to excess tail risk exposure (Section~\ref{sec:theorybenchmarks}), we now analyze when algorithmic liability insurance creates value for firms 
deploying AI systems. We first specify a concrete insurance mechanism suited to AI's unique liability risk profile, and then derive precise conditions under which purchasing insurance dominates the alternative of retaining risk and holding capital against potential losses. 

This section is organized to directly address these questions. Section~\ref{sec:insurance_motivation} presents our proposed insurance structure, involving a dual-limit design with mandatory threshold requirements that leverages AI systems' programmability. Section~\ref{sec:value_mechanisms} decomposes insurance value into three components: risk transfer, capital relief, and calibration improvement, revealing why poorly calibrated firms gain disproportionately from coverage. Finally, Section~\ref{sec:market_viability} establishes precise conditions for value creation, showing that \rev{full coverage creates value whenever the firm's capital charge per dollar of tail risk exceeds the insurer's loadings, and otherwise when the tail-to-average ratio of the insured loss is large enough}.

\subsection{Insurance Design for Algorithmic Liability}\label{sec:insurance_motivation}

We propose an insurance structure that addresses AI's distinctive risk profile, where a single threshold choice generates heterogeneous losses across deployment contexts, with extreme scenarios dominating liability exposure. Drawing from medical malpractice markets, we adopt a dual-limit structure where $L_{\text{occ}} > 0$ caps the insurer's payment per claim and $L_{\text{agg}} \geq L_{\text{occ}}$ caps total payments over the policy period. This design balances three objectives: protecting insurers from catastrophic individual claims, managing concentration risk when tail scenarios generate disproportionate losses, and maintaining insurable premium levels.

The insurer's payment obligation for a firm making $\rev{D}$ decisions (stochastic) during the policy period becomes:
\begin{equation}\label{eq:insurer_payment}
C_{\text{tot}}(\tau, L_{\text{occ}}, L_{\text{agg}}) = \min\left\{\sum_{t=1}^{\rev{D}} \min\{S_t(\tau), L_{\text{occ}}\}, L_{\text{agg}}\right\},
\end{equation}
where $S_t(\tau)$ represents the loss from decision $t$ when operating at threshold $\tau$. \rev{When the loss of a decision comprises an operational component borne by the firm (for instance, the work-up cost of a false positive) and a liability component, the insurer's payment is computed on the liability component only; Proposition~\ref{prop:limit_behavior} accounts for the uninsured component.}

While we model losses deterministically for clarity in this formulation, the framework extends naturally to stochastic claim emergence. \rev{When every cost, including the operational one, is scaled by the same emergence rate $\phi$, the optimal threshold remains unchanged, since CVaR is positively homogeneous.}  When claim emergence rates differ by error type—transforming costs to $\tilde{K}_{ji} = \phi_{\text{FP}} K_{ji}$ and $\tilde{L}_{ji} = \phi_{\text{FN}} L_{ji}$—the optimal threshold shifts according to Theorem~\ref{thm:optimal_threshold} with these modified parameters. This flexibility allows practitioners to interpret our cost parameters as either maximum potential liability or expected realized losses after accounting for litigation friction.

Premium pricing follows our CVaR framework, reflecting the tail risk nature of liability:
\begin{equation}\label{eq:premium}
\color{black}\Pi(L_{\text{occ}}, L_{\text{agg}}, \tau) = (1 + \mu) \cdot \mathbb{E}[C_{\text{tot}}] + r \cdot \left(\text{CVaR}_\beta[C_{\text{tot}}] - \mathbb{E}[C_{\text{tot}}]\right), \qquad C_{\text{tot}} = C_{\text{tot}}(\tau, L_{\text{occ}}, L_{\text{agg}}),\color{black}
\end{equation}
\rev{where $\mu > 0$ is the loading on expected payouts covering administrative costs and profit margins, $r > 0$ is the insurer's cost of holding capital against the tail of the insured payments (the cost-of-capital risk margin of Solvency II), and $\beta$ captures the insurer's tail risk aversion. Setting $r = 1 + \mu$ applies the full loading to the tail of the insured payments, $\Pi = (1+\mu)\,\text{CVaR}_\beta[C_{\text{tot}}]$, a more conservative premium that we refer to as tail-priced and also report in Section~\ref{sec:insurance_sensitivity_ec}.}

The distinctive feature of our proposed insurance is the mandatory threshold requirement. Unlike traditional professional liability coverage, where human behavior is difficult to monitor and control, AI systems can be programmatically constrained to operate at specific thresholds. Insurers can require operation at the CVaR-optimal threshold $\tau^{\ast}$ as a coverage condition, verified through system audits and embedded controls. This requirement serves three purposes\rev{. It aligns the deployed threshold with the tail-risk-optimal one, it makes the deployed threshold observable, which limits the scope for adverse selection, and it creates value for suboptimally calibrated firms through mandated improvements.}

\rev{We propose a verification mechanism that exploits the fact that a threshold is a single configuration parameter. The insurer can require a signed configuration, decision logs that record the score and the threshold in force, and periodic or claim-triggered audits. Coverage can then exclude any claim arising from a decision taken at another threshold. As with warranties and audit rights in professional liability policies, this exclusion makes the mandate self-enforcing (Section~\ref{sec:limitations} discusses the limits of enforceability). The threshold mandate transforms insurance from pure risk transfer into a governance mechanism.} Firms must choose between maintaining their current threshold and retaining all liability risk, or adopting the risk-optimal threshold to access coverage. As we show in the following sections, this choice creates substantial value for firms whose calibration approaches ignore tail risk, while still benefiting risk-aware firms through traditional risk transfer and capital relief.

\subsection{Value Creation Mechanisms}\label{sec:value_mechanisms}
To understand when insurance creates value, we compare a firm's economics with and without coverage. Consider a firm deploying an AI system generating revenue $R$ per period. Without insurance, the firm deploys only if revenue exceeds expected losses plus required capital cost:
\begin{equation}\label{eq:deployment_no_insurance}
R \geq \mathbb{E}[S(\tau_\theta)] + \rho\gamma \cdot \text{CVaR}_\beta[S(\tau_\theta)],
\end{equation}
where $\tau_\theta$ represents the firm's chosen threshold (varying by type $\theta$), $\rho$ is the cost of capital, and $\gamma$ is the regulatory capital multiplier for tail risk. With insurance, the deployment condition changes in two ways. First, the firm operates at the mandated threshold $\tau^{\ast}$ rather than $\tau_\theta$. Second, regulatory capital requirements decrease due to risk transfer:
\begin{equation}\label{eq:deployment_with_insurance}
R \geq \mathbb{E}[S_{\text{net}}(\tau^{\ast})] + \Pi + \rho\gamma_{\text{ins}} \cdot \text{CVaR}_\beta[S_{\text{net}}(\tau^{\ast})],
\end{equation}
where $S_{\text{net}} = S - C_{\text{tot}}$ represents retained losses, $\Pi$ is the insurance premium, and $\gamma_{\text{ins}} < \gamma$ reflects regulatory recognition of risk transfer. The latter assumption captures the regulatory view that insurance reduces systemic risk and provides external monitoring of operational practices. The magnitude of capital relief $(\gamma - \gamma_{\text{ins}})$ depends on factors including insurer credit ratings, contract terms, and regulatory frameworks \citep{basel2006international}. The value of insurance equals the reduction in required revenue:

\begin{theorem}[Insurance Value Decomposition]\label{thm:insurance_value}
The value of insurance with limits $(L_{\text{occ}}, L_{\text{agg}})$ for a firm of type $\theta$ equals:
\begin{equation}\label{eq:value_full}
V_{\theta}(L_{\text{occ}}, L_{\text{agg}}) = V_{\mathrm{base}}(L_{\text{occ}}, L_{\text{agg}}) + V_{\text{calib}}^{\theta},
\end{equation}
where the base insurance value
\begin{align}\label{eq:base_value}
V_{\mathrm{base}} &= \underbrace{\mathbb{E}[C_{\text{tot}}(\tau^{\ast})] - \Pi}_{\text{Risk Transfer}} + \underbrace{\rho\gamma \cdot \text{CVaR}_\beta[S(\tau^{\ast})] - \rho\gamma_{\text{ins}} \cdot \text{CVaR}_\beta[S_{\text{net}}(\tau^{\ast})]}_{\text{Capital Relief}}
\end{align}
represents traditional insurance benefits\rev{, with all insured quantities evaluated at the mandated threshold $\tau^{\ast}$ so that $V_{\mathrm{base}}$ does not depend on the firm type}, while the calibration value
\begin{align}\label{eq:calibration_value}
V_{\text{calib}}^{\theta} &= \underbrace{\mathbb{E}[S(\tau_\theta) - S(\tau^{\ast})]}_{\text{Expected Loss Reduction}} + \underbrace{\rho\gamma[\text{CVaR}_\beta[S(\tau_\theta)] - \text{CVaR}_\beta[S(\tau^{\ast})]]}_{\text{Tail Risk Reduction}}
\end{align}
captures operational improvements from threshold optimization.
\end{theorem}

This decomposition reveals three distinct value creation mechanisms. First, the risk transfer component $\mathbb{E}[C_{\text{tot}}(\tau^{\ast})] - \Pi$ \rev{is never positive. By Equation~\eqref{eq:premium} it equals $-\mu\,\mathbb{E}[C_{\text{tot}}] - r\,(\text{CVaR}_\beta[C_{\text{tot}}] - \mathbb{E}[C_{\text{tot}}]) \leq 0$, since $\mu, r > 0$ and $\text{CVaR}_\beta \geq \mathbb{E}$. It is strictly negative whenever the insurer's expected payment is positive.} This apparent cost is offset by the value of converting volatile tail losses into predictable premium expenses, particularly for firms facing capital constraints or earnings volatility penalties.

Second, capital relief emerges through the differential regulatory treatment of insured versus uninsured risks. The value generated equals $\rho[\gamma \cdot \text{CVaR}_\beta[S(\tau^{\ast})] - \gamma_{\text{ins}} \cdot \text{CVaR}_\beta[S_{\text{net}}(\tau^{\ast})]]$, which depends on three factors: the tail risk transferred to insurers, the regulatory capital differential $(\gamma - \gamma_{\text{ins}})$, and the cost of capital $\rho$.\footnote{\rev{Regulatory recognition is bounded in practice. The Basel framework caps the reduction of the operational-risk capital charge obtained through insurance at 20\% \citep{basel2006international}, so that $\gamma_{\text{ins}} \geq 0.8\gamma$, but even this bounded relief generates value when tail exposures are severe (Section~\ref{sec:valueinsurance_case}).}}

Third, calibration improvement provides value exclusively to firms operating at suboptimal thresholds\rev{, and is positive for them only when the reduction in the tail-risk charge outweighs any increase in expected loss}. The magnitude of $V_{\text{calib}}^{\theta}$ depends directly on the efficiency gap $|\tau_\theta - \tau^{\ast}|$ established in Section~\ref{sec:theorybenchmarks}. For accuracy-maximizing firms, where risk penalties grow \rev{at least} quadratically with threshold deviation, this component can dominate total insurance value. Risk-aware firms, already operating at $\tau^{\ast}$, receive $V_{\text{calib}}^{\text{risk-aware}} = 0$ and benefit solely from risk transfer and capital relief.

The relative magnitudes of the value creation mechanisms vary systematically with firm characteristics. Risk-aware firms with strong balance sheets benefit primarily from capital efficiency, as their optimal calibration eliminates operational improvement opportunities. Conversely, poorly calibrated firms with limited capital access gain from \rev{the capital relief and calibration channels} simultaneously, creating substantially higher total value. This heterogeneity in value creation has direct implications for insurance adoption and optimal contract design, which we examine in the following section.

\subsection{Market Viability and Conditions for Value Creation}\label{sec:market_viability}

Having decomposed insurance value into its components, we now establish precise conditions for when purchasing third-party insurance coverage dominates retaining risk within the firm. The analysis reveals that value creation depends critically on the relationship between tail risk severity, insurance pricing, and the firm's calibration approach.

Since risk-aware firms already operate at optimal thresholds, they capture value only through traditional insurance mechanisms. The value creation dynamics depend critically on coverage limits.
\begin{revblock}
\begin{proposition}[Limit Behavior of Value]\label{prop:limit_behavior}
Write $S(\tau^{\ast}) = s_{\mathrm{op}} + S_L(\tau^{\ast})$, with $s_{\mathrm{op}} \geq 0$ the uninsured operational component, constant across scenarios at a fixed threshold, and $S_L$ the liability component on which the insurer pays. At the limit, the base value function behaves as:
\begin{enumerate}
    \item As $L_{\text{occ}}, L_{\text{agg}} \to 0$: 
    $V_{\mathrm{base}} \to \rho(\gamma - \gamma_{\text{ins}}) \cdot \text{CVaR}_\beta[S(\tau^{\ast})]$
    (pure capital relief with no risk transfer)
    \item As $L_{\text{occ}}, L_{\text{agg}} \to \infty$: 
    $V_{\mathrm{base}} \to (\rho\gamma - r)\,\text{CVaR}_\beta[S_L(\tau^{\ast})] - (\mu - r)\,\mathbb{E}[S_L(\tau^{\ast})] + \rho(\gamma - \gamma_{\text{ins}})\,s_{\mathrm{op}}$
    (full risk transfer of the liability with capital relief, net of the insurer's loadings)
\end{enumerate}
\end{proposition}

\end{revblock}
Proposition~\ref{prop:limit_behavior} reveals two distinct sources of value. With minimal coverage, value arises purely from regulatory capital relief—the differential treatment of insured versus uninsured risks. \rev{With full coverage, the firm gains when its capital charge per dollar of tail risk ($\rho\gamma$) exceeds the insurer's loadings $\mu$ and $r$. Corollary~\ref{cor:viability} in Section~\ref{sec:theoryinsurance} makes the condition explicit. With $s_{\mathrm{op}} = 0$ and $\rho\gamma > r$, full coverage creates value if and only if the tail-to-average ratio of the insured loss exceeds $(\mu - r)/(\rho\gamma - r)$, and it always does when $\rho\gamma > \max\{\mu, r\}$.} As coverage increases, risk transfer grows but is offset by premium loadings. The optimal coverage balances marginal premium costs against marginal risk reduction and capital relief benefits.

These results reveal how operational decisions at the deployment level drive insurance value creation. When ML clients employ risk-aware thresholds, AI providers capture value primarily through capital relief, with risk transfer benefits constrained by the relationship between tail risk severity and premium loadings (see \rev{Figure~\ref{fig:valuecomponents_case}}); when clients operate with standard accuracy-based calibration, providers capture additional value through insurance-mandated threshold improvements at client sites. The three-party structure—insurer, provider, and client—\rev{thus} transforms algorithmic insurance from traditional indemnification into a governance mechanism, and suggests market segmentation based on client operational sophistication, with early adoption concentrated among providers whose clients have not internalized tail risk in their threshold decisions.

\section{Framework Extensions}\label{sec:extensions}

Our baseline framework assumes stationary model performance and fully automated decisions. In practice, performance degradation over time and the potential for human oversight to avert errors can also critically affect algorithmic liability. This section analyzes how these considerations shape insurance contract design. Section \ref{sec:drift} characterizes optimal contract duration when models decay exponentially. Section \ref{subsec:interpretability} demonstrates how interpretable AI systems that facilitate error detection can enable lower premiums, with optimal transparency levels determined by the trade-off between implementation costs and risk reduction benefits. These extensions provide practical guidance for structuring algorithmic insurance in dynamic human-in-the-loop environments.

\subsection{Performance Decay due to ML Model Drift}\label{sec:drift}

Deployed ML models invariably experience performance degradation due to distribution shifts between training and deployment environments. This decay directly impacts error rates and insurer tail risk exposure, altering optimal contract design. We extend our trigonometric model by allowing model quality to decay over time:
\begin{equation}
k(t) = k_0 \cdot h(t),
\end{equation}
\rev{where $k_0 \in (0, \pi/2]$ represents initial model quality and $h(t)$ is a continuous degradation function satisfying $h(0) = 1$ (no degradation at deployment) and non-increasing (monotonic decay).} This degradation directly affects the error probabilities $\mathbb{P}_{\mathrm{FP}}(\tau, t)$ and $\mathbb{P}_{\mathrm{FN}}(\tau, t)$ through the time-varying quality parameter.
We adopt the exponential decay $h(t) = e^{-\delta t}$ where $\delta > 0$ is the drift rate. This specification aligns with \rev{the performance degradation documented} across domains including credit scoring and fraud detection \citep{gama2014survey,sousa2014two}. Under this model, false positive and false negative rates increase over time, leading to monotonically growing tail risk for any fixed threshold (see Lemma \ref{lem:cvar_drift}\rev{, which holds for any family of models whose ROC curves are nested in the quality parameter}).\rev{ The same decay applies to the separation parameter $d$ of the binormal model (Section~\ref{sec:binormal}).}

The degradation of model quality creates a tension in contract design. Extended coverage periods amortize administrative costs but require insurers to price increasing tail risk from model decay. \rev{The premium is set at inception, so drift accumulates an unpriced excess tail risk until renewal, which restores the inception quality $k_0$ at an administrative and retraining cost $\mathcal{C}_a > 0$. Carrying a unit of excess tail risk for a unit of time costs the firm its capital charge on retained risk, $\lambda = \rho\gamma_{\text{ins}}$ (Section~\ref{sec:value_mechanisms}). We define the excess-risk cost rate under the mandated threshold,
\begin{equation}\label{eq:excess_rate}
e(t) := \lambda\bigl[\text{CVaR}_\beta(S(\tau^{\ast}, t)) - \text{CVaR}_\beta(S(\tau^{\ast}, 0))\bigr] \geq 0, \qquad e(0) = 0,
\end{equation}
which is continuous and non-decreasing by Lemma~\ref{lem:cvar_drift}. Renewing every $T$ periods then costs, per unit time, $\mathcal{A}(T) = \bigl(\mathcal{C}_a + \int_0^T e(t)\,dt\bigr)/T$.} We formalize this trade-off between operational efficiency and risk accumulation:

\begin{revblock}
\begin{proposition}[Optimal Contract Duration under Drift]
\label{prop:optimal_duration_full}
Let $H(T) := \int_0^T [e(T) - e(t)]\,dt$ and $H_\infty := \sup_{T > 0} H(T)$.
\begin{enumerate}
\item[(i)] If $\mathcal{C}_a < H_\infty$, the average cost $\mathcal{A}$ has a finite minimizer, every minimizer $T^{\ast}$ satisfies
\begin{equation}\label{eq:F0_condition_full}
\int_0^{T^{\ast}} \bigl[e(T^{\ast}) - e(t)\bigr]\,dt = \mathcal{C}_a ,
\end{equation}
and $T^{\ast}$ is unique when $e$ is strictly increasing. If $\mathcal{C}_a > H_\infty$, no finite renewal time is optimal.
\item[(ii)] $T^{\ast}$ is non-decreasing in the renewal cost $\mathcal{C}_a$.
\item[(iii)] If $e_1 \geq e_2$ are the excess-risk rates of two drift processes and $e_1 - e_2$ is non-decreasing on $[0, T_2^{\ast}]$, where $T_2^{\ast}$ is the largest minimizer under $e_2$, then the smallest minimizer under $e_1$ satisfies $T_1^{\ast} \leq T_2^{\ast}$.
\end{enumerate}
\end{proposition}

\end{revblock}

\rev{Two corollaries in Section~\ref{sec:theorydrift} complement the proposition. Under linearly growing excess risk the optimal renewal time is $T^{\ast} = \sqrt{2\mathcal{C}_a/s}$, where $s$ is the growth rate of the excess-risk cost (Corollary~\ref{cor:sqrt}), and Corollary~\ref{cor:recalibration} compares re-optimizing the threshold on a degraded model with retraining it.} This result reveals that optimal contract duration balances the marginal cost of renewal against accumulated value loss from operating with degraded models. When drift rates increase (whether in rapidly evolving domains like content moderation or due to underlying market shifts) insurers should offer shorter contracts despite higher administrative burden. Conversely, stable applications like industrial quality control support longer coverage periods.

Our findings indicate that insurers should segment markets by drift characteristics, offering shorter contracts for high-drift applications (financial fraud detection, recommendation systems) while extending longer period coverage for stable domains (medical imaging, industrial inspection). Premium structures should explicitly account for expected degradation, with contracts potentially including performance-based renewal triggers when $k(t)$ falls below predetermined thresholds.
\rev{For ML providers, these results quantify the insurance value of model maintenance and separate two responses to drift. Re-optimizing the threshold on the degraded model slows the growth of tail risk within the contract, but only retraining restores the inception risk. A slower drift, in turn, enables longer, more cost-effective insurance contracts.} This creates a self-reinforcing mechanism where insurance incentivizes proactive model management, ultimately reducing system-wide liability exposure.

\subsection{Incorporating Human Oversight in Risk Assessment}
\label{subsec:interpretability}

High-stakes AI deployments rarely operate without human oversight. The degree of model interpretability directly affects whether humans can identify and prevent algorithmic errors before they materialize into claims \citep{DeBockEtAl2024ExplainOR}, fundamentally altering liability exposure and insurance pricing. Our interpretability approach aligns with established frameworks in the explainable AI literature, where transparency enables human oversight of algorithmic decisions \citep{rudin2019stop}. This operational view of interpretability focuses on humans identifying and preventing errors in real-time. It differs from post-hoc model updating or feedback-based improvements, which represent a separate paradigm not captured in our current framework.

\begin{figure}[htbp]
    \centering
    \begin{subfigure}[h]{0.49\textwidth}
        \centering
\includegraphics[width=\linewidth]{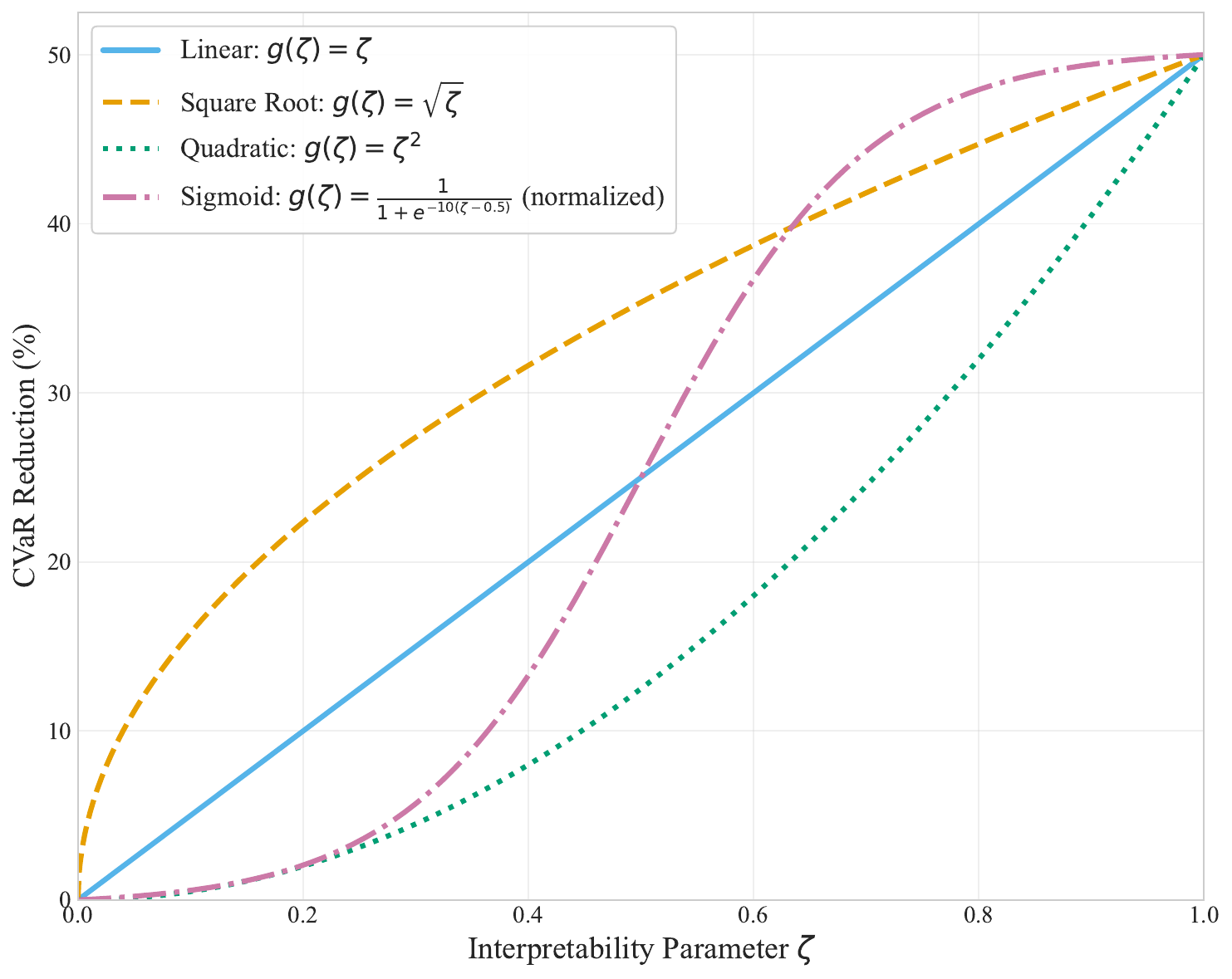}
        \vspace{4pt}
        \caption{CVaR effect of $g(\zeta)$.}
    \end{subfigure}
    \hfill
    \begin{subfigure}[h]{0.49\textwidth}
        \centering        \includegraphics[width=\linewidth]{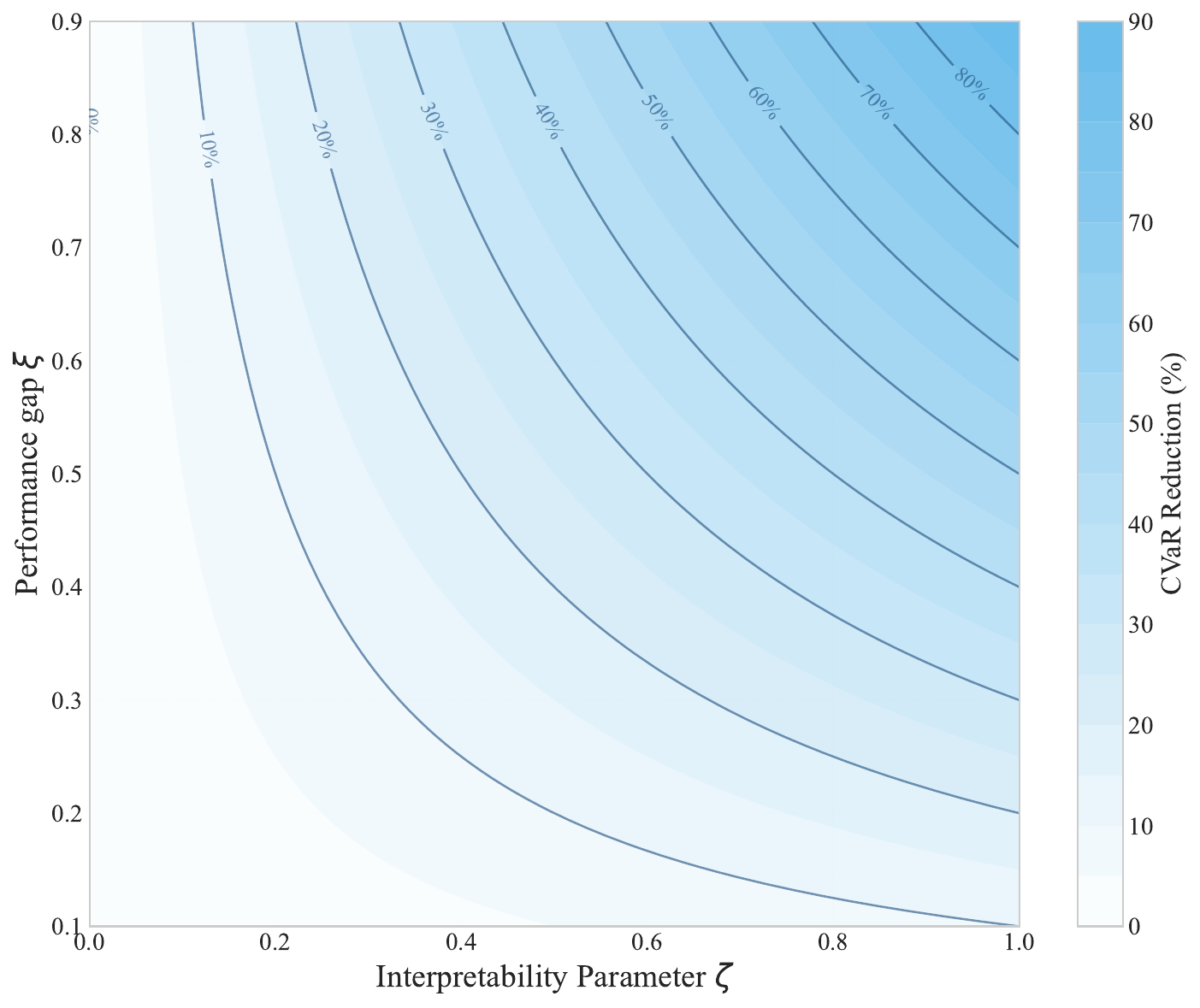}
        \vspace{4pt}
        \caption{CVaR effect of $\xi$ and $\zeta$.}
    \end{subfigure}
    \caption{Interpretability and residual tail risk. CVaR reduction as a function of $\zeta$ for $\xi = 0.5$ (a), and of $\zeta$ and $\xi$ for $g(\zeta)=\zeta$ (b).}
    \label{fig:interpretability_ill}
\end{figure}

We model interpretability's risk-reducing effect through the parameter $\zeta \in [0,1]$, where $\zeta = 0$ represents a black box and $\zeta = 1$ indicates full transparency. When $\tau^{\ast}$ is fixed, residual risk becomes:
\begin{equation}
\text{CVaR}_\beta(S(\tau^{\ast}, \zeta)) = \text{CVaR}_\beta(S(\tau^{\ast})) \cdot (1 - \xi g(\zeta)),
\end{equation}
where $\xi = \text{CVaR}_\beta(S(\tau^{\ast}))/\text{CVaR}_\beta(S_{\text{human}}) \in (0,1)$ captures the algorithm-versus-human performance gap, and $g: [0,1] \to [0,1]$ is \rev{a continuous} increasing function satisfying $g(0) = 0$ and $g(1) = 1$ that characterizes the risk reduction achieved through interpretability\footnote{The function $g(\zeta)$ may be linear (constant marginal benefits), concave (diminishing returns from cognitive limits), or convex (threshold effects where understanding jumps at critical explanation levels).}. Figure~\ref{fig:interpretability_ill} in Section~\ref{sec:suppfigures} illustrates how increasing interpretability reduces residual tail risk, with steeper reductions under convex mappings and \rev{higher values of $\xi$}.

This multiplicative specification reflects our modeling assumption that interpretability enables humans to catch a fraction $\xi g(\zeta)$ of errors that would otherwise occur\rev{. Losses and insured payments then scale by $1 - \xi g(\zeta)$. When the coverage limits do not bind, the premium of Equation~\eqref{eq:premium} at the mandated threshold scales by the same factor, $\Pi(\zeta) = \Pi(\tau^{\ast})(1 - \xi g(\zeta))$, and we retain this scaling as a simplifying assumption when they bind}. The effectiveness depends on both the human-AI performance differential and the quality of explanations, consistent with empirical findings in manufacturing and healthcare settings \citep{senoner2022using,caruana2015intelligible}. While alternative specifications could model interpretability as affecting error distributions directly, our approach captures the primary mechanism in practice: transparent models allow human experts to identify and override problematic decisions before they generate liability\footnote{This specification also ensures threshold invariance—the optimal $\tau^{\ast}$ remains unchanged regardless of $\zeta$—which aligns with empirical observations that interpretability tools typically augment rather than replace existing decision rules.}.

Achieving interpretability level $\zeta$ requires investment $c(\zeta)$, where $c$ is \rev{continuous,} increasing and convex, reflecting escalating engineering effort and user training costs. The optimal transparency level balances implementation costs against risk reduction benefits:

\begin{revblock}
\begin{proposition}[Optimal interpretability investment]
\label{prop:optimal_zeta}
Fix $\tau = \tau^{\ast}$ and let $A := \Pi(\tau^{\ast})$. Let $g$ and $c$ be differentiable on $(0,1)$ with $g' > 0$ and $c$ increasing, and let $\psi(\zeta) := c'(\zeta)/g'(\zeta)$ be continuous and strictly increasing on $(0,1)$. Then the value-maximizing interpretability level $\zeta^\ast$ is unique. It is interior if and only if $\psi(0^+) < A\xi < \psi(1^-)$, in which case $c'(\zeta^\ast) = A\,\xi\,g'(\zeta^\ast)$. The condition on $\psi$ holds whenever $c$ is convex and $g$ concave with at least one of them strict, and for $c(\zeta) = c_0\zeta^a$, $g(\zeta) = \zeta^b$ with $a > b > 0$, where
\begin{equation}\label{eq:zetastar}
\zeta^\ast = \min\Bigl\{1,\ \Bigl(\frac{A\,\xi\,b}{c_0\,a}\Bigr)^{1/(a-b)}\Bigr\}.
\end{equation}
\end{proposition}
\end{revblock}

\noindent \rev{The proposition characterizes the oversight level that the insurer's premium schedule induces, with the premium saving as the benefit of oversight. In the power family, $\zeta^\ast$ has elasticity $1/(a-b)$ with respect to the premium and $-1/(a-b)$ with respect to the cost scale. A firm that also counts the reduction of its retained losses and of its capital charge would choose a higher level. With this larger benefit base the closed form of Equation~\eqref{eq:zetastar} applies unchanged and, at the inputs of the case study, yields full oversight (Section~\ref{sec:renewal_oversight_ec}).} This finding highlights that insurers could offer premium discounts for verifiable interpretability features, with the discount schedule following $\xi g(\zeta)$. ML providers facing high tail risk, such as medical diagnostics or autonomous systems, can justify substantial interpretability investments through insurance savings. 

\section{Case Study}\label{sec:casestudy}

To \rev{illustrate} our theoretical framework, we examine AI-assisted mammography, a clinical application in healthcare where algorithmic deployment creates novel liability risks that diverge from established medical malpractice patterns. Mammography exemplifies our modeling features through heavy-tailed risk distributions from missed cancer diagnoses and heterogeneous cost structures across clinical contexts. We analyze two contrasting deployment scenarios with varying cost structures and operational constraints. \rev{Section \ref{sec:casesetup} specifies the settings, Section \ref{sec:calibration_case} compares threshold selection methods, Section \ref{sec:valueinsurance_case} decomposes insurance value, and Section \ref{sec:extensions_case} analyzes renewal under model drift and human oversight.}

\subsection{Case Study Setup}\label{sec:casesetup}

We examine two mammography settings that differ fundamentally in disease prevalence, operational context, and liability exposure. In the absence of algorithm-specific litigation history, we employ a structured scenario-based approach that synthesizes malpractice databases, settlement analyses, and treatment cost studies to model liability outcomes. \rev{Each error carries two kinds of cost. A false positive incurs an operational work-up cost, whether or not a claim follows. We treat the work-up cost as part of the ML provider's loss, as if the provider and its client acted as a single decision-maker. A missed cancer, and more rarely a false positive, generates liability when it leads to a paid malpractice claim. Following Section~\ref{sec:insurance_motivation}, the insurer pays on the liability component only. The construction of the scenarios ($J = 1{,}000$), their calibration to published claim statistics, and the provenance of every input, with all amounts in 2025 US dollars, are detailed in Section~\ref{sec:scenariolosses}. Every result below is the mean over 100 independent draws of the $J = 1{,}000$ scenarios, and the tables report the 2.5--97.5\% range across draws in parentheses.}

\paragraph{Setting I -- Population Screening:}
Screening mammography targets asymptomatic women aged 40-74 to identify breast cancer before clinical manifestation, representing a high-volume and low-prevalence setting. \rev{We set the cancer prevalence to $0.8\%$ ($\alpha = 0.992$), a round value close to the 5.9 cancers per 1{,}000 screens of US screening practice (5.1 detected and 0.8 missed) \citep{lehman2017national}, and consider an annual volume of 50,000 examinations.\footnote{\rev{The volumes represent a large hospital network or regional screening service and a separate enriched diagnostic cohort.}} Section~\ref{sec:robustness_ec} varies the prevalence between 0.4\% and 1.2\%. Reported AUCs of AI systems in screening populations range from about 0.81 to 0.93 \citep{mckinney2020international, chen2023performance}; since performance in new populations is typically below that in retrospective development studies \citep{yu2022external}, we set $\text{AUC} = 0.85$. This prevalence lies outside the range of the trigonometric model (Section~\ref{app:trigonometric_validation}), so we use the binormal model of Section~\ref{sec:binormal} (Lemma~\ref{lem:binormal}, $d = 1.47$) for this setting.} The cost structure reflects screening's fundamental asymmetry as false positives generate bounded costs through unnecessary \rev{work-up}, while false negatives incur severe liability from delayed diagnosis. Tail scenarios capture operational contexts with heterogeneous cost structures, where scenario-specific factors can induce either positive or negative correlation between false positive and false negative losses.

\paragraph{Setting II -- Diagnostic Testing:} This setting evaluates symptomatic women or abnormal screening findings\rev{, at an annual volume of 6{,}000 examinations. We set the cancer prevalence to $15\%$ ($\alpha = 0.85$), a value representative of an enriched diagnostic cohort. In US diagnostic practice, cancer is detected in about 3.5\% of all diagnostic examinations and 6.5\% of those for a palpable lump, while 28--35\% of the lesions recommended for biopsy prove malignant \citep{sprague2017national}, so our value lies between these rates. Well-validated models reach AUCs of 0.93--0.97 on external test sets that include a symptomatic cohort \citep{lotter2021robust} and 0.84 on enriched reader-study sets \citep{rodriguez2019stand}. We set $\text{AUC} = 0.90$, which the trigonometric model attains at $k = 1.549$.} Cost asymmetry becomes more prominent as missing cancer in symptomatic patients constitutes clear negligence under heightened standards of care. Liability for missed cancers intensifies under diagnostic care standards, while false positive costs remain bounded despite more aggressive interventions. The scenario-based cost structure, detailed in Section \ref{sec:scenariolosses}, captures how heterogeneity in diagnostic settings can create varying correlation patterns between false positive and false negative losses.

\paragraph{\rev{Costs and Claims:}} \rev{We set the cost of a false positive to \$1{,}191 per false-positive screen, the health-plan expenditure on care within 12 months of a false-positive mammogram \citep{ong2015}, and \$2{,}516 per benign biopsy in the diagnostic setting \citep{vlahiotis2018}, both inflated with the CPI-U medical care index (Section~\ref{sec:provenance}). Medical malpractice exhibits substantial friction, as only a small fraction of negligent injuries generate claims, and only part of those are paid. We set the probability that a missed cancer leads to a paid claim at $2\%$, consistent with published claim-filing and payment rates and with the share of missed screening cancers that led to claims \citep{studdert2006claims, vanbreestsmallenburg2012}. For false positives we assume $0.1\%$, since lawsuits over unnecessary work-up are uncommon (Section~\ref{sec:provenance}). Before the scenario multipliers, a paid missed-cancer claim costs \$250{,}000 in screening and \$500{,}000 in the diagnostic setting, below the published averages of \$438{,}047 \citep{dick2009} and \$978{,}858 for litigated cases \citep{lee2020breast}. We assume \$15{,}000 and \$25{,}000 per paid false-positive claim and calibrate scenario multipliers to national closed-claim statistics \citep{studdert2006claims, schaffer2017} to generate the tail costs (Section~\ref{sec:scenariolosses}).}

\subsection{ML Calibration under Heterogeneous Costs}\label{sec:calibration_case}

We now quantify how alternative threshold selection methods affect tail risk exposure in our mammography settings. Table \ref{tab:comprehensive_risk_analysis} reports the efficiency gaps and risk penalties that arise when traditional calibration approaches are applied instead of CVaR optimization\rev{, together with the operating point (share of examinations flagged, sensitivity, positive predictive value) and the expected numbers of errors and paid claims at the risk-aware threshold}. The analysis compares three calibration methods (CVaR minimization at the 95th percentile, expected loss, and accuracy maximization), presenting how heterogeneous cost structures determine varying levels of optimal thresholds and liability exposure across clinical contexts.

\begin{table}[htbp]
\scriptsize
\centering
\color{blue}
\setlength{\tabcolsep}{3pt}
\begin{tabular}{@{}lcc@{}}
\toprule
\textbf{Metric} & \textbf{Screening} & \textbf{Diagnostic} \\ \midrule
\multicolumn{3}{c}{\textit{Scenario Parameters}} \\ \midrule
Cancer prevalence ($1-\alpha$) & 0.8\% & 15\% \\
Performance model (AUC) & binormal (0.85) & trigonometric (0.90) \\
Annual volume & 50,000 & 6,000 \\
Work-up cost per false positive & \$1,191 & \$2,516 \\
Paid-claim probability (FP / FN) & 0.1\% / 2\% & 0.1\% / 2\% \\ \midrule
\multicolumn{3}{c}{\textit{Optimal Thresholds}} \\ \midrule
$\tau^{\ast}$ (CVaR-optimal) & 0.8410\,(0.8226--0.8566) & 0.2903\,(0.2801--0.3008) \\
$\tau_{\text{EL}}^\ast$ (Expected Loss) & 0.9915\,(0.9905--0.9926) & 0.6324\,(0.6181--0.6526) \\
$\tau_{\mathrm{acc}}$ (Accuracy) & 0.9999 & 0.8815 \\ \midrule
\multicolumn{3}{c}{\textit{Operating Points}} \\ \midrule
Share flagged at $\tau^{\ast}$ & 15.9\%\,(14.3\%--17.7\%) & 71.0\%\,(69.9\%--72.0\%) \\
Share flagged at $\tau_{\text{EL}}^\ast$ / $\tau_{\mathrm{acc}}$ & 0.8\%\,(0.7\%--0.9\%) / 0.01\% & 36.8\%\,(34.7\%--38.2\%) / 11.85\% \\
Sensitivity at $\tau^{\ast}$ & 0.67\,(0.65--0.70) & 0.98\,(0.98--0.99) \\
Sensitivity at $\tau_{\text{EL}}^\ast$ / $\tau_{\mathrm{acc}}$ & 0.16\,(0.15--0.17) / 0.01 & 0.86\,(0.85--0.87) / 0.61 \\
Positive predictive value at $\tau^{\ast}$ & 0.03\,(0.03--0.04) & 0.21\,(0.21--0.21) \\
Expected errors per year at $\tau^{\ast}$ (FP / FN) & 7,682\,(6,912--8,592) / 131\,(120--140) & 3,372\,(3,310--3,432) / 14\,(12--15) \\
Expected paid claims per year at $\tau^{\ast}$ (FP / FN) & 7.7\,(6.9--8.6) / 2.6\,(2.4--2.8) & 3.4\,(3.3--3.4) / 0.3\,(0.2--0.3) \\ \midrule
\multicolumn{3}{c}{\textit{Efficiency Gaps}} \\ \midrule
Gap: EL vs CVaR & 0.151\,(0.136--0.168) & 0.342\,(0.331--0.356) \\
Gap: Accuracy vs CVaR & 0.159\,(0.143--0.177) & 0.591\,(0.581--0.601) \\ \midrule
\multicolumn{3}{c}{\textit{Risk Penalties}} \\ \midrule
Risk Penalty (EL) \% & 51.1\,(46.1--56.6) & 74.3\,(69.2--81.5) \\
Risk Penalty (Acc) \% & 77.4\,(70.4--85.9) & 296.9\,(274.7--320.2) \\ \midrule
\multicolumn{3}{c}{\textit{Liability Exposure at $\tau^{\ast}$ (\$M)}} \\ \midrule
CVaR of the annual loss & 22.0\,(20.8--23.4) & 10.0\,(9.9--10.1) \\
Expected annual loss & 10.9\,(10.0--11.9) & 8.9\,(8.8--9.1) \\
\bottomrule
\end{tabular}
\caption{\rev{Risk analysis of the two mammography settings. FP and FN stand for false positive and false negative.}}
\label{tab:comprehensive_risk_analysis}
\end{table}

\begin{revblock}
Accuracy-maximizing thresholds deviate from CVaR-optimal values by 0.159 in screening and 0.591 in diagnostic settings, generating tail risk penalties of 77\% and 297\% respectively. Expected loss optimization yields a 0.151 gap with a 51\% penalty in screening and a 0.342 gap with a 74\% penalty in the diagnostic setting. In operating terms, the risk-aware threshold flags 15.9\% of screening and 71.0\% of diagnostic examinations, far more than the expected-loss and accuracy thresholds (Table~\ref{tab:comprehensive_risk_analysis}; Section~\ref{sec:robustness_ec} compares the screening policy with US clinical practice). In both settings and in every draw the optimum is a smooth point of the CVaR objective, with no scenarios tied at $\text{VaR}_\beta$. The divergence follows from the cost structure. Work-up costs account for 84\% and 95\% of the expected loss, so the expected-loss threshold flags few examinations, and the accuracy threshold, which ignores costs, fewer still. The tail is driven by missed-cancer liability, whose per-error cost exceeds that of a false positive by factors of 82 (screening) and 44 (diagnostic) in the tail scenarios, against 10.0 and 9.9 on average. The risk-aware threshold therefore sits where the calibrated probability equals a tail-weighted false positive cost share of 1.2\% and 2.2\% (Theorem~\ref{thm:optimal_threshold}), an order of magnitude below the one-half of accuracy maximization.
\end{revblock}

\subsection{Insurance Value Decomposition}\label{sec:valueinsurance_case}

To quantify the economic value of algorithmic insurance under realistic market conditions, we implement the framework developed in Section \ref{sec:insurance_value} with parameters calibrated to the medical malpractice insurance market. \rev{We set the loading on expected payouts to $\mu = 0.15$ and the insurer's cost of tail capital to $r = 6\%$, the cost-of-capital rate of the Solvency II risk margin \citep{solvency2015}, of which our premium is a single-period form. The firm's cost of capital is its cost of equity, $\rho = 0.08$, between the 7.5\% of hospitals and the 8.9\% of healthcare technology firms \citep{damodaran2026}. In the absence of a capital rule for AI liability, we borrow the capital multipliers from bank regulation, with $\gamma = 4.0$ the upper value of the market-risk multiplier \citep{cfr217} and $\gamma_{\text{ins}} = 3.2$, which reflects the 20\% relief that the Basel framework allows for insurance \citep{basel2006international}. Coverage limits are \$1 million per occurrence and \$3 million in aggregate, the most common US physician malpractice limits \citep{ama2025}. We evaluate the insurer's payment, Equation~\eqref{eq:insurer_payment}, on the liability component with the expected numbers of paid claims (Section~\ref{sec:scenario_construction}).}

\begin{figure}[htbp]
    \centering
    \includegraphics[width=\linewidth]{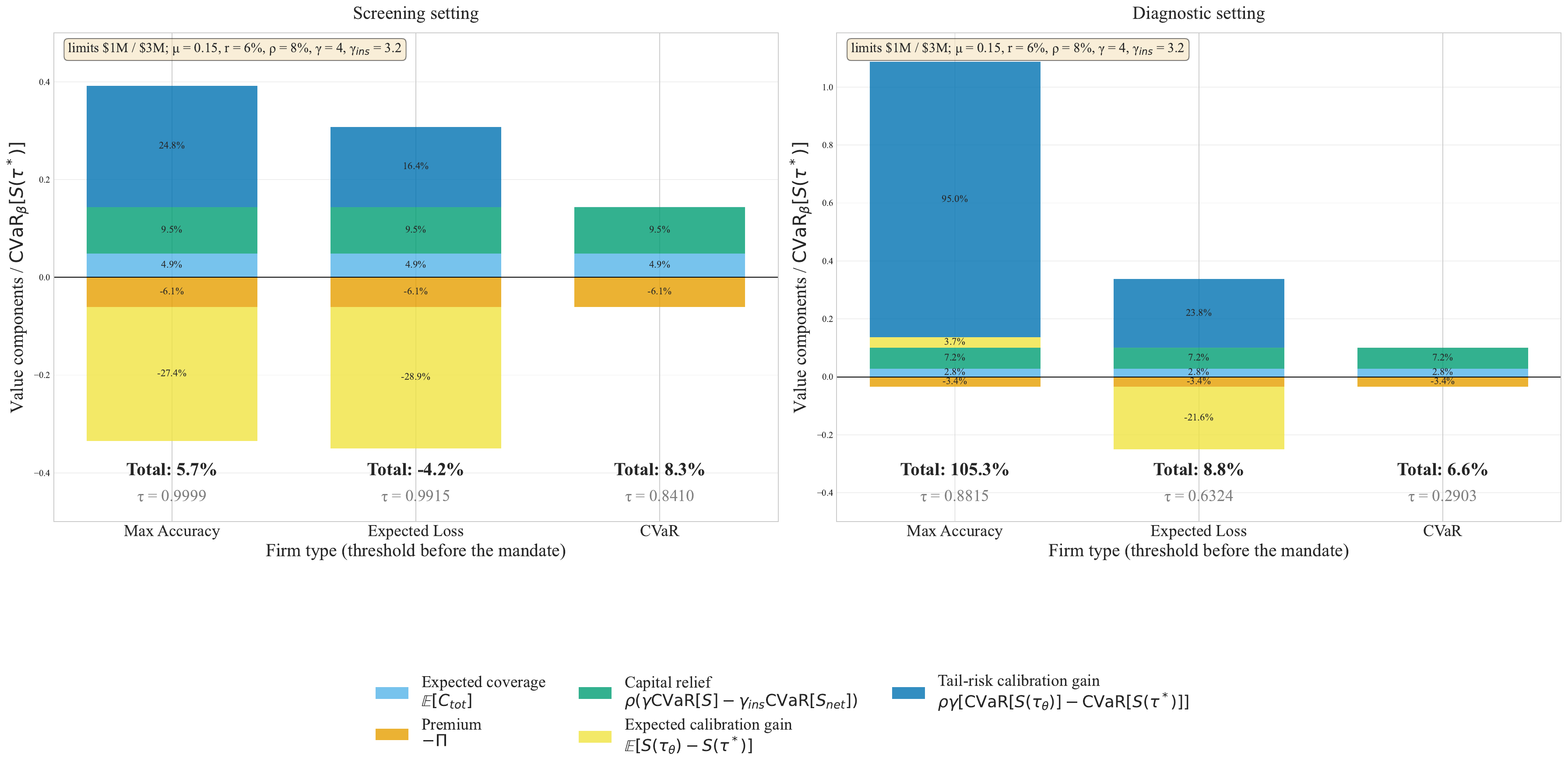}
    \caption{\rev{Decomposition of insurance value (Theorem~\ref{thm:insurance_value}) relative to $\text{CVaR}_\beta[S(\tau^{\ast})]$.}}
    \label{fig:valuecomponents_case}
\end{figure}

\rev{Figure \ref{fig:valuecomponents_case} presents the value decomposition from Theorem \ref{thm:insurance_value} across firm types, with every component divided by $\text{CVaR}_\beta[S(\tau^{\ast})]$, the tail risk under the mandated threshold. Base insurance value, comprising risk transfer and capital relief, equals 8.3\% of tail risk in screening and 6.6\% in the diagnostic setting. It is the same for every firm type because it is evaluated at $\tau^{\ast}$. Expected coverage (4.9\% and 2.8\%) falls short of the premium (6.1\% and 3.4\%), so risk transfer is negative ($-1.2\%$ and $-0.6\%$), in accordance with Equation~\eqref{eq:premium} (Section~\ref{sec:value_mechanisms}). Capital relief (9.5\% and 7.2\%) more than offsets this cost. The premium implies a loading of 24\% (screening) and 20\% (diagnostic) over expected payouts, of the order of the 22--32\% by which US medical professional liability premiums exceed expected losses and loss-adjustment expenses \citep{naic2026, burns2024}. Under the cost-of-capital premium, the condition of Corollary~\ref{cor:viability} is met, since $\rho\gamma = 0.32 > \max\{\mu, r\}$. The tail-to-average ratio of the insured loss is 7.4 in screening and 3.4 in the diagnostic setting (see also Section~\ref{sec:insurance_sensitivity_ec}).}

\begin{revblock}
The calibration value components reveal significant heterogeneity across firm types and settings. The tail risk reduction component is positive for every firm type and largest for the accuracy-maximizing firm in the diagnostic setting, 95.0\% of tail risk, which shows how severely traditional ML metrics misalign with liability minimization in high-stakes contexts. The expected loss reduction component is negative for both suboptimal firm types in screening and for the expected-loss firm in the diagnostic setting, because the mandated threshold flags many more examinations than theirs (Table~\ref{tab:comprehensive_risk_analysis}) and each additional false positive costs a work-up. Whether the mandate creates value therefore depends on how the firm weighs the two components, here through the capital cost $\rho\gamma$ per dollar of tail risk.

Insurance thus creates value through two channels that reach different firms. Firms far from the risk-optimal threshold gain through the mandate itself (the accuracy-maximizing firm in the diagnostic setting captures 105.3\%), and risk-aware firms through traditional insurance mechanisms (8.3\% and 6.6\%). Between these cases, the mandate can destroy value when the increase in expected loss outweighs the reduction in the tail charge. The expected-loss firm in screening has a total value of $-4.2\%$ and the accuracy-maximizing firm in screening only 5.7\%, because the operational cost of the additional recalls outweighs the capital cost of the tail risk they remove. These findings imply that insurance contracts should offer threshold flexibility, such as partial mandates or premium schedules indexed to the firm's threshold, for such partially aligned firms, while maintaining strict mandates where calibration benefits dominate. Section~\ref{sec:insurance_sensitivity_ec} reports the sensitivity of every component to $\pm 25\%$ changes in $\mu$, $\rho$, $\gamma$, and $\gamma_{\text{ins}}$, to the relief $1 - \gamma_{\text{ins}}/\gamma$, and to the insurer's cost of tail capital $r$.
\end{revblock}

\subsection{The Impact of Model Drift and Interpretability}\label{sec:extensions_case}

We examine how model performance degradation and interpretability investments could affect optimal insurance design in mammography AI deployment (see also Section \ref{sec:extensions}).

\begin{figure}[htbp]
    \centering
    \includegraphics[width=\linewidth]{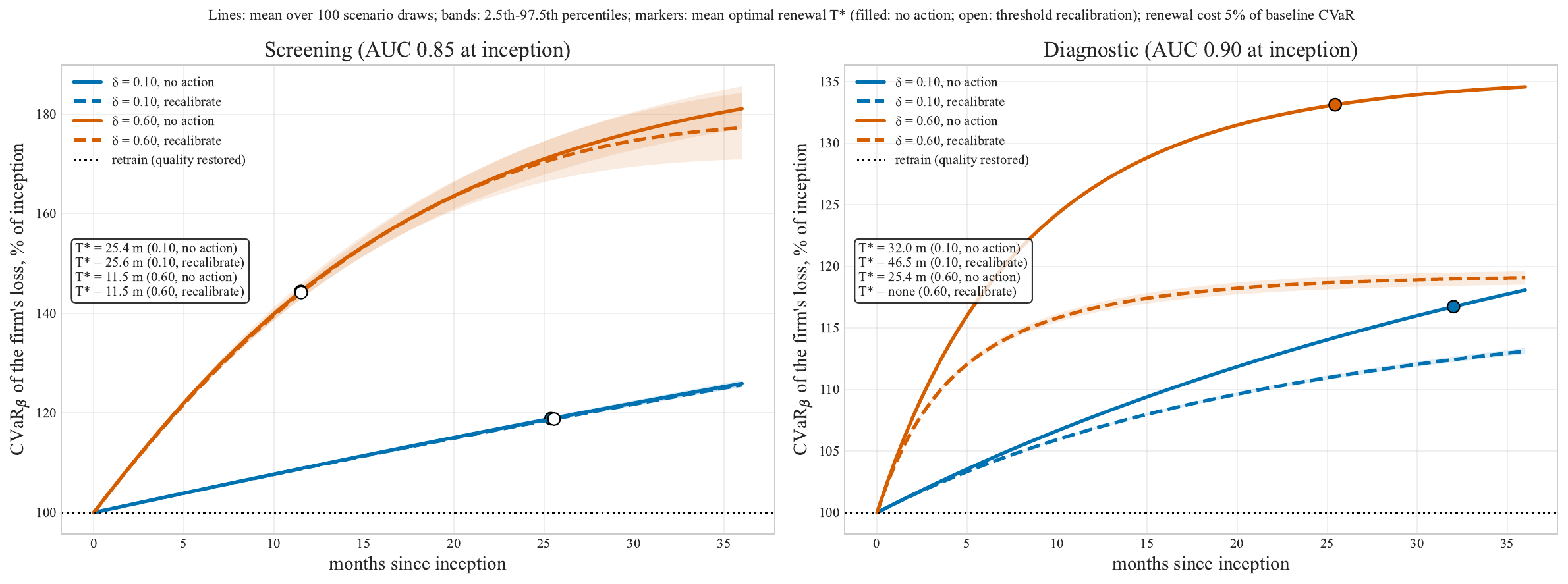}
    \caption{\rev{Model drift and renewal. Tail risk relative to inception without action and with threshold recalibration, with optimal renewal times (markers) and 2.5--97.5\% bands across draws.}}
    \label{fig:extensions_combined}
\end{figure}

\paragraph{Model Drift and Contract Duration.} \rev{Figure \ref{fig:extensions_combined} shows how the tail risk of the firm's loss evolves when the model's discrimination parameter decays exponentially at annual rates $\delta \in \{0.10, 0.60\}$, a decay we attribute to dataset shift \citep{singh2025experiences}. The parameter is the binormal separation $d$ in screening and the trigonometric parameter $k$ in the diagnostic setting (Section~\ref{sec:casesetup}). We compare keeping the mandated threshold (no action) with re-optimizing it on the degraded model (recalibration). We assume a renewal cost of 5\% of the baseline CVaR, reflecting administrative burdens of re-underwriting, system validation, and retraining, and compute the optimal renewal time from the condition of Equation~\eqref{eq:F0_condition_full} of Proposition~\ref{prop:optimal_duration_full}. Without recalibration the optimal renewal occurs after 25.4 months (screening) and 32.0 months (diagnostic) under slow drift, and after 11.5 and 25.4 months under fast drift. In the diagnostic setting, recalibrating the threshold within the contract slows the growth of tail risk and lengthens the optimal renewal to 46.5 months under slow drift. Under fast drift it removes the case for renewal altogether, since the recalibrated excess risk never accumulates enough to cover the renewal cost (the case $\mathcal{C}_a > H_\infty$ of Proposition~\ref{prop:optimal_duration_full}). In screening it leaves the renewal times almost unchanged. In neither setting does recalibration return the tail risk to its inception level (Corollary~\ref{cor:recalibration}), which only retraining achieves. Section~\ref{sec:renewal_oversight_ec} reports the renewal times for other renewal costs.} Our results suggest insurers should offer flexible contract durations calibrated to expected degradation rates. Moreover, applications where risk grows faster than the renewal cost threshold require continuous monitoring and adaptive coverage, while slower-degrading systems could maintain longer fixed terms. While drift analysis addresses temporal risk evolution, interpretability investments offer a complementary approach to risk mitigation through enhanced human oversight.

\paragraph{Interpretability as Risk Mitigation.}
\rev{Beyond temporal model drift, we examine how investments in human oversight affect the value of insurance in both settings. Applying the framework of Section \ref{subsec:interpretability}, we let $\zeta = 1$ denote a radiologist reviewing every examination alongside the AI system, so that $\zeta$ is the share of examinations reviewed and the cost scale $c_0 = c(1)$ is the annual professional fee for a radiologist interpretation of every examination\footnote{\rev{We assume \$34.93 per screening and \$45.29 per diagnostic mammogram under the 2025 Medicare physician fee schedule \citep{cms2025}, which gives \$1.75 million and \$0.27 million at the annual volumes.}}. We assume a cubic cost, $c(\zeta) = c_0\zeta^3$, since partial review can focus on the cases the system flags as uncertain while complete review requires additional reader capacity. We further set $g(\zeta) = \zeta^{1.5}$ and $\xi = 0.5$. The closed form of Equation~\eqref{eq:zetastar} then gives $\zeta^\ast = 0.33$ in screening and $0.46$ in the diagnostic setting. Section~\ref{sec:renewal_oversight_ec} and Figure~\ref{fig:interpretability_premium} report the resulting net benefits and show that the optimum is interior in almost every alternative cost and benefit specification.}

For insurers designing premium structures, optimal human-AI oversight levels require empirical calibration specific to each deployment context. The framework thus demonstrates how algorithmic insurance can guide AI safety through economic incentives on the  operational design of model integration, rather than prescriptive regulation.

\section{Discussion}\label{sec:discussion}

Our analysis establishes that operational design parameters of ML systems determine to a large extent the degree of AI liability exposure.  Algorithmic insurance can function as both a risk transfer instrument and a governance mechanism that directly improves system deployment decisions. We demonstrate that ML calibration decisions fundamentally shape liability exposure\rev{, with suboptimal threshold selection amplifying tail risk by a factor of up to four in our case study.} Our results reveal that algorithmic insurance creates an efficient coordination mechanism between liability and operations management, transforming insurance contracts from passive risk pooling arrangements into active operational governance instruments for deployment practices.

\subsection{Managerial Implications}\label{sec:ominsights}

We find that organizations deploying AI systems in high-stakes, liability-sensitive environments face fundamentally different operationalization challenges than traditional settings. We establish that optimizing system performance for tail risk becomes essential when algorithmic errors generate asymmetric liability consequences. \rev{Established calibration methods that maximize accuracy or minimize expected loss increase tail risk by 51\% to 297\% compared to risk-aware approaches in our case study.} 

Our analysis shows that algorithmic insurance creates value for ML providers and their clients through mechanisms that extend beyond traditional risk transfer by leveraging the programmable nature of AI systems ex post. Unlike conventional insurance that merely shifts financial exposure, algorithmic insurance can contractually specify and verify operational parameters such as the classification thresholds and the degree of human oversight. \rev{Operational mandates, such as threshold calibration, can generate substantial value for risk-unaware systems while traditional risk transfer and capital relief benefits remain available to all insured parties. For partially aligned systems, however, a strict mandate can cost more in operational expenses than it saves in tail risk, so contracts should allow threshold flexibility or premium schedules indexed to the threshold.}

Moreover, we demonstrate that optimal contract design depends on the interplay between algorithmic characteristics and operational environment. \rev{Contract duration should decrease with model drift rates as long as the excess risk keeps accumulating, and lengthen when thresholds are recalibrated within the contract. The value of coverage rests on the insurer holding tail capital more cheaply than the insured. Finally, investments in algorithmic transparency to facilitate human supervision need to balance implementation costs against premium reductions.} These findings provide guidance for structuring adaptive insurance products that evolve with system performance rather than treating algorithmic risk as static.

\rev{Our analysis provides concrete guidance for the primary stakeholders of the algorithmic insurance ecosystem. \textit{ML clients} gain external validation for operational improvements when insurance mandates risk-aware thresholds, and those already using optimal calibration can negotiate better service terms. \textit{ML providers} should evaluate coverage based on the gap between client practices and optimal thresholds, since a large gap creates value beyond risk transfer through mandatory improvements. \textit{Insurers} should exploit AI's verifiability through threshold-contingent pricing, with early opportunities where operational mandates generate immediate value. \textit{Regulators} can catalyze adoption through capital relief policy, since recognizing insured deployments as lower systemic risk creates market incentives for safer AI practices without prescriptive rules.}

\subsection{Limitations and Future Research Directions}\label{sec:limitations}

Several modeling choices define the scope and applicability of our framework. Guided by current algorithmic insurance offerings, we adopt the assumption of stationarity within contract periods to ensure analytical tractability. Our focus on binary classification reflects the prevalence and strategic importance of dichotomous decisions in high-stakes applications, encompassing the core liability challenges where algorithmic errors carry the highest financial consequences. Extensions beyond binary classification to regression models, reinforcement learning, and generative AI represent natural next steps for addressing risks from continuous outcomes, sequential decisions, and content generation. \rev{Such extensions require dedicated research. For predictive models with continuous outputs, \citet{sanchez2025aiinsurance} study parametric insurance contracts triggered when prediction errors exceed contractual bounds.} The framework's reliance on scenario-based cost estimates follows established risk management practice and leverages domain expertise through actuarial methods used successfully in analogous insurance markets. Future research could strengthen this approach through robust optimization techniques that account for ambiguity in scenario probabilities and cost parameters, providing performance guarantees under worst-case parameter realizations.

\rev{The assumption of threshold enforceability is a simplification but reflects the programmability that distinguishes algorithmic from human decisions. The audit provisions of Section~\ref{sec:insurance_motivation} make potential deviations detectable, yet imperfect enforcement, information asymmetry, and adverse selection would reduce the value we compute, and their treatment is a natural extension. Likewise, we acknowledge that the severity distributions in practice are estimated from censored and evolving claim data.} 

\rev{Our single-organization perspective enables precise characterization of individual risk optimization problems, though market-level dynamics and extensions to multi-agent problems present opportunities for future research. One such extension concerns algorithmic collusion, where pricing algorithms sustain supra-competitive outcomes without explicit agreement and expose the firms that deploy them to antitrust liability \citep{fish2024collusion, keppo2024pricing}. An insurer could price this liability and mandate constraints on the algorithm, such as its exploration rate, similarly to the threshold mandate for classifiers. Whether such coverage is desirable, given its effect on antitrust deterrence, remains an open question.}

\subsection{Conclusions}\label{sec:conclusions}

Our work aims to set the foundations in the novel area of algorithmic insurance. We propose, to the best of our knowledge, the first quantitative framework that allows ML providers and insurance companies to estimate the liability risk of ML models as a function of their predictive performance. Our findings bridge a critical gap between the operationalization of modern AI systems and the financial instruments necessary for their responsible deployment. We demonstrate that organizations can systematically manage algorithmic liability through operational design choices, transforming a significant barrier to AI adoption in high-stakes applications into a manageable and insurable risk.  Our work can serve as the basis of a new research area and guide an expanding industry market that will facilitate responsible AI deployment in liability-sensitive applications.
{\small
\bibliographystyle{informs2014} 
\bibliography{ref}}

\clearpage
\newpage
\ECSwitch
\ECHead{Electronic Companion}

\noindent This Electronic Companion provides supplementary materials that support the theoretical and empirical analyses presented in the main manuscript. Section~\ref{sec:suptheory} establishes the supporting results for our theoretical framework. Section~\ref{sec:proofs} contains complete proofs for all theorems and propositions stated in the main text\rev{ and for the corollaries of Section~\ref{sec:suptheory}}. \rev{Section~\ref{app:trigonometric_validation} assesses how well the trigonometric model fits real classifiers and delimits the range in which it should be used.} Section~\ref{sec:scenariolosses} provides implementation details for our case study. Sections \ref{sec:suppfigures}--\ref{tab:notationsummary} include supplementary figures and tables. Throughout, we maintain consistent notation with the main manuscript and provide additional technical details that support our primary theoretical contributions.

\FloatBarrier
\section{Supplementary Structural Results}\label{sec:suptheory}

In this section, we present the rigorous properties required to support the analytical framework and develop additional benchmarks and extensions referenced in the main text. Section~\ref{proofstrigonoproperties} verifies the essential properties of the trigonometric model. Section~\ref{sec:cvarECresults} establishes \rev{the strict convexity and differentiability} properties of the CVaR objective\rev{, presents an algorithm for the optimal threshold, shows that the threshold is continuous in the costs, and lists the results that hold for any calibration map}. Section \ref{sec:acccomparison} extends the comparisons with the accuracy benchmark. Section~\ref{sec:explossmaximizer} analyzes expected loss minimization as an intermediate benchmark. \rev{Section~\ref{sec:theoryinsurance} states the viability condition for full coverage.} Section~\ref{sec:theorydrift} examines tail risk growth under model degradation\rev{, and Section~\ref{sec:binormal} gives the counterparts of the closed forms for the binormal model}.

\FloatBarrier
\subsection{ML Model Properties}\label{proofstrigonoproperties}

\begin{lemma}[Boundary Conditions]\label{lem:boundary}
Let $\mathbb{P}(Y = 1 \mid \overline{\gamma})$ be defined as in Equation~\eqref{eq:trig-model}. Then the model satisfies: $\mathbb{P}(Y = 1 \mid \overline{\gamma} = 0) = 0$, $\mathbb{P}(Y = 1 \mid \overline{\gamma} = \alpha) = 1 - \alpha$, and $\mathbb{P}(Y = 1 \mid \overline{\gamma} = 1) = 1$.
\end{lemma}

\begin{proof}
\textnormal{\noindent We verify each boundary condition:} \textnormal{At} $\overline{\gamma} = 0$: $\mathbb{P}(Y=1|0) = \frac{1-\alpha}{\sin k}[\sin(-k) + \sin k] = 0$.
At $\overline{\gamma} = \alpha$: Case 1 gives $\frac{1-\alpha}{\sin k}\sin k = 1-\alpha$; Case 2 gives $(1-\alpha) + \frac{\alpha}{k}[\arcsin(-\sin k) + k] = (1-\alpha) + \frac{\alpha}{k}[-k + k] = 1-\alpha$.
At $\overline{\gamma} = 1$: $\mathbb{P}(Y=1|1) = (1-\alpha) + \alpha = 1$. These conditions ensure that the model output is calibrated at the extrema and remains continuous at the population proportion $\alpha$.
\end{proof}

\begin{lemma}[Probability Conservation]\label{lem:conservation}
Let $\mathbb{P}(Y = 1 \mid \overline{\gamma})$ be defined as in Equation~\eqref{eq:trig-model}. Then the model satisfies the following integral identity: $\int_0^1 \mathbb{P}(Y = 1 \mid \overline{\gamma}) \, d\overline{\gamma} = 1 - \alpha.$
\end{lemma}

\begin{proof}
\textnormal{We split the integral:}
$\int_0^1 \mathbb{P}(Y=1|\overline{\gamma}) \, d\overline{\gamma} = \int_0^\alpha \mathbb{P}(Y=1|\overline{\gamma}) \, d\overline{\gamma} + \int_\alpha^1 \mathbb{P}(Y=1|\overline{\gamma}) \, d\overline{\gamma}$. Similarly to Lemma~\ref{lem:fn_case}, we define the following area functions over the probability function and derive an analytical form from calculus principles (see Equation \ref{eq:S1}):
\begin{align*}
S_1 = \int_0^\alpha \mathbb{P}(Y=1|\overline{\gamma}) \, d\overline{\gamma} =  \frac{(1-\alpha)\alpha}{k\sin k}(k\sin k + \cos k - 1) \quad \mathrm{ and }  \quad
S_2 = \int_\alpha^1 \mathbb{P}(Y=1|\overline{\gamma}) \, d\overline{\gamma}.
\end{align*}

For the second integral $S_2 = \int_\alpha^1 \mathbb{P}(Y=1|\overline{\gamma}) \, d\overline{\gamma}$, we use the second case of the model since $\overline{\gamma} > \alpha$:

\begin{align*}
S_2 &= \int_\alpha^1 \left[(1-\alpha) + \frac{\alpha}{k}\left[\arcsin\left(\frac{(\overline{\gamma}-1)\sin k}{1-\alpha}\right) + k\right]\right] d\overline{\gamma}
\end{align*}

For the arcsin integral, we use substitution $u = \frac{(\overline{\gamma}-1)\sin k}{1-\alpha}$, so $\overline{\gamma} = 1 + \frac{u(1-\alpha)}{\sin k}$ and $d\overline{\gamma} = \frac{1-\alpha}{\sin k} du$. We observe that when $\overline{\gamma} = \alpha$: $u = \frac{(\alpha-1)\sin k}{1-\alpha} = -\sin k$ and when $\overline{\gamma} = 1$: $u = 0$. Thus, we derive: $\int_\alpha^1 \arcsin\left(\frac{(\overline{\gamma}-1)\sin k}{1-\alpha}\right) d\overline{\gamma} = \frac{1-\alpha}{\sin k} \int_{-\sin k}^0 \arcsin(u) \, du$. Using integration by parts with $v = \arcsin(u)$ and $dw = du$: $
\int \arcsin(u) \, du = u \arcsin(u) + \sqrt{1-u^2} + C$. Then, we evaluate: $\int_{-\sin k}^0 \arcsin(u) \, du = \left[u \arcsin(u) + \sqrt{1-u^2}\right]_{-\sin k}^0 = 1 - k\sin k - \cos k$.
Finally, we substitute back:
\begin{align*}
S_2 &= (1-\alpha)^2 + \frac{\alpha}{k}\left[\frac{1-\alpha}{\sin k}(1 - k\sin k - \cos k) + k(1-\alpha)\right] = (1-\alpha)\left[(1-\alpha) + \alpha \cdot \frac{1-\cos k}{k\sin k}\right]
\end{align*}

By summing $S_1 + S_2 = 1-\alpha.$ This condition ensures that the total expected proportion of class 1 instances under the stylized model \rev{equals the population share $1-\alpha$ of class 1}.
\end{proof}

\begin{lemma}[Monotonicity of Error Rates]\label{lem:monotonicity}
Under the trigonometric model defined in equation~\eqref{eq:trig-model}, for every $k \in (0, \pi/2]$\rev{, the map $\overline{\gamma} \mapsto \mathbb{P}(Y=1 \mid \overline{\gamma})$ is continuous and strictly increasing on $[0,1]$, from $0$ to $1$, with derivative $\frac{(1-\alpha)k}{\alpha\sin k}\cos\bigl(\frac{k\overline{\gamma}}{\alpha} - k\bigr)$ on $(0,\alpha)$ and $\frac{\alpha\sin k}{k(1-\alpha)}\bigl[1-\bigl(\frac{(\overline{\gamma}-1)\sin k}{1-\alpha}\bigr)^2\bigr]^{-1/2}$ on $(\alpha,1)$, both positive (the first vanishes only at $\overline{\gamma} = 0$ when $k = \pi/2$). Moreover}:
\begin{enumerate}
    \item The false negative rate $\mathbb{P}_{\mathrm{FN}}(\tau)$ is \rev{strictly increasing in $\tau$};
    \item The false positive rate $\mathbb{P}_{\mathrm{FP}}(\tau)$ is \rev{strictly decreasing in $\tau$}.
\end{enumerate}
\end{lemma}

\begin{proof}
\rev{\textnormal{Continuity} at $\overline{\gamma} = \alpha$ and the boundary values follow from Lemma~\ref{lem:boundary}. Differentiating Equation~\eqref{eq:trig-model} on each branch gives the two derivatives, which are positive because $\frac{k\overline{\gamma}}{\alpha} - k \in (-k, 0) \subset (-\pi/2, 0)$ on the first branch and the square root is positive on the second. A continuous function with positive derivative off a finite set is strictly increasing.} We demonstrate the monotonicity properties of the error rates by analyzing the respective derivatives. First, we show that  $\mathbb{P}_{\mathrm{FN}}(\tau)$ is strictly increasing in $\tau$. By definition, $\mathbb{P}_{\mathrm{FN}}(\tau) = \int_0^\tau \mathbb{P}(Y=1|\overline{\gamma}) \, d\overline{\gamma}$ and by the Fundamental Theorem of Calculus it follows that
$\frac{d\mathbb{P}_{\mathrm{FN}}(\tau)}{d\tau} = \mathbb{P}(Y=1|\tau)$.

Since $\mathbb{P}(Y=1|\tau) \in [0,1]$ for all $\tau \in [0,1]$, and from the model definition, $\mathbb{P}(Y=1|\tau) \geq 0$ for all $\tau \geq 0$, we have $ \frac{d\mathbb{P}_{\mathrm{FN}}(\tau)}{d\tau} \geq 0 \quad \text{for all } \tau \in [0,1]$
with strict inequality $\frac{d\mathbb{P}_{\mathrm{FN}}(\tau)}{d\tau} > 0$ for $\tau > 0$ \rev{(by the monotonicity of the map)}. Moreover, at $\tau = 0$, we have $\mathbb{P}(Y=1|0) = 0$ from Lemma~\ref{lem:boundary}, so the derivative is non-negative at the boundary. Since the model is continuous and strictly positive for $\tau > 0$, $\mathbb{P}_{\mathrm{FN}}(\tau)$ is strictly increasing on $[0,1]$.

Next, we demonstrate that $\mathbb{P}_{\mathrm{FP}}(\tau)$ is strictly decreasing in $\tau$. From the proof of Lemma~\ref{lem:fp_case}, we established that:
$\mathbb{P}_{\mathrm{FP}}(\tau) = \alpha - \tau + \mathbb{P}_{\mathrm{FN}}(\tau)$. Taking the derivative $\frac{d\mathbb{P}_{\mathrm{FP}}(\tau)}{d\tau} = -1 + \frac{d\mathbb{P}_{\mathrm{FN}}(\tau)}{d\tau} = -1 + \mathbb{P}(Y=1|\tau)$. Since $\mathbb{P}(Y=1|\tau) \in [0,1]$ for all $\tau \in [0,1]$, we have $\frac{d\mathbb{P}_{\mathrm{FP}}(\tau)}{d\tau} = \mathbb{P}(Y=1|\tau) - 1 \leq 0$.

To show strict inequality, note that $\mathbb{P}(Y=1|\tau) = 1$ only when $\tau = 1$ (from Lemma~\ref{lem:boundary}). For all $\tau \in [0,1)$, we have $\mathbb{P}(Y=1|\tau) < 1$, which implies $\frac{d\mathbb{P}_{\mathrm{FP}}(\tau)}{d\tau} < 0 \quad \text{for all } \tau \in [0,1)$. At $\tau = 1$, we have $\mathbb{P}_{\mathrm{FP}}(1) = 0$ (no false positives when everything is classified as negative), which is the minimum value. Therefore, $\mathbb{P}_{\mathrm{FP}}(\tau)$ is strictly decreasing on $[0,1]$.
\end{proof}

\begin{lemma}[False Negative Probability]\label{lem:fn_case}
The false negative probability $\mathbb{P}_{\mathrm{FN}}(\tau)$ is given by:
\begin{equation}
\mathbb{P}_{\mathrm{FN}}(\tau) =
\begin{cases}
(1-\alpha)\tau + \frac{(1-\alpha)\alpha}{k\sin k} \left[\cos k - \cos\left(\frac{k\tau}{\alpha} - k\right)\right], & \text{if } \tau \leq \alpha \\[0.5em]
(\tau - \alpha)(1 - \alpha + f(\tau)) \\ + \frac{1 - \alpha}{k \sin k} \left[(\alpha - f(\tau))k \sin k + \alpha \cos\left(\frac{k f(\tau)}{\alpha} - k\right) - \alpha \right], & \text{if } \tau > \alpha,
\end{cases}
\end{equation}
\noindent where $f(\tau) := \frac{\alpha}{k} \left[ \arcsin\left( \frac{(\tau - 1)\sin k}{1 - \alpha} \right) + k \right]$. 
\end{lemma}

\begin{proof}
\textnormal{For} $\tau \leq \alpha$, false negatives occur when $\overline{\gamma} < \tau$ and $Y = 1$. Since the entire interval $[0,\tau]$ lies within the interval $[0,\alpha]$:
\begin{align*}
\mathbb{P}_{\mathrm{FN}}(\tau) &= \int_0^\tau \mathbb{P}(Y=1|\overline{\gamma}) \, d\overline{\gamma} = \int_0^\tau \frac{1-\alpha}{\sin k}\left[\sin\left(\frac{k\overline{\gamma}}{\alpha} - k\right) + \sin k\right] d\overline{\gamma}
\end{align*}

We factor out constants and split the integral: $\frac{1-\alpha}{\sin k}\left[\int_0^\tau \sin\left(\frac{k\overline{\gamma}}{\alpha} - k\right) d\overline{\gamma} + \sin k \int_0^\tau d\overline{\gamma}\right].$ For the trigonometric integral, use substitution $u = \frac{k\overline{\gamma}}{\alpha} - k$, so $d\overline{\gamma} = \frac{\alpha}{k} du$: (i) when $\overline{\gamma} = 0$: $u = -k$; (ii) when $\overline{\gamma} = \tau$: $u = \frac{k\tau}{\alpha} - k$. 
\begin{align*}
\int_0^\tau \sin\left(\frac{k\overline{\gamma}}{\alpha} - k\right) d\overline{\gamma} &= \frac{\alpha}{k} \int_{-k}^{\frac{k\tau}{\alpha}-k} \sin u \, du = \frac{\alpha}{k}\left[-\cos u\right]_{-k}^{\frac{k\tau}{\alpha}-k} = \frac{\alpha}{k}\left[\cos k - \cos\left(\frac{k\tau}{\alpha} - k\right)\right]
\end{align*}
The second integral yields $\sin k \cdot \tau$. Combining:
\begin{align*}
\mathbb{P}_{\mathrm{FN}}(\tau) &= \frac{1-\alpha}{\sin k}\left[\frac{\alpha}{k}\left[\cos k - \cos\left(\frac{k\tau}{\alpha} - k\right)\right] + \sin k \cdot \tau\right] \\
&= (1-\alpha)\tau + \frac{(1-\alpha)\alpha}{k\sin k}\left[\cos k - \cos\left(\frac{k\tau}{\alpha} - k\right)\right]
\end{align*}

\textnormal{For} $\tau > \alpha$, the integration domain spans both cases of the model. We split the integral:
\begin{align*}
\mathbb{P}_{\mathrm{FN}}(\tau)&= \int_0^\tau \mathbb{P}(Y=1|\overline{\gamma}) \, d\overline{\gamma} = \underbrace{\int_0^\alpha \mathbb{P}(Y=1|\overline{\gamma}) \, d\overline{\gamma}}_{S_1} + \underbrace{\int_\alpha^\tau \mathbb{P}(Y=1|\overline{\gamma}) \, d\overline{\gamma}}_{I_1}
\end{align*}
The integration domain $[0,\alpha]$ lies entirely within the first case of the model. Similarly to the case of $\tau\leq \alpha$, we use substitution for the first integral $u = \frac{k\overline{\gamma}}{\alpha} - k$, so $d\overline{\gamma} = \frac{\alpha}{k} du$. Thus, when $\overline{\gamma} = 0$: $u = -k$ and when $\overline{\gamma} = \alpha$: $u = 0$
\begin{align}\label{eq:S1}
S_1 &= \int_0^\alpha \frac{1-\alpha}{\sin k}\left[\sin\left(\frac{k}{\alpha} \cdot \overline{\gamma} - k\right) + \sin k\right] d\overline{\gamma} = \frac{(1-\alpha)\alpha}{k \sin k}(k \sin k + \cos k - 1)
\end{align}

For $\tau>\alpha$, we define
\begin{align*}
f(\tau)\;=\;
\frac{\alpha}{k}\Bigl[
  \arcsin\!\Bigl(\tfrac{(\tau-1)\sin k}{1-\alpha}\Bigr)+k
\Bigr],
\qquad 0<f(\tau)\rev{\leq}\alpha.
\end{align*}
Based on this definition, one can verify that 
\begin{align}\label{eq:A}
P\!\bigl(Y=1\mid\overline\gamma=f(\tau)\bigr)=
\frac{1-\alpha}{\sin k}
  \Bigl[
    \sin\!\Bigl(\tfrac{k}{\alpha}f(\tau)-k\Bigr)+\sin k
  \Bigr]
=\tau-\alpha.    
\end{align}

To analyze $I_1 = \int_\alpha^\tau \mathbb{P}(Y=1|\overline{\gamma})\rev{\,d\overline{\gamma}}$, we set
\[
G(\tau)\;:=\;
(\tau-\alpha)\,\mathbb{P}(Y=1\mid\tau)
\;-\;\int_{\alpha}^{\tau}\mathbb{P}(Y=1\mid x)\,dx
\;-\;\int_{0}^{f(\tau)}\mathbb{P}(Y=1\mid x)\,dx,
\quad\alpha\le \tau\le 1.
\]

First, observe that at $\tau=\alpha$ each integrand’s upper limit
equals its lower limit, so $G(\alpha)=0$. Next, we differentiate, using Equation~\eqref{eq:trig-model} and the fact that
$f(\tau)=\mathbb{P}(Y=1\mid\tau)- (1-\alpha)$:
\[
\begin{aligned}
G'(\tau)
&=\mathbb{P}(Y=1\mid\tau)+(\tau-\alpha)\,\partial_\tau\mathbb{P}(Y=1\mid\tau)
      -\mathbb{P}(Y=1\mid\tau)                      -\mathbb{P}\bigl(Y=1\mid\overline\gamma=f(\tau)\bigr)\,
        f'(\tau)                                     \\[2mm]
&=(\tau-\alpha)\,\partial_\tau\mathbb{P}(Y=1\mid\tau)-\mathbb{P}\bigl(Y=1\mid\overline\gamma=f(\tau)\bigr)\,f'(\tau).
\end{aligned}
\]
\rev{By Equation~\eqref{eq:A}, $\mathbb{P}(Y=1\mid\overline\gamma=f(\tau))=\tau-\alpha$, and on $(\alpha,1]$ we have $\mathbb{P}(Y=1\mid\tau)=1-\alpha+f(\tau)$, so $\partial_\tau\mathbb{P}(Y=1\mid\tau)=f'(\tau)$; hence $G'(\tau)=(\tau-\alpha)f'(\tau)-(\tau-\alpha)f'(\tau)=0$.} Thus, we can conclude that since $G(\alpha)=0$ and $G'(\tau)=0$, we have
$G(\tau)\equiv0$ for all $\tau\!>\!\alpha$, i.e.
\[
(\tau-\alpha)\,\mathbb{P}(Y=1\mid\tau)
=\int_{\alpha}^{\tau}\mathbb{P}(Y=1\mid x)\,dx
+\int_{0}^{f(\tau)}\mathbb{P}(Y=1\mid x)\,dx.
\]
This finding can also be verified using a geometric argument, visualized in Figure~\ref{fig:trigonometricregions}.
\begin{figure}[h]
  \centering
\includegraphics[width=0.6\linewidth]{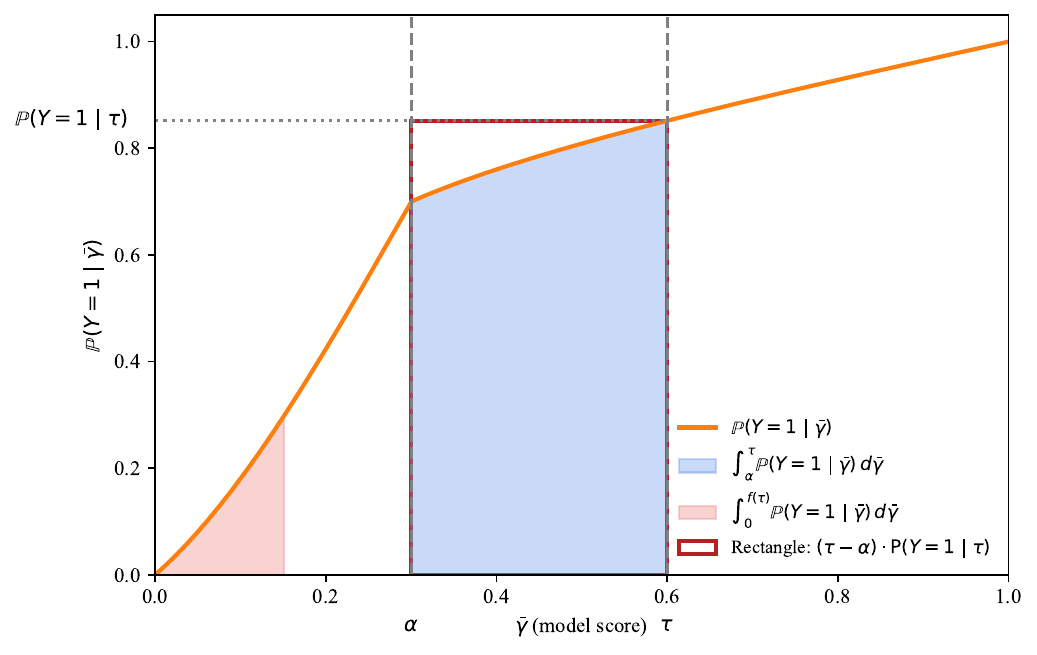}
  \caption{\rev{Geometric verification of the identity in Lemma~\ref{lem:fn_case}. The blue region and a region of the same area as the pink one make up the rectangle of height $P(Y=1\mid\tau)$ and width $\tau-\alpha$.}}
  \label{fig:trigonometricregions}
\end{figure}
\rev{By the decomposition of this rectangle shown in Figure~\ref{fig:trigonometricregions},}
\begin{align}\label{eq:B}
(\tau-\alpha)\,P(Y=1\mid\tau)
=
\underbrace{\int_{\alpha}^{\tau}P(Y=1\mid\overline\gamma)\,d\overline\gamma}_{I_{1}}
+
\underbrace{\int_{0}^{f(\tau)}P(Y=1\mid\overline\gamma)\,d\overline\gamma}_{I_{2}}.
\end{align}
Hence $
I_1=(\tau-\alpha)\,P(Y=1\mid\tau)-I_2.$ The mirror integral $I_{2}$ is on the left branch, so using the same
substitution that produced \(S_{1}\) we obtain
\[
I_{2}=
(1-\alpha)f(\tau)
+\frac{(1-\alpha)\alpha}{k\sin k}
  \Bigl[
    \cos k-\cos\!\Bigl(\tfrac{k f(\tau)}{\alpha}-k\Bigr)
  \Bigr].
\]

Collecting terms and using
\(P(Y=1\mid\tau)=1-\alpha+f(\tau)\) (the right-branch formula) gives
{\color{black}\begin{align}\label{eq:C}
I_{1}
&=(\tau-\alpha)\bigl(1-\alpha+f(\tau)\bigr)-(1-\alpha)f(\tau)
-\frac{(1-\alpha)\alpha}{k\sin k}
  \Bigl[\cos k-\cos\!\Bigl(\tfrac{k f(\tau)}{\alpha}-k\Bigr)\Bigr].
\end{align}}

Adding \(S_{1}\) from Equation~\eqref{eq:S1} to $I_{1}$ in Equation~\eqref{eq:C} and simplifying yields
\begin{align*}
\mathbb{P}_{\mathrm{FN}}(\tau)=
(\tau-\alpha)\bigl(1-\alpha+f(\tau)\bigr)
+\frac{1-\alpha}{k\sin k}
  \Bigl[(\alpha-f(\tau))k\sin k
        +\alpha\cos\!\Bigl(\tfrac{k f(\tau)}{\alpha}-k\Bigr)-\alpha\Bigr],
\qquad \tau>\alpha.
\end{align*}
\end{proof}

\begin{lemma}[False Positive Probability]\label{lem:fp_case}
The false positive probability $\mathbb{P}_{\mathrm{FP}}(\tau)$ is given by:
\begin{equation}
\mathbb{P}_{\mathrm{FP}}(\tau) =
\begin{cases}
\alpha(1 - \tau) + \frac{(1 - \alpha)\alpha}{k \sin k} \left[\cos k - \cos\left(\frac{k\tau}{\alpha} - k\right)\right], & \text{if } \tau \leq \alpha \\[0.5em]
(\tau - \alpha)(f(\tau) - \alpha) \\ + \frac{1 - \alpha}{k \sin k} \left[(\alpha - f(\tau))k \sin k + \alpha \cos\left(\frac{k f(\tau)}{\alpha} - k\right) - \alpha \right], & \text{if } \tau > \alpha.
\end{cases}
\end{equation}
\end{lemma}

\begin{proof}
\textnormal{Since} false positives occur when $\overline{\gamma} \geq \tau$ and $Y = 0$, we have 
$\mathbb{P}_{\mathrm{FP}}(\tau) = \int_\tau^1 [1 - \mathbb{P}(Y=1|\overline{\gamma})] d\overline{\gamma}$.
Using the conservation property $\int_0^1 \mathbb{P}(Y=1|\overline{\gamma}) d\overline{\gamma} = 1-\alpha$, 
this yields $\mathbb{P}_{\mathrm{FP}}(\tau) = \alpha - \tau + \mathbb{P}_{\mathrm{FN}}(\tau)$.
Substituting the expressions from Lemma \ref{lem:fn_case} for each case gives the stated results.
\end{proof}

\FloatBarrier
\subsection{CVaR Properties}\label{sec:cvarECresults}

\begin{lemma}[Differentiability of Scenario Loss]\label{lem:differentiability}
Under the trigonometric model defined in Equation~\eqref{eq:trig-model}, for each scenario~$j$, the loss function $S_j(\tau)$ defined in Equation \eqref{eq:loss_scenario}
is continuously differentiable on $[0,1]$, with first derivative given explicitly by
\[
\frac{dS_j(\tau)}{d\tau} 
= \sum_{i=1}^{N}\left[\frac{d\mathbb{P}_{\mathrm{FP}}(\tau)}{d\tau} K_{ji}+\frac{d\mathbb{P}_{\mathrm{FN}}(\tau)}{d\tau} L_{ji}\right].
\]
Furthermore, $S_j(\tau)$ is twice continuously differentiable on $(0,\alpha)\cup(\alpha,1)$, \rev{and its second derivative is in general discontinuous at $\tau=\alpha$ (the one-sided limits coincide only if $(1-\alpha)^2k^2\cos k = \alpha^2\sin^2 k$)}.
\end{lemma}

\begin{proof}
\textnormal{The} loss function for scenario $j$ can be expressed as:
\begin{equation*}
S_j(\tau) = \mathbb{P}_{\mathrm{FP}}(\tau) K_j + \mathbb{P}_{\mathrm{FN}}(\tau) L_j,
\end{equation*}
where $K_j = \sum_{i=1}^{N} K_{ji}$ and $L_j = \sum_{i=1}^{N} L_{ji}$ are constants for each scenario.

From Lemmas~\ref{lem:fn_case} and \ref{lem:fp_case}, the error probabilities $\mathbb{P}_{\mathrm{FN}}(\tau)$ and $\mathbb{P}_{\mathrm{FP}}(\tau)$ have explicit forms involving compositions of elementary functions. For $\tau \in (0, \alpha)$, both probabilities are expressed through linear terms and cosine functions of linear transformations of $\tau$. For $\tau \in (\alpha, 1)$, they involve the auxiliary function $f(\tau) = \frac{\alpha}{k}[\arcsin(\frac{(\tau-1)\sin k}{1-\alpha}) + k]$, where the argument of the arcsine remains in its domain $(-1, 0]$ for all valid $\tau$ values.

The differentiability of $S_j(\tau)$ on the open intervals $(0, \alpha)$ and $(\alpha, 1)$ follows immediately from the differentiability of these elementary functions and their compositions. At the boundary point $\tau = \alpha$, continuity is established in Lemma~\ref{lem:boundary}. Moreover, from the proof of Lemma~\ref{lem:monotonicity}, we have $\frac{d\mathbb{P}_{\mathrm{FN}}(\tau)}{d\tau} = \mathbb{P}(Y=1|\tau)$ and $\frac{d\mathbb{P}_{\mathrm{FP}}(\tau)}{d\tau} = \mathbb{P}(Y=1|\tau) - 1$, both of which are continuous at $\tau = \alpha$ since $\mathbb{P}(Y=1|\overline{\gamma})$ is continuous. Therefore, $S_j(\tau)$ is continuously differentiable on $[0,1]$ with derivative:
\begin{equation*}
\frac{dS_j(\tau)}{d\tau} = K_j \frac{d\mathbb{P}_{\mathrm{FP}}(\tau)}{d\tau} + L_j \frac{d\mathbb{P}_{\mathrm{FN}}(\tau)}{d\tau} = \sum_{i=1}^{N} \left[K_{ji} \frac{d\mathbb{P}_{\mathrm{FP}}(\tau)}{d\tau} + L_{ji} \frac{d\mathbb{P}_{\mathrm{FN}}(\tau)}{d\tau}\right]
\end{equation*}

On the open intervals $(0,\alpha)$ and $(\alpha,1)$, $\mathbb{P}_{\mathrm{FN}}(\tau)$ and $\mathbb{P}_{\mathrm{FP}}(\tau)$ are infinitely differentiable as compositions of smooth elementary functions. However, the second derivatives \rev{may be discontinuous} at $\tau = \alpha$. \rev{For $k < \pi/2$} the left and right limits of $\frac{d^2S_j(\tau)}{d\tau^2}$ at $\tau = \alpha$ are
\begin{align*}
\lim_{\tau \to \alpha^-} \frac{d^2S_j(\tau)}{d\tau^2} &= (K_j + L_j) \cdot \frac{(1-\alpha)k}{\alpha\sin k} \qquad\text{and}\qquad
\lim_{\tau \to \alpha^+} \frac{d^2S_j(\tau)}{d\tau^2} = (K_j + L_j) \cdot \frac{\alpha\sin k}{k(1-\alpha)\cos k},
\end{align*}
\rev{which coincide only if $(1-\alpha)^2k^2\cos k = \alpha^2\sin^2 k$. At $k = \pi/2$ the right limit is infinite.} Thus, $S_j(\tau)$ is twice continuously differentiable on $(0,\alpha) \cup (\alpha,1)$ but\rev{, in general,} not on the entire interval $[0,1]$.

\end{proof}

\begin{revblock}
Throughout the electronic companion we write $\Psi(\tau) := \text{CVaR}_\beta(S(\tau))$ and $\Psi_q(\tau) := \sum_j q_j S_j(\tau)$ for $q \in \mathcal{Q}_\beta$, so that $\Psi = \max_{q \in \mathcal{Q}_\beta} \Psi_q$ by Equation~\eqref{eq:envelope}.

\begin{lemma}[Strict Convexity of Scenario Losses and CVaR]\label{lem:convexity}
Under the trigonometric model with parameters $\alpha \in (0,1)$ and $k \in (0, \pi/2]$, each scenario loss $S_j(\tau) = K_j\mathbb{P}_{\mathrm{FP}}(\tau) + L_j\mathbb{P}_{\mathrm{FN}}(\tau)$ is strictly convex on $[0,1]$, with $S_j'(\tau) = (K_j + L_j)\mathbb{P}(Y=1|\tau) - K_j$, $S_j'(0) = -K_j < 0$ and $S_j'(1) = L_j > 0$. For every $q \in \mathcal{Q}_\beta$ the function $\tau \mapsto \sum_j q_j S_j(\tau)$ is strictly convex, and so is $\tau \mapsto \text{CVaR}_\beta(S(\tau))$. Consequently $\text{CVaR}_\beta(S(\tau))$ has a unique minimizer $\tau^{\ast} \in (0,1)$.
\end{lemma}
\end{revblock}

\begin{proof}
\textnormal{We} prove each $S_j(\tau) = \mathbb{P}_{\mathrm{FP}}(\tau) K_j + \mathbb{P}_{\mathrm{FN}}(\tau) L_j$ is convex, then extend to CVaR. From Lemma~\ref{lem:differentiability}, each $S_j(\tau) = \mathbb{P}_{\mathrm{FP}}(\tau) K_j + \mathbb{P}_{\mathrm{FN}}(\tau) L_j$ is continuously differentiable on $[0,1]$ and twice continuously differentiable on $(0,\alpha) \cup (\alpha,1)$.

\textit{Convexity on the intervals $(0,\alpha)$ and $(\alpha,1)$:}
From Lemma~\ref{lem:monotonicity}, we have $\frac{d\mathbb{P}_{\mathrm{FN}}(\tau)}{d\tau} = \mathbb{P}(Y=1|\tau)$ and $\frac{d\mathbb{P}_{\mathrm{FP}}(\tau)}{d\tau} = \mathbb{P}(Y=1|\tau) - 1$, yielding:
\begin{equation*}
\frac{d^2S_j(\tau)}{d\tau^2} = (K_j + L_j)\frac{d\mathbb{P}(Y=1|\tau)}{d\tau}.
\end{equation*}

For $\tau \in (0, \alpha)$:
\begin{equation*}
\frac{d\mathbb{P}(Y=1|\tau)}{d\tau} = \frac{(1-\alpha)k}{\alpha\sin k}\cos\left(\frac{k\tau}{\alpha} - k\right) > 0,
\end{equation*}
since $\frac{k\tau}{\alpha} - k \in (-k, 0) \subset (-\pi/2, 0)$, ensuring $\cos(\cdot) > 0$.

For $\tau \in (\alpha, 1)$:
\begin{equation*}
\frac{d\mathbb{P}(Y=1|\tau)}{d\tau} = f'(\tau) = \frac{\alpha \sin k}{k(1-\alpha)} \cdot \frac{1}{\sqrt{1-\left(\frac{(\tau-1)\sin k}{1-\alpha}\right)^2}} > 0
\end{equation*}
since the argument of the square root lies in $(0,1]$ for $\tau \in (\alpha, 1]$.

Therefore, $\frac{d^2S_j(\tau)}{d\tau^2} > 0$ on $(0,\alpha) \cup (\alpha,1)$, establishing strict convexity on these open intervals.

\textit{Convexity at $\tau = \alpha$:}
To establish convexity on the entire interval $[0,1]$, we verify the one-sided derivative condition at $\tau = \alpha$. For a continuously differentiable function to be convex at a point where the second derivative is discontinuous, we require $\lim_{\tau \to \alpha^+} \frac{dS_j(\tau)}{d\tau} \geq \lim_{\tau \to \alpha^-} \frac{dS_j(\tau)}{d\tau}$.

From our analysis:
\begin{align*}
\lim_{\tau \to \alpha^-} \frac{dS_j(\tau)}{d\tau} &= K_j(-\alpha) + L_j(1-\alpha) = L_j - \alpha(K_j + L_j)\\
\lim_{\tau \to \alpha^+} \frac{dS_j(\tau)}{d\tau} &= K_j(-\alpha) + L_j(1-\alpha) = L_j - \alpha(K_j + L_j)
\end{align*}

The equality of these limits confirms that $S_j(\tau)$ has no corner at $\tau = \alpha$. Combined with strict convexity on $(0,\alpha)$ and $(\alpha,1)$, \rev{this establishes strict convexity of $S_j(\tau)$ on $[0,1]$: $S_j'$ is continuous and strictly increasing on $[0,1]$. The endpoint values $S_j'(0) = -K_j$ and $S_j'(1) = L_j$ follow from Lemma~\ref{lem:boundary}, and $S_j'(\tau) = (K_j+L_j)\mathbb{P}(Y=1|\tau) - K_j$ from Lemma~\ref{lem:monotonicity}. A convex combination $\sum_j q_j S_j$ with weights summing to one is then strictly convex as well.}

\textit{\rev{Strict} convexity of CVaR:}
\rev{Each $\Psi_q$ is strictly convex by the previous step. Let $\tau_0 \neq \tau_1$, $t \in (0,1)$ and $\tau_t = t\tau_1 + (1-t)\tau_0$, and let $\hat q$ attain the maximum in $\Psi(\tau_t) = \max_q \Psi_q(\tau_t)$. Then
\[
\Psi(\tau_t) = \Psi_{\hat q}(\tau_t) < t\,\Psi_{\hat q}(\tau_1) + (1-t)\,\Psi_{\hat q}(\tau_0) \leq t\,\Psi(\tau_1) + (1-t)\,\Psi(\tau_0),
\]
so $\Psi$ is strictly convex. A strictly convex continuous function on $[0,1]$ has a unique minimizer, which is interior because every $\Psi_q$ has right derivative $-\sum_j q_j K_j < 0$ at $0$ and left derivative $\sum_j q_j L_j > 0$ at $1$.}
\end{proof}

\begin{revblock}
\begin{lemma}[Finitely Many Tail Sets]\label{lem:tailsets}
For scenarios $i \neq j$ with distinct cost pairs $(K_i, L_i) \neq (K_j, L_j)$, the difference $S_i - S_j$ has at most two zeros on $[0,1]$. Hence the ordering of the scenario losses changes at most $J(J-1)$ times as $\tau$ runs over $[0,1]$, and at most $J(J-1)+1$ distinct tail sets occur.
\end{lemma}

\begin{proof}
\textnormal{By} Lemma~\ref{lem:convexity}, $(S_i - S_j)'(\tau) = (K_i - K_j + L_i - L_j)\,\mathbb{P}(Y=1|\tau) - (K_i - K_j)$. Since $\mathbb{P}(Y=1|\cdot)$ is strictly increasing and the two coefficients do not both vanish, this derivative is strictly monotone or a non-zero constant and changes sign at most once, so $S_i - S_j$ has at most one interior stationary point and at most two zeros. A tail set can change only where two losses cross.
\end{proof}

\begin{lemma}[Discrete CVaR and Its Derivatives]\label{lem:cvar_technical}
Let $n := \lfloor m \rfloor$ and let $S_{(1)}(\tau) \geq \dots \geq S_{(J)}(\tau)$ be the ordered scenario losses. Then
\begin{equation}\label{eq:exactcvar}
\text{CVaR}_\beta(S(\tau)) = \frac{1}{m}\Bigl[\sum_{i=1}^{n} S_{(i)}(\tau) + (m-n)\,S_{(n+1)}(\tau)\Bigr],
\end{equation}
which is the tail average $\frac{1}{m}\sum_{i \leq m} S_{(i)}(\tau)$ when $m$ is an integer. Moreover, with $\mathcal{Q}^\ast(\tau)$ the set of maximizers in \eqref{eq:envelope}:
\begin{enumerate}
\item $\text{CVaR}_\beta(S(\cdot))$ has one-sided derivatives everywhere. The right derivative $\partial_\tau^{+}\text{CVaR}_\beta(S(\tau))$ is the maximum and the left derivative $\partial_\tau^{-}\text{CVaR}_\beta(S(\tau))$ the minimum of $\sum_j q_j \rho_j(\tau)$ over $q \in \mathcal{Q}^\ast(\tau)$, and the subdifferential at $\tau$ is the interval between them;
\item if $\mathcal{Q}^\ast(\tau) = \{q\}$ is a singleton, $\text{CVaR}_\beta(S(\cdot))$ is differentiable at $\tau$ with derivative $\sum_j q_j \rho_j(\tau)$; this holds outside a finite set of thresholds when the scenarios have distinct cost pairs $(K_j, L_j)$, and on the open intervals between those thresholds $\text{CVaR}_\beta(S(\cdot))$ is twice continuously differentiable except at $\tau = \alpha$, with $\partial_{\tau\tau}\text{CVaR}_\beta(S(\tau)) = \sum_j q_j (K_j + L_j)\,\partial_\tau\mathbb{P}(Y=1|\tau) > 0$. By part (1), $\text{CVaR}_\beta(S(\cdot))$ is also differentiable at a threshold where all scenarios tied at $\text{VaR}_\beta$ have the same sensitivity;
\item at a threshold where $\mathcal{Q}^\ast(\tau)$ is not a singleton and the tied scenarios have different sensitivities $\rho_j(\tau)$, the one-sided derivatives differ, so $\text{CVaR}_\beta(S(\cdot))$ is not differentiable there.
\end{enumerate}
\end{lemma}

\begin{proof}
\textnormal{Equation}~\eqref{eq:exactcvar} is the value of the linear program \eqref{eq:envelope}. Part (1) follows from Danskin's theorem \citepec{danskin1966maxmin} for a maximum of continuously differentiable functions over a compact set. For part (2), the ordering of the losses changes only where two of them cross, which happens at finitely many thresholds by Lemma~\ref{lem:tailsets}; away from those, the maximizer is unique and the formulas follow from Lemma~\ref{lem:differentiability}. Part (3) follows from part (1), since the maximum and the minimum of $\sum_j q_j\rho_j(\tau)$ over $\mathcal{Q}^\ast(\tau)$ differ exactly when the tied scenarios have different sensitivities.
\end{proof}

\begin{algorithm}[t]
\color{black}
\caption{Bisection on a CVaR subgradient}\label{alg:bisection}
\begin{algorithmic}[1]
\REQUIRE costs $(K_j, L_j)_{j \leq J}$, level $\beta$, calibration map $\mathbb{P}(Y=1\mid\cdot)$, tolerance $\varepsilon$
\STATE $\tau_{\mathrm{lo}} \leftarrow 0$, $\tau_{\mathrm{hi}} \leftarrow 1$
\WHILE{$\tau_{\mathrm{hi}} - \tau_{\mathrm{lo}} > \varepsilon$}
\STATE $\tau \leftarrow (\tau_{\mathrm{lo}}+\tau_{\mathrm{hi}})/2$; compute and sort $S_j(\tau)$, $j = 1,\dots,J$
\STATE $q \leftarrow$ tail weights at $\tau$: $1/m$ on the $\lfloor m \rfloor$ largest losses, $(m - \lfloor m \rfloor)/m$ on the next, $0$ otherwise
\STATE $v \leftarrow \sum_j q_j \left[(K_j + L_j)\,\mathbb{P}(Y=1\mid\tau) - K_j\right]$
\STATE \textbf{if} $v > 0$ \textbf{then} $\tau_{\mathrm{hi}} \leftarrow \tau$ \textbf{else} $\tau_{\mathrm{lo}} \leftarrow \tau$
\ENDWHILE
\RETURN $(\tau_{\mathrm{lo}}+\tau_{\mathrm{hi}})/2$
\end{algorithmic}
\end{algorithm}

\begin{proposition}[Correctness and Complexity of Algorithm~\ref{alg:bisection}]\label{prop:algorithm}
Algorithm~\ref{alg:bisection} returns a point within $\varepsilon$ of $\tau^{\ast}$ after $\lceil \log_2(1/\varepsilon) \rceil$ iterations, each costing $O(J\log J)$ operations.
\end{proposition}

\begin{proof}
\textnormal{The} function $\Psi$ is convex by Lemma~\ref{lem:convexity}. The weights $q$ of line 4 belong to $\mathcal{Q}^\ast(\tau)$, so by Lemma~\ref{lem:cvar_technical}(1) the quantity $v$ of line 5 is a subgradient of $\Psi$ at $\tau$. The subgradient inequality $\Psi(\tau^{\ast}) \geq \Psi(\tau) + v(\tau^{\ast} - \tau)$ together with $\Psi(\tau^{\ast}) \leq \Psi(\tau)$ gives $v(\tau^{\ast} - \tau) \leq 0$. Hence $\tau^{\ast} \leq \tau$ if $v > 0$, and $\tau^{\ast} \geq \tau$ if $v < 0$. If $v = 0$, then $0$ is a subgradient of $\Psi$ at $\tau$, so $\tau$ minimizes $\Psi$ and $\tau = \tau^{\ast}$ by Lemma~\ref{lem:convexity}. The invariant $\tau^{\ast} \in [\tau_{\mathrm{lo}}, \tau_{\mathrm{hi}}]$ is therefore maintained while the interval halves at each iteration, and sorting dominates the cost of an iteration.
\end{proof}

\begin{lemma}[Continuity of the Optimal Threshold in the Costs]\label{lem:continuity}
Let $k < \pi/2$ and $\rev{p_{\min}} := \min_{\tau \in [0,1]} \partial_\tau\mathbb{P}(Y=1|\tau) = \min\left\{\frac{(1-\alpha)k\cos k}{\alpha\sin k}, \frac{\alpha\sin k}{k(1-\alpha)}\right\} > 0$. Let $\tau^{\ast}$ and $\tilde\tau^{\ast}$ be the optimal thresholds for cost vectors $(K, L)$ and $(\tilde K, \tilde L)$, and set
\[
\Delta_{\mathrm{cost}} := \max_j \bigl[\alpha|K_j - \tilde K_j| + (1-\alpha)|L_j - \tilde L_j|\bigr], \qquad
\sigma := \rev{p_{\min}}\,\min_j \min\{K_j + L_j,\ \tilde K_j + \tilde L_j\}.
\]
Then $|\tau^{\ast} - \tilde\tau^{\ast}| \leq \sqrt{2\Delta_{\mathrm{cost}}/\sigma}$.
\end{lemma}

\begin{proof}
\textnormal{By} the derivative formulas in the proof of Lemma~\ref{lem:convexity}, $\partial_\tau\mathbb{P}(Y=1|\tau)$ is bounded below on $[0,\alpha)$ by its value at $\tau = 0$ and on $(\alpha,1]$ by its value at $\tau = 1$, which gives $\rev{p_{\min}}$. For every $q \in \mathcal{Q}_\beta$, $\Psi_q$ has an absolutely continuous derivative with $\Psi_q'' \geq \sigma$ almost everywhere, so it is $\sigma$-strongly convex. A maximum of $\sigma$-strongly convex functions is $\sigma$-strongly convex, because $\Psi - \frac{\sigma}{2}\tau^2$ is a maximum of convex functions. Since $\tau^{\ast}$ is an interior minimizer of $\Psi$, $0$ is a subgradient of $\Psi$ at $\tau^{\ast}$, and strong convexity gives
\[
\Psi(\tau) \geq \Psi(\tau^{\ast}) + \tfrac{\sigma}{2}(\tau - \tau^{\ast})^2 \quad\text{for all } \tau, \qquad \tilde\Psi(\tau) \geq \tilde\Psi(\tilde\tau^{\ast}) + \tfrac{\sigma}{2}(\tau - \tilde\tau^{\ast})^2 \quad\text{for all } \tau,
\]
where $\tilde\Psi$ is the CVaR objective under the costs $(\tilde K, \tilde L)$. Next, $|S_j(\tau) - \tilde S_j(\tau)| \leq |K_j - \tilde K_j|\mathbb{P}_{\mathrm{FP}}(\tau) + |L_j - \tilde L_j|\mathbb{P}_{\mathrm{FN}}(\tau) \leq \Delta_{\mathrm{cost}}$, because $\mathbb{P}_{\mathrm{FP}} \leq \alpha$ and $\mathbb{P}_{\mathrm{FN}} \leq 1-\alpha$. As $\text{CVaR}_\beta$ is monotone and translation equivariant, $\|\Psi - \tilde\Psi\|_\infty \leq \Delta_{\mathrm{cost}}$. Evaluating the first inequality at $\tilde\tau^{\ast}$ and the second at $\tau^{\ast}$ and adding them,
\[
\sigma(\tau^{\ast} - \tilde\tau^{\ast})^2 \leq \bigl[\Psi(\tilde\tau^{\ast}) - \tilde\Psi(\tilde\tau^{\ast})\bigr] + \bigl[\tilde\Psi(\tau^{\ast}) - \Psi(\tau^{\ast})\bigr] \leq 2\Delta_{\mathrm{cost}}.
\]
\end{proof}

\end{revblock}\begin{revblock}
\begin{remark}[Scope of the Results]\label{rem:scope}
Except for the closed forms (the two cases of Theorem~\ref{thm:optimal_threshold}, Lemma~\ref{lem:accuracy_threshold} and the constants of Proposition~\ref{prop:efficiency_gap}), the results of Sections~\ref{sec:baselineriskmanagement}--\ref{sec:extensions} use only that $\mathbb{P}(Y=1 \mid \cdot)$ is continuous and strictly increasing from $0$ to $1$. Four results need one further property. Proposition~\ref{prop:risk_penalty} needs an absolutely continuous derivative, Proposition~\ref{prop:monotonicity_beta} a map that is continuously differentiable near $\tau^{\ast}$ with a positive derivative at $\tau^{\ast}$, Lemma~\ref{lem:continuity} a positive lower bound on the derivative, and Lemma~\ref{lem:cvar_drift} a quality-ordered family of ROC curves. Section~\ref{sec:binormal} gives the counterparts of the closed forms for the equal-variance binormal model, used alongside the trigonometric model in Section~\ref{sec:casestudy}.
\end{remark}
\end{revblock}

\FloatBarrier
\subsection{Comparison with Accuracy Benchmark}\label{sec:acccomparison}

\begin{lemma}[Simplified Accuracy Expression]\label{lem:accuracy_form}
Under the trigonometric model, accuracy can be expressed as:
\begin{equation}\label{eq:accuracydef}
\text{Acc}(\tau) = \tau + (1-\alpha) - 2\mathbb{P}_{\mathrm{FN}}(\tau).
\end{equation}
\end{lemma}

\begin{proof}
\textnormal{From} the definition in Section \ref{sec:theorybenchmarks}, $\text{Acc}(\tau) = \int_\tau^1 \mathbb{P}(Y=1|\overline{\gamma}) d\overline{\gamma} + \int_0^\tau [1 - \mathbb{P}(Y=1|\overline{\gamma})] d\overline{\gamma}$. 
Using $\int_0^1 \mathbb{P}(Y=1|\overline{\gamma}) d\overline{\gamma} = 1-\alpha$ and $\mathbb{P}_{\mathrm{FN}}(\tau) = \int_0^\tau \mathbb{P}(Y=1|\overline{\gamma}) d\overline{\gamma}$, 
the result follows by manipulation.
\end{proof}

\begin{lemma}[Fundamental Divergence Between Accuracy and Risk]\label{lem:accuracy_risk_divergence}
The accuracy-optimal and CVaR-optimal thresholds coincide ($\tau_{\mathrm{acc}} = \tau^{\ast}$) if and only if the tail-weighted cost ratio equals unity:
\begin{equation}
\frac{\sum_{j \in \mathcal{J}_\beta} \sum_{i=1}^N K_{ji}}{\sum_{j \in \mathcal{J}_\beta} \sum_{i=1}^N L_{ji}} = 1
\end{equation}
where $\mathcal{J}_\beta$ denotes the \rev{tail set and the sums carry the worst-case weights of Theorem~\ref{thm:optimal_threshold}, as in Section~\ref{sec:theorybenchmarks}}. In all other cases, $\tau_{\mathrm{acc}} \neq \tau^{\ast}$.
\end{lemma}

\begin{proof}
\textnormal{The} CVaR-optimal threshold satisfies $\mathbb{P}(Y=1|\tau^{\ast}) = \frac{\sum_{j \in \mathcal{J}_\beta} \sum_{i=1}^N K_{ji}}{\sum_{j \in \mathcal{J}_\beta} \sum_{i=1}^N (K_{ji} + L_{ji})}$ from the first-order condition. \rev{Since $\mathbb{P}(Y=1|\tau_{\mathrm{acc}}) = \frac{1}{2}$ and $\mathbb{P}(Y=1|\cdot)$ is strictly increasing, $\tau_{\mathrm{acc}} = \tau^{\ast}$ if and only if the two probabilities coincide, i.e., if and only if the numerator equals half the denominator and the tail costs are balanced.}
\end{proof}

\FloatBarrier
\subsection[Expected Loss-Minimizing Threshold]{Expected Loss-{\color{black}Minimizing} Threshold}\label{sec:explossmaximizer}

\begin{definition}[Expected Loss under $\tau$]
The expected loss for threshold $\tau$ is:
\begin{equation}\label{def:expectedloss}
\mathbb{E}[S(\tau)] = \frac{1}{J} \sum_{j=1}^{J} S_j(\tau) = \sum_{i=1}^{N} \left[\bar{K}_i \mathbb{P}_{\mathrm{FP}}(\tau) + \bar{L}_i \mathbb{P}_{\mathrm{FN}}(\tau)\right]
\end{equation}
where $\bar{K}_i = \frac{1}{J}\sum_{j=1}^{J} K_{ji}$ and $\bar{L}_i = \frac{1}{J}\sum_{j=1}^{J} L_{ji}$ are the average costs across scenarios.
\end{definition}

\begin{revblock}
\begin{proposition}[Expected Loss-Optimal Threshold]\label{prop:expected_loss_threshold}
The threshold $\tau_{\text{EL}}^\ast \in (0,1)$ that minimizes $\mathbb{E}[S(\tau)]$ is unique and satisfies $\mathbb{P}(Y=1|\tau_{\text{EL}}^\ast) = \bar{p} := \bar{K}/(\bar{K}+\bar{L})$, where $\bar{K} := \sum_i \bar{K}_i$ and $\bar{L} := \sum_i \bar{L}_i$. Under the trigonometric model it takes the closed forms of Theorem~\ref{thm:optimal_threshold} with $\bar{p}$ in place of $p^\ast$: $\tau_{\text{EL}}^\ast = \alpha + \frac{\alpha}{k}\arcsin\left(\left(\frac{\bar{p}}{1-\alpha} - 1\right)\sin k\right)$ if $\bar{p} \leq 1-\alpha$, and $\tau_{\text{EL}}^\ast = 1 - (1-\alpha)\frac{\sin\left(\frac{k}{\alpha}(1-\bar{p})\right)}{\sin k}$ otherwise.
\end{proposition}

\begin{proof}
\textnormal{The} derivative $\frac{d\mathbb{E}[S(\tau)]}{d\tau} = \bar{K}(\mathbb{P}(Y=1|\tau) - 1) + \bar{L}\mathbb{P}(Y=1|\tau)$ is negative at $0$ and positive at $1$, and $\mathbb{E}[S(\tau)] = \bar{K}\mathbb{P}_{\mathrm{FP}}(\tau) + \bar{L}\mathbb{P}_{\mathrm{FN}}(\tau)$ is strictly convex by Lemma~\ref{lem:convexity}. Hence the unique minimizer is the root of the first-order condition $\mathbb{P}(Y=1|\tau_{\text{EL}}^\ast) = \bar{p}$. The closed forms follow by inverting Equation~\eqref{eq:trig-model} exactly as in the proof of Theorem~\ref{thm:optimal_threshold}, with $\bar{p}$ in place of $p^\ast$.
\end{proof}
\end{revblock}

The expected loss-optimal threshold has the same functional form as the CVaR-optimal threshold, but uses average costs $(\bar{K}, \bar{L})$ rather than tail-weighted costs. When cost structures are homogeneous across scenarios, the thresholds coincide; otherwise they diverge proportionally to the difference between average and tail cost ratios.

\begin{lemma}[Fundamental Divergence Between Expected Loss and CVaR]\label{lem:el_cvar_divergence}
The expected loss-optimal and CVaR-optimal thresholds coincide ($\tau_{\text{EL}}^\ast = \tau^{\ast}$) if and only if the tail-weighted cost ratio equals the average cost ratio:
\begin{equation}
\frac{\sum_{j \in \mathcal{J}_\beta} \sum_{i=1}^N K_{ji}}{\sum_{j \in \mathcal{J}_\beta} \sum_{i=1}^N L_{ji}} = \frac{\sum_{j=1}^{J} \sum_{i=1}^N K_{ji}}{\sum_{j=1}^{J} \sum_{i=1}^N L_{ji}}
\end{equation}
\end{lemma}

\begin{proof}
\textnormal{From} the first-order conditions, the CVaR-optimal threshold satisfies: $\mathbb{P}(Y=1|\tau^{\ast}) = \frac{\sum_{j \in \mathcal{J}_\beta} \sum_{i=1}^N K_{ji}}{\sum_{j \in \mathcal{J}_\beta} \sum_{i=1}^N (K_{ji} + L_{ji})}$,
while the expected loss-optimal threshold satisfies:
\begin{equation*}
\mathbb{P}(Y=1|\tau_{\text{EL}}^\ast) = \frac{\bar{K}}{\bar{K} + \bar{L}} = \frac{\sum_{i=1}^N \bar{K}_i}{\sum_{i=1}^N (\bar{K}_i + \bar{L}_i)} = \frac{\sum_{j=1}^{J} \sum_{i=1}^N K_{ji}}{\sum_{j=1}^{J} \sum_{i=1}^N (K_{ji} + L_{ji})}
\end{equation*}
where the last equality follows since $\sum_{i=1}^N \bar{K}_i = \frac{1}{J}\sum_{j=1}^{J} \sum_{i=1}^N K_{ji}$ and the $\frac{1}{J}$ factors cancel.

Since $\mathbb{P}(Y=1|\cdot)$ is strictly monotonic by Lemma~\ref{lem:monotonicity}, we have $\tau_{\text{EL}}^\ast = \tau^{\ast}$ if and only if $\mathbb{P}(Y=1|\tau_{\text{EL}}^\ast) = \mathbb{P}(Y=1|\tau^{\ast})$. This equality holds if and only if:
\begin{equation*}
\frac{\sum_{j \in \mathcal{J}_\beta} \sum_{i=1}^N K_{ji}}{\sum_{j \in \mathcal{J}_\beta} \sum_{i=1}^N (K_{ji} + L_{ji})} = \frac{\sum_{j=1}^{J} \sum_{i=1}^N K_{ji}}{\sum_{j=1}^{J} \sum_{i=1}^N (K_{ji} + L_{ji})}
\end{equation*}

Cross-multiplying and simplifying shows this holds if and only if the cost ratios are equal:
$$\frac{\sum_{j \in \mathcal{J}_\beta} \sum_{i=1}^N K_{ji}}{\sum_{j \in \mathcal{J}_\beta} \sum_{i=1}^N L_{ji}} = \frac{\sum_{j=1}^{J} \sum_{i=1}^N K_{ji}}{\sum_{j=1}^{J} \sum_{i=1}^N L_{ji}}.$$

Equivalently, $\tau_{\text{EL}}^\ast = \tau^{\ast}$ if and only if the tail-weighted cost ratio equals the average cost ratio, $\bar{r}_{\text{tail}} = \bar{r}_{\text{avg}}$\rev{, where $\bar{r}_{\text{avg}} := \bar{K}/\bar{L}$}.
\end{proof}

\begin{proposition}[Efficiency Gap Between Expected Loss and CVaR]\label{prop:efficiency_gap_el}
Fix $\alpha \in (0,1)$ and $k\in(0,\pi/2)$. Let $\tau^{\ast}$ be the CVaR-optimal threshold and $\tau_{\text{EL}}^\ast$ the expected loss-optimal threshold.

\medskip\noindent
\textit{(i) \rev{Proportional} costs.}  
If $K_{ji}=c_jK_i$ and $L_{ji}=c_jL_i$ for all $i,j$, then $\Delta_\tau^{\text{EL}}=|\tau^{\ast} - \tau_{\text{EL}}^\ast|=0.$

\smallskip\noindent
\textit{(ii) Heterogeneous costs.}  
Assume $\tau^{\ast},\tau_{\text{EL}}^\ast$ lie in the same region \rev{(both in $(0,\alpha)$ or both in $(\alpha,1)$)}. Define the cost ratio gap:
\[
\Delta_r := \left|\frac{\bar{r}_{\text{tail}}}{1+\bar{r}_{\text{tail}}} - \frac{\bar{r}_{\text{avg}}}{1+\bar{r}_{\text{avg}}}\right|,
\]
where $\bar{r}_{\text{tail}} = \frac{\sum_{j\in\mathcal{J}_\beta}\sum_i K_{ji}}{\sum_{j\in\mathcal{J}_\beta}\sum_i L_{ji}}$ and $\bar{r}_{\text{avg}} = \frac{\bar{K}}{\bar{L}}$\rev{, the tail sums carrying the worst-case weights of Theorem~\ref{thm:optimal_threshold} as in Section~\ref{sec:theorybenchmarks}}. With the same branch-specific derivative bounds $\rev{p_{\min}}$ and $\rev{p_{\max}}$ as in Proposition~\ref{prop:efficiency_gap}, and $C_1:=1/\rev{p_{\max}}$ and $C_2:=1/\rev{p_{\min}}$: $
C_1 \Delta_r \leq \Delta_\tau^{\text{EL}} \leq C_2 \Delta_r.$
\end{proposition}

\begin{proof}
\textnormal{Part} (i): When costs are proportional, all scenarios have the same relative cost trade-offs:
\[
\bar{r}_{\text{tail}} = \frac{\sum_{j\in\mathcal{J}_\beta} c_j \sum_i K_i}{\sum_{j\in\mathcal{J}_\beta} c_j \sum_i L_i} = \frac{\sum_i K_i}{\sum_i L_i} = \frac{\sum_{j=1}^J c_j \sum_i K_i}{\sum_{j=1}^J c_j \sum_i L_i} = \bar{r}_{\text{avg}}
\]
By Lemma~\ref{lem:el_cvar_divergence}, this implies $\tau^{\ast} = \tau_{\text{EL}}^\ast$.

Part (ii): The first-order conditions yield:
\begin{align*}
\mathbb{P}(Y=1|\tau^{\ast}) &= \frac{\bar{r}_{\text{tail}}}{1+\bar{r}_{\text{tail}}} \qquad\text{and}\qquad
\mathbb{P}(Y=1|\tau_{\text{EL}}^\ast) = \frac{\bar{r}_{\text{avg}}}{1+\bar{r}_{\text{avg}}}
\end{align*}

The probability gap is thus: $\Delta_p = \left|\mathbb{P}(Y=1|\tau^{\ast}) - \mathbb{P}(Y=1|\tau_{\text{EL}}^\ast)\right| = \Delta_r$. \rev{Following the same integration argument as in Proposition~\ref{prop:efficiency_gap}}, with $\rev{p_{\min}} \leq \partial_\tau\mathbb{P}(Y=1|\tau) \leq \rev{p_{\max}}$ on each region, we obtain the stated bounds.
\end{proof}

\begin{remark}
If $\tau^{\ast}$ and $\tau_{\text{EL}}^\ast$ lie in opposite regions \rev{(one in $(0,\alpha)$ and the other in $(\alpha,1)$)}, then $\Delta_\tau^{\text{EL}} \geq \max\{|\tau^{\ast}-\alpha|, |\tau_{\text{EL}}^\ast-\alpha|\}$.    
\end{remark}
When comparing Propositions~\ref{prop:efficiency_gap} and \ref{prop:efficiency_gap_el}, we observe that while accuracy maximization ignores cost structures entirely—treating all errors as equally important—expected loss optimization occupies a middle ground between this cost-blind approach and full risk awareness \rev{when the average cost share of false positives lies between the tail share and one half (Section~\ref{sec:theorybenchmarks})}. \rev{In that case the gap of expected loss optimization is at most that of accuracy optimization.} Accuracy optimization produces gaps proportional to $|\bar{r}_{\text{tail}}/(1+\bar{r}_{\text{tail}}) - 1/2|$, which can approach 0.5 when tail costs are severely imbalanced, whereas expected loss optimization yields gaps proportional to $|\bar{r}_{\text{tail}} - \bar{r}_{\text{avg}}|/\rev{[}(1+\bar{r}_{\text{tail}})(1+\bar{r}_{\text{avg}})\rev{]}$, which vanishes when tail and average scenarios share similar cost structures. 

\begin{proposition}[Risk Penalty of Expected Loss Optimization]\label{prop:risk_penalty_el}
Fix $\alpha \in (0,1)$, $k\in(0,\pi/2)$, and set $\tau^{\ast}$ (CVaR-optimal) and $\tau_{\text{EL}}^\ast$ (expected loss-optimal). Define 
\[
\Delta_{\text{CVaR}}^{\text{EL}} := \frac{\text{CVaR}_\beta(S(\tau_{\text{EL}}^\ast)) - \text{CVaR}_\beta(S(\tau^{\ast}))}{\text{CVaR}_\beta(S(\tau^{\ast}))}, \qquad
\Delta_\tau^{\text{EL}} := |\tau^{\ast} - \tau_{\text{EL}}^\ast|.
\]
\noindent
\textit{(i) \rev{Proportional} costs.}  
If $K_{ji}=c_jK_i$ and $L_{ji}=c_jL_i$ (all $i,j$), then $\Delta_{\text{CVaR}}^{\text{EL}} = 0$.

\smallskip\noindent
\rev{\textit{(ii) Heterogeneous costs.}  
The bounds of Proposition~\ref{prop:risk_penalty}(ii) hold with $\tau_{\text{EL}}^\ast$ in place of $\tau_{\mathrm{acc}}$: with $q_{\text{EL}}$ any maximizing weights in \eqref{eq:envelope} at $\tau_{\text{EL}}^\ast$, $I$ the interval between $\tau^{\ast}$ and $\tau_{\text{EL}}^\ast$, and $\tilde{m}_{\text{EL}}, \tilde{M}_{\text{EL}}$ defined as $\tilde{m}, \tilde{M}$ there,
\[
\frac{\tilde{m}_{\text{EL}}}{2\,\text{CVaR}_\beta(S(\tau^{\ast}))} (\Delta_\tau^{\text{EL}})^2 
\leq 
\Delta_{\text{CVaR}}^{\text{EL}} 
\leq 
\frac{\tilde{M}_{\text{EL}}(\Delta_\tau^{\text{EL}})^2/2 + \Delta_\tau^{\text{EL}}\,|\sum_j q_{\text{EL},j}\,\rho_j(\tau^{\ast})|}{\text{CVaR}_\beta(S(\tau^{\ast}))}.
\]}
\end{proposition}

\begin{proof}
\textnormal{For} Part (i), when costs are homogeneous, from Proposition~\ref{prop:efficiency_gap_el} and Lemma~\ref{lem:el_cvar_divergence} it follows $\tau^{\ast} = \tau_{\text{EL}}^\ast$. Hence, $\Delta_{\text{CVaR}}^{\text{EL}} = 0$. \rev{Part (ii) follows by repeating the proof of Proposition~\ref{prop:risk_penalty}(ii) with $\tau_{\text{EL}}^\ast$ and $q_{\text{EL}}$ in place of $\tau_{\mathrm{acc}}$ and $q_{\mathrm{acc}}$; no step uses the definition of the benchmark threshold.}
\end{proof}

\rev{As for Proposition~\ref{prop:risk_penalty}, the first-order term vanishes when $q_{\text{EL}}$ can be taken equal to $q^\ast$.} The bounds of Proposition~\ref{prop:risk_penalty_el} reveal that tail losses rise 
\rev{at least} quadratically with threshold misalignment. Unlike inventory management where safety stock 
provides a buffer, binary classification offers no protection when thresholds are calibrated 
for average rather than tail conditions. Moreover, scenarios with the largest second derivatives—where 
calibration matters most—are typically rare in historical data and hardest to price ex ante.

\FloatBarrier
\subsection{Insurance Value}\label{sec:theoryinsurance}

\begin{revblock}
\begin{corollary}[Viability of Full Coverage]\label{cor:viability}
In the full-coverage limit of Proposition~\ref{prop:limit_behavior}(2), the base value is positive if and only if
$(\rho\gamma - r)\,\text{CVaR}_\beta[S_L(\tau^{\ast})] + \rho(\gamma - \gamma_{\text{ins}})\,s_{\mathrm{op}} > (\mu - r)\,\mathbb{E}[S_L(\tau^{\ast})]$.
Since the limit equals $(\rho\gamma - r)\bigl(\text{CVaR}_\beta[S_L] - \mathbb{E}[S_L]\bigr) + (\rho\gamma - \mu)\,\mathbb{E}[S_L] + \rho(\gamma - \gamma_{\text{ins}})\,s_{\mathrm{op}}$, full coverage creates value whenever $\rho\gamma \geq \max\{\mu, r\}$ and at least one of the three terms is positive, i.e., $\rho\gamma > \mu$, or $\rho\gamma > r$ with $\text{CVaR}_\beta[S_L] > \mathbb{E}[S_L]$, or $s_{\mathrm{op}} > 0$ with $\gamma > \gamma_{\text{ins}}$. When $s_{\mathrm{op}} = 0$ and $\rho\gamma > r$, the condition reads
\[
\chi > \frac{\mu - r}{\rho\gamma - r}, \qquad \chi := \frac{\text{CVaR}_\beta[S_L(\tau^{\ast})]}{\mathbb{E}[S_L(\tau^{\ast})]} \geq 1,
\]
where $\chi$ is the tail-to-average ratio of the insured loss.
\end{corollary}
\end{revblock}

\FloatBarrier
\subsection{Framework Extensions: Model Drift}\label{sec:theorydrift}

\begin{revblock}
\begin{corollary}[Square-Root Rule]\label{cor:sqrt}
If the excess-risk cost grows linearly, $e(t) = s\,t$ with $s > 0$, then the optimal renewal time of Proposition~\ref{prop:optimal_duration_full} is $T^{\ast} = \sqrt{2\mathcal{C}_a/s}$.
\end{corollary}

\begin{corollary}[Recalibration Cannot Restore the Baseline]\label{cor:recalibration}
Let $\tau^{\ast}_t := \arg\min_\tau \text{CVaR}_\beta(S(\tau, t))$ be the threshold re-optimized on the degraded model at date $t$. If $h(t) < 1$, then $\text{CVaR}_\beta(S(\tau^{\ast}_t, t)) > \text{CVaR}_\beta(S(\tau^{\ast}, 0))$, so the recalibrated excess $e_{\mathrm{rc}}(t) := \lambda[\text{CVaR}_\beta(S(\tau^{\ast}_t, t)) - \text{CVaR}_\beta(S(\tau^{\ast}, 0))]$ satisfies $0 < e_{\mathrm{rc}}(t) \leq e(t)$. Re-optimizing the threshold removes only the excess risk due to the threshold, and only retraining returns the tail risk to its inception level. If $e - e_{\mathrm{rc}}$ is non-decreasing on $[0, T^{\ast}_{\mathrm{rc}}]$, where $T^{\ast}_{\mathrm{rc}}$ is the largest optimal renewal time under recalibration, then recalibrating within the contract weakly lengthens the optimal renewal time.
\end{corollary}
\end{revblock}

\begin{revblock}
\begin{lemma}[Monotonic Growth of Tail Risk]\label{lem:cvar_drift}
Let $\{P_\eta\}$ be a family of calibration maps indexed by a quality parameter $\eta$, with error probabilities $\mathbb{P}_{\mathrm{FN}}(\tau; \eta)$ and $\mathbb{P}_{\mathrm{FP}}(\tau; \eta)$, and suppose the ROC curve at $\eta$ strictly dominates the one at $\eta' < \eta$ (for every false positive rate $x \in (0,1)$ the sensitivity at $\eta$ is strictly larger). Then for every $\tau \in (0,1)$,
\[
\mathbb{P}_{\mathrm{FN}}(\tau; \eta') > \mathbb{P}_{\mathrm{FN}}(\tau; \eta) \quad\text{and}\quad \mathbb{P}_{\mathrm{FP}}(\tau; \eta') > \mathbb{P}_{\mathrm{FP}}(\tau; \eta),
\]
so every scenario loss and $\text{CVaR}_\beta(S(\tau; \cdot))$ are strictly larger at $\eta'$. In particular, under $k(t) = k_0 h(t)$ with $h$ non-increasing, $t \mapsto \text{CVaR}_\beta(S(\tau, t))$ is non-decreasing for every fixed $\tau$, and strictly increasing where $h$ strictly decreases. The hypothesis holds for the trigonometric model in $k \in (0, \pi/2]$ and for the equal-variance binormal model of Section~\ref{sec:binormal} in $d > 0$.
\end{lemma}

\begin{proof}
\textnormal{At} a fixed rank threshold $\tau$ the flagged share is $1 - \tau$ whatever the quality, and with $x = \mathbb{P}_{\mathrm{FP}}/\alpha$ (false positive rate) and $y = 1 - \mathbb{P}_{\mathrm{FN}}/(1-\alpha)$ (sensitivity) the operating point satisfies $\alpha x + (1-\alpha)y = 1 - \tau$. The operating points $(x, y)$ at $\eta$ and $(x', y')$ at $\eta'$ lie on this decreasing line and on their increasing ROC curves $R_\eta$ and $R_{\eta'}$. If $x \geq x'$, then $y = R_\eta(x) \geq R_\eta(x') > R_{\eta'}(x') = y'$, so $\alpha x + (1-\alpha)y > \alpha x' + (1-\alpha)y'$, a contradiction; hence $x < x'$ and then $y > y'$. The claims for $S_j$ and $\text{CVaR}_\beta$ follow by monotonicity of both.

\emph{Trigonometric model.} Since $\mathbb{P}_{\mathrm{FP}} = \alpha - \tau + \mathbb{P}_{\mathrm{FN}}$ (Lemma~\ref{lem:fp_case}), ROC dominance in $k$ is equivalent to $\mathbb{P}_{\mathrm{FN}}(\tau; k)$ decreasing in $k$ for every $\tau$. On $[0,\alpha]$, Equation~\eqref{eq:trig-model} gives $\mathbb{P}(Y=1 \mid \overline{\gamma}; k) = (1-\alpha)[1 - \sin(ku)/\sin k]$ with $u = 1 - \overline{\gamma}/\alpha \in (0,1)$, and $\sin(ku)/\sin k$ is increasing in $k$ because $\frac{d}{dk}\frac{\sin ku}{\sin k} = \frac{\sin k \cos ku}{\sin^2 k}\bigl[u - \frac{\tan ku}{\tan k}\bigr] \geq 0$ by convexity of $\tan$ on $[0, \pi/2)$. Hence $\mathbb{P}(Y=1 \mid \overline{\gamma}; k)$ is decreasing in $k$ for $\overline{\gamma} \in (0,\alpha)$, and so is $\mathbb{P}_{\mathrm{FN}}(\tau; k) = \int_0^\tau \mathbb{P}(Y=1 \mid \overline{\gamma}; k)\,d\overline{\gamma}$ for $\tau \leq \alpha$. On $(\alpha, 1)$, $\mathbb{P}(Y=1 \mid \overline{\gamma}; k) = 1 - \alpha + \alpha[1 - \arcsin(w \sin k)/k]$ with $w = (1-\overline{\gamma})/(1-\alpha) \in (0,1)$. The function $\vartheta(k) = \arcsin(w \sin k)$ satisfies $\vartheta(0) = 0$ and $\vartheta''(k) = w \sin k\,(w^2 - 1)/(1 - w^2\sin^2 k)^{3/2} < 0$, so $\vartheta(k)/k$ is decreasing and $\mathbb{P}(Y=1 \mid \overline{\gamma}; k)$ is increasing in $k$ for $\overline{\gamma} \in (\alpha, 1)$. By Lemma~\ref{lem:conservation}, $\mathbb{P}_{\mathrm{FN}}(\tau; k) = (1-\alpha) - \int_\tau^1 \mathbb{P}(Y=1 \mid \overline{\gamma}; k)\,d\overline{\gamma}$ is then decreasing in $k$ for $\tau > \alpha$ as well, strictly for $\tau \in (0,1)$.

\emph{Binormal model.} The sensitivity at false positive rate $x$ is $\Phi(d - \Phi^{-1}(1-x))$, strictly increasing in $d$.
\end{proof}
\end{revblock}

\FloatBarrier
\subsection{The Binormal Model}\label{sec:binormal}

\begin{revblock}
In the equal-variance binormal model \citepec{pepe2003statistical}, the latent score satisfies $z \mid Y=0 \sim N(0,1)$ and $z \mid Y=1 \sim N(d,1)$ with $d = \sqrt{2}\,\Phi^{-1}(\mathrm{AUC}) > 0$, where $\Phi$ is the standard normal distribution function and $\bar\Phi := 1 - \Phi$. Let $\overline{\gamma} = F(z)$, $F(z) = \alpha\Phi(z) + (1-\alpha)\Phi(z-d)$, be its rank. Then
\begin{equation}\label{eq:binormalP}
\mathbb{P}(Y=1 \mid \overline{\gamma}) = \Bigl[1 + \exp\bigl(-\operatorname{logit}(1-\alpha) - d\,F^{-1}(\overline{\gamma}) + d^2/2\bigr)\Bigr]^{-1},
\end{equation}
and the error probabilities at rank threshold $\tau$ are $\mathbb{P}_{\mathrm{FP}}(\tau) = \alpha\,\bar\Phi(F^{-1}(\tau))$ and $\mathbb{P}_{\mathrm{FN}}(\tau) = (1-\alpha)\,\Phi(F^{-1}(\tau) - d)$.

\begin{lemma}[Binormal Counterparts]\label{lem:binormal}
(a) The map \eqref{eq:binormalP} is continuous and strictly increasing with $\mathbb{P}(Y=1 \mid 0^+) = 0$ and $\mathbb{P}(Y=1 \mid 1^-) = 1$, and its derivative is bounded below by a positive constant on $(0,1)$; hence all results listed in Remark~\ref{rem:scope} apply. (b) The threshold with $\mathbb{P}(Y=1 \mid \tau) = p$ is $\tau = F(z_p)$ with
\[
z_p = \frac{\operatorname{logit}(p) - \operatorname{logit}(1-\alpha) + d^2/2}{d};
\]
in particular $\tau^{\ast} = F(z_{p^\ast})$ with $p^\ast$ of Equation~\eqref{eq:foc}, $\tau_{\mathrm{acc}} = F(z_{1/2})$ and $\tau_{\text{EL}}^\ast = F(z_{\bar p})$. (c) On the latent scale the efficiency gap is exact, $|z_{p^\ast} - z_{1/2}| = |\ln \bar r_{\text{tail}}|/d$ with $\bar r_{\text{tail}} = K_{q^\ast}/L_{q^\ast}$, so the gap vanishes if and only if $\bar r_{\text{tail}} = 1$ and shrinks with the discrimination $d$. Its rank-scale value is $\Delta_\tau = |F(z_{p^\ast}) - F(z_{1/2})|$.
\end{lemma}

\begin{proof}
\textnormal{(a)} $\operatorname{logit}\mathbb{P}(Y=1 \mid F(z)) = \operatorname{logit}(1-\alpha) + dz - d^2/2$ is affine and increasing in $z$, and $F$ is a strictly increasing bijection of $\mathbb{R}$ onto $(0,1)$; the limits follow as $z \to \mp\infty$. Writing $P$ for $\mathbb{P}(Y=1 \mid \tau)$, the derivative $\partial_\tau\mathbb{P}(Y=1 \mid \tau) = d\,P(1-P)/F'(F^{-1}(\tau))$ is continuous and positive on $(0,1)$ and tends to $+\infty$ at both ends, because $P(1-P)$ decays like $e^{-d|z|}$ while $F'(z)$ decays like a Gaussian density, so it is bounded below by a positive constant. (b) follows by solving the affine equation for $z$. (c) $z_p$ is affine in $\operatorname{logit} p$, with $\operatorname{logit} p^\ast = \ln \bar r_{\text{tail}}$ and $\operatorname{logit}\frac12 = 0$.
\end{proof}
\end{revblock}

\FloatBarrier
\section{Proofs}\label{sec:proofs}

We now present the complete proofs of the propositions and theorems from the main manuscript\rev{, and of the corollaries stated in Section~\ref{sec:suptheory}}.

\begin{revblock}
\FloatBarrier
\subsection{Proof of Theorem \ref{thm:optimal_threshold}}

\begin{proof}
\textnormal{Uniqueness} and $\tau^{\ast} \in (0,1)$ follow from Lemma~\ref{lem:convexity}. The function $\Psi_q(\tau) = \sum_j q_j S_j(\tau)$ is continuous, convex in $\tau$ on the compact interval $[0,1]$ and linear in $q$ on the compact convex set $\mathcal{Q}_\beta$. By Sion's minimax theorem \citepec{sion1958minimax}, $\min_\tau \max_q \Psi_q(\tau) = \max_q \min_\tau \Psi_q(\tau)$, and both extrema are attained. Let $q^\ast \in \arg\max_q \min_\tau \Psi_q(\tau)$ and let $\tau^{\ast} = \arg\min_\tau \max_q \Psi_q(\tau)$, which is the CVaR minimizer by Equation~\eqref{eq:envelope}. Then
\[
\Psi_{q^\ast}(\tau^{\ast}) \leq \max_q \Psi_q(\tau^{\ast}) = \min_\tau \max_q \Psi_q(\tau) = \max_q \min_\tau \Psi_q(\tau) = \min_\tau \Psi_{q^\ast}(\tau) \leq \Psi_{q^\ast}(\tau^{\ast}),
\]
so all terms are equal and $(\tau^{\ast}, q^\ast)$ is a saddle point. Hence $q^\ast$ maximizes the linear program of Equation~\eqref{eq:envelope} at $\tau^{\ast}$, which gives it the stated tail structure, and $\tau^{\ast}$ minimizes $\Psi_{q^\ast}(\tau) = K_{q^\ast}\mathbb{P}_{\mathrm{FP}}(\tau) + L_{q^\ast}\mathbb{P}_{\mathrm{FN}}(\tau)$, a strictly convex function (Lemma~\ref{lem:convexity}) whose derivative $(K_{q^\ast} + L_{q^\ast})\mathbb{P}(Y=1|\tau) - K_{q^\ast}$ is negative at $0$ and positive at $1$. Its minimizer is the unique root of this derivative, which is \eqref{eq:foc}; $p^\ast \in (0,1)$ since $K_{q^\ast}, L_{q^\ast} > 0$, and $\mathbb{P}(Y=1|\cdot)$ is a continuous strictly increasing bijection of $[0,1]$ (Lemmas~\ref{lem:boundary} and \ref{lem:monotonicity}).

The closed forms follow by inverting Equation~\eqref{eq:trig-model} on each branch. On $[0,\alpha]$, $\mathbb{P}(Y=1|\tau) = \frac{1-\alpha}{\sin k}[\sin(\frac{k\tau}{\alpha} - k) + \sin k]$, so $\mathbb{P}(Y=1|\tau) = p^\ast$ reads $\sin(\frac{k\tau}{\alpha} - k) = (\frac{p^\ast}{1-\alpha} - 1)\sin k$; the right-hand side lies in $[-1, 0]$ exactly when $p^\ast \leq 1-\alpha$, and since $\frac{k\tau}{\alpha} - k \in [-k, 0] \subset [-\pi/2, 0]$ the principal branch of the arcsine applies, giving case 1. On $(\alpha, 1]$, $\mathbb{P}(Y=1|\tau) = 1 - \alpha + f(\tau)$ with $f(\tau) = \frac{\alpha}{k}[\arcsin(\frac{(\tau-1)\sin k}{1-\alpha}) + k]$, and $\mathbb{P}(Y=1|\tau) = p^\ast > 1-\alpha$ gives $\arcsin(\frac{(\tau-1)\sin k}{1-\alpha}) = -\frac{k}{\alpha}(1-p^\ast)$, whose right-hand side lies in $(-k, 0) \subset (-\pi/2, 0)$ because $1-p^\ast < \alpha$; solving for $\tau$ gives case 2. The two cases meet at $p^\ast = 1-\alpha$, where both give $\tau^{\ast} = \alpha$ (Lemma~\ref{lem:boundary}).
\end{proof}

\FloatBarrier
\subsection{Proof of Proposition \ref{prop:cost_monotonicity}}

\begin{proof}
\textnormal{Write} $\bar r$ for $\bar{r}_{\text{tail}}$. Since $\bar r \mapsto \bar r/(1+\bar r)$ is strictly increasing onto $(0,1)$ and $\mathbb{P}(Y=1|\cdot)$ is a continuous strictly increasing bijection, $\tau^{\ast}(\bar r) = \mathbb{P}(Y=1|\cdot)^{-1}(\bar r/(1+\bar r))$ is continuous and strictly increasing. Explicitly, in case 1 of Theorem~\ref{thm:optimal_threshold}, $\tau^{\ast}(\bar r) = \alpha + \frac{\alpha}{k}\arcsin(\Theta(\bar r)\sin k)$ with $\Theta(\bar r) = \frac{\bar r}{(1-\alpha)(1+\bar r)} - 1$, so
\[
\frac{d\tau^{\ast}}{d\bar r} = \frac{\alpha}{k}\cdot\frac{\sin k}{\sqrt{1 - \Theta(\bar r)^2\sin^2 k}}\cdot\frac{1}{(1-\alpha)(1+\bar r)^2} > 0,
\]
and in case 2, $\tau^{\ast}(\bar r) = 1 - (1-\alpha)\frac{\sin(B(\bar r))}{\sin k}$ with $B(\bar r) = \frac{k}{\alpha(1+\bar r)} \in (0, k)$, so
\[
\frac{d\tau^{\ast}}{d\bar r} = \frac{(1-\alpha)k\cos(B(\bar r))}{\alpha\sin k\,(1+\bar r)^2} > 0 .
\]
The boundary $p^\ast = 1-\alpha$ corresponds to $\bar r = (1-\alpha)/\alpha$, where both expressions give $\tau^{\ast} = \alpha$.
\end{proof}

\FloatBarrier
\subsection{Proof of Proposition \ref{prop:monotonicity_beta}}

\begin{proof}
\textnormal{Under} the strict separation of the $n$-th, $(n+1)$-th and $(n+2)$-th largest losses at $\tau^{\ast}$, the ordering of the losses at these ranks is preserved in a neighborhood of $(\tau^{\ast}, \beta)$ by continuity, so by Lemma~\ref{lem:cvar_technical},
\[
\mathcal{F}(\tau, m) := \partial_\tau\text{CVaR}_\beta(S(\tau)) = \frac{1}{m}\Bigl[\sum_{i=1}^{n}\rho_{(i)}(\tau) + (m-n)\,\rho_{j^{\ast}}(\tau)\Bigr]
\]
is continuously differentiable there (one-sided if $\tau^{\ast} = \alpha$), where $\rho_{(i)}$ denotes the sensitivity of the scenario with the $i$-th largest loss. At the optimum $\mathcal{F}(\tau^{\ast}, m) = 0$ and $\partial_\tau \mathcal{F} = \partial_{\tau\tau}\text{CVaR}_\beta > 0$ by Lemma~\ref{lem:cvar_technical}(2). Differentiating in $m$,
\[
\partial_m \mathcal{F} = \frac{\rho_{j^{\ast}}(\tau)}{m} - \frac{1}{m^2}\Bigl[\sum_{i \leq n}\rho_{(i)}(\tau) + (m-n)\rho_{j^{\ast}}(\tau)\Bigr] = \frac{\rho_{j^{\ast}}(\tau) - \mathcal{F}(\tau, m)}{m},
\]
so $\partial_m \mathcal{F}|_{\tau^{\ast}} = \rho_{j^{\ast}}(\tau^{\ast})/m$. Since $m = (1-\beta)J$, $\partial_\beta \mathcal{F}|_{\tau^{\ast}} = -J\,\partial_m \mathcal{F}|_{\tau^{\ast}} = -\rho_{j^{\ast}}(\tau^{\ast})/(1-\beta)$, and the implicit function theorem yields
\[
\frac{d\tau^{\ast}}{d\beta} = -\frac{\partial_\beta \mathcal{F}}{\partial_\tau \mathcal{F}} = \frac{\rho_{j^{\ast}}(\tau^{\ast})}{(1-\beta)\,\partial_{\tau\tau}\text{CVaR}_\beta(S(\tau^{\ast}))},
\]
whose sign is that of $\rho_{j^{\ast}}(\tau^{\ast})$. If $m$ is an integer, a small increase of $\beta$ makes $n = m-1$ with the $m$-th largest scenario as marginal scenario, and a small decrease makes $n = m$ with the $(m+1)$-th largest; applying the argument on each side gives the one-sided statement.
\end{proof}

\end{revblock}

\FloatBarrier
\subsection{Proof of Lemma~\ref{lem:accuracy_threshold}}
\begin{proof}
\rev{\textnormal{The} derivative $\frac{d\text{Acc}(\tau)}{d\tau} = 1 - 2\mathbb{P}(Y=1|\tau)$ is strictly decreasing by Lemma~\ref{lem:monotonicity}, so accuracy is strictly concave and its maximizer is the unique root of $\mathbb{P}(Y=1|\tau_{\mathrm{acc}}) = \frac{1}{2}$.} The specific solutions follow from inverting the trigonometric model at this level.

\textit{Case} $\alpha < \frac{1}{2}$: Since $\mathbb{P}(Y=1|\alpha) = 1-\alpha > \frac{1}{2}$ and $\mathbb{P}(Y=1|\cdot)$ is increasing, we need $\tau_{\mathrm{acc}} < \alpha$. Using the first branch of the trigonometric model for $\tau \leq \alpha$:
$$\mathbb{P}(Y=1|\tau_{\mathrm{acc}})=\frac{1-\alpha}{\sin k}\left[\sin\left(\frac{k\tau_{\mathrm{acc}}}{\alpha} - k\right) + \sin k\right] = \frac{1}{2}$$
We solve $\sin\left(\frac{k\tau_{\mathrm{acc}}}{\alpha} - k\right) = \sin k\left(\frac{2\alpha - 1}{2(1-\alpha)}\right)$, which yields
$$\tau_{\mathrm{acc}} = \alpha + \frac{\alpha}{k}\arcsin\left(\sin k \cdot \frac{2\alpha - 1}{2(1-\alpha)}\right)$$
Since $\alpha < \frac{1}{2}$ implies $|2\alpha - 1|/(2(1-\alpha)) < 1$, the arcsin is well-defined and $\tau_{\mathrm{acc}} \in (0, \alpha)$ as required.

\textit{Case} $\alpha = \frac{1}{2}$: We have $\mathbb{P}(Y=1|\alpha) = \frac{1}{2}$, so $\tau_{\mathrm{acc}} = \alpha$.

\textit{Case} $\alpha > \frac{1}{2}$: Since $\mathbb{P}(Y=1|\alpha) = 1-\alpha < \frac{1}{2}$, we need $\tau_{\mathrm{acc}} > \alpha$. Using the second branch of the trigonometric model for $\tau>\alpha$:
$$\mathbb{P}(Y=1|\tau_{\mathrm{acc}}) = (1-\alpha) + \frac{\alpha}{k}\left[\arcsin\left(\frac{(\tau_{\mathrm{acc}}-1)\sin k}{1-\alpha}\right) + k\right] = \frac{1}{2}$$
This simplifies to $\arcsin\left(\frac{(\tau_{\mathrm{acc}}-1)\sin k}{1-\alpha}\right) = -\frac{k}{2\alpha}$, yielding
$$\tau_{\mathrm{acc}} = 1 - \frac{(1-\alpha)\sin\left(\frac{k}{2\alpha}\right)}{\sin k}$$
Since $0 < k/(2\alpha) < k$ when $\alpha > \frac{1}{2}$, we have $\tau_{\mathrm{acc}} \in (\alpha, 1)$.
\end{proof}

\FloatBarrier
\subsection{Proof of Proposition \ref{prop:efficiency_gap}}

\begin{proof}
\textnormal{Part (i):} Under proportional costs, every scenario loss equals $c_jS_0(\tau)$, fixing $\mathcal{J}_\beta$. 
The first-order condition gives $\mathbb{P}(Y=1|\tau^{\ast}) = \frac{\sum_i K_i}{\sum_i(K_i+L_i)}$. 
This equals $\frac{1}{2}$ if and only if $\sum_i K_i = \sum_i L_i$, forcing $\tau^{\ast} = \tau_{\mathrm{acc}}$; otherwise they differ.

\smallskip
Part (ii): The CVaR and accuracy conditions yield the probability gap 
$\Delta_p = \left|\frac{\bar{r}_{\text{tail}}}{1+\bar{r}_{\text{tail}}} - \frac{1}{2}\right|$.
\rev{Since $\mathbb{P}(Y=1|\cdot)$ is absolutely continuous with $p_{\min} \leq \partial_\tau\mathbb{P}(Y=1|\tau) \leq p_{\max}$ at every $\tau \neq \alpha$, integrating between the two thresholds gives $p_{\min}\Delta_\tau \leq \Delta_p \leq p_{\max}\Delta_\tau$, i.e., $\Delta_\tau \in [\Delta_p/p_{\max}, \Delta_p/p_{\min}] = [C_1\Delta_p, C_2\Delta_p]$. The bounds $p_{\min}$ and $p_{\max}$ are the extreme values, over both branches, of the derivative}
\[
\partial_\tau\mathbb{P}(Y=1\mid\tau)=
\begin{cases}
\displaystyle
  \frac{(1-\alpha)k}{\alpha\sin k}\cos\!\bigl(\tfrac{k\tau}{\alpha}-k\bigr),
  &\tau\in(0,\alpha),\\[6pt]
\displaystyle
  \frac{\alpha\sin k}{k(1-\alpha)}
        \frac{1}{\sqrt{1-\bigl(\frac{(\tau-1)\sin k}{1-\alpha}\bigr)^2}},
  &\tau\in(\alpha,1).
\end{cases}
\]
\end{proof}

\FloatBarrier
\subsection{Proof of Proposition \ref{prop:risk_penalty}}

\begin{proof}
\textnormal{Part (i):} Under proportional costs $S_j(\tau) = c_j S_0(\tau)$, the tail set $\mathcal{J}_\beta$ is fixed consisting of the indices with the largest $c_j$ values, regardless of the threshold. Thus $\text{CVaR}_\beta(S(\tau)) = \bar{c}_{\text{tail}} S_0(\tau)$ where $\bar{c}_{\text{tail}} = \frac{1}{(1-\beta)J}\sum_{j \in \mathcal{J}_\beta} c_j$ is the average tail scaling factor, yielding:
$$\Delta_{\text{CVaR}} = \frac{S_0(\tau_{\mathrm{acc}}) - S_0(\tau^{\ast})}{S_0(\tau^{\ast})}$$
When $\sum_i K_i = \sum_i L_i$, we have $\tau^{\ast} = \tau_{\mathrm{acc}}$ by Lemma~\ref{lem:accuracy_threshold} and Lemma~\ref{lem:accuracy_risk_divergence}, hence $\Delta_{\text{CVaR}} = 0$.

\rev{Part (ii): Write $\tau_{\mathrm{acc}} = \tau^{\ast} + \omega$ with $|\omega| = \Delta_\tau$. By Lemma~\ref{lem:differentiability}, $\partial_\tau S_j$ is absolutely continuous, so for every $j$
\[
S_j(\tau_{\mathrm{acc}}) = S_j(\tau^{\ast}) + \rho_j(\tau^{\ast})\,\omega + R_j, \qquad R_j := \int_{\tau^{\ast}}^{\tau_{\mathrm{acc}}} (\tau_{\mathrm{acc}} - t)\,\partial_{\tau\tau}S_j(t)\,dt,
\]
and $\frac{\tilde m_j}{2}\omega^2 \leq R_j \leq \frac{\tilde M_j}{2}\omega^2$ with $\tilde m_j, \tilde M_j$ the essential bounds of $\partial_{\tau\tau}S_j$ on $I$.

\emph{Lower bound.} Since $q^\ast \in \mathcal{Q}_\beta$ is feasible in \eqref{eq:envelope} at $\tau_{\mathrm{acc}}$,
\begin{align*}
\text{CVaR}_\beta(S(\tau_{\mathrm{acc}})) &\geq \sum_j q^\ast_j S_j(\tau_{\mathrm{acc}}) = \text{CVaR}_\beta(S(\tau^{\ast})) + \omega\sum_j q^\ast_j \rho_j(\tau^{\ast}) + \sum_j q^\ast_j R_j \\
&\geq \text{CVaR}_\beta(S(\tau^{\ast})) + \frac{\tilde m}{2}(\Delta_\tau)^2,
\end{align*}
where $\sum_j q^\ast_j S_j(\tau^{\ast}) = \text{CVaR}_\beta(S(\tau^{\ast}))$ by the saddle-point property of Theorem~\ref{thm:optimal_threshold}, the first-order term vanishes because $\tau^{\ast}$ is the interior minimizer of $\sum_j q^\ast_j S_j$, and $\sum_j q^\ast_j R_j \geq \frac{\tilde m}{2}\omega^2$ since $q^\ast$ is supported on scenarios with $\tilde m_j \geq \tilde m$.

\emph{Upper bound.} Since $q_{\mathrm{acc}}$ attains the maximum in \eqref{eq:envelope} at $\tau_{\mathrm{acc}}$ and is feasible at $\tau^{\ast}$,
\begin{align*}
\text{CVaR}_\beta(S(\tau_{\mathrm{acc}})) &= \sum_j q_{\mathrm{acc},j} S_j(\tau_{\mathrm{acc}}) \\
&\leq \text{CVaR}_\beta(S(\tau^{\ast})) + \omega\sum_j q_{\mathrm{acc},j}\rho_j(\tau^{\ast}) \\
&\quad + \frac{\tilde M}{2}(\Delta_\tau)^2,
\end{align*}
using $\sum_j q_{\mathrm{acc},j} S_j(\tau^{\ast}) \leq \text{CVaR}_\beta(S(\tau^{\ast}))$ and $R_j \leq \frac{\tilde M}{2}\omega^2$ on the support of $q_{\mathrm{acc}}$. Bounding $\omega\sum_j q_{\mathrm{acc},j}\rho_j(\tau^{\ast})$ by $\Delta_\tau|\sum_j q_{\mathrm{acc},j}\rho_j(\tau^{\ast})|$ and dividing both bounds by $\text{CVaR}_\beta(S(\tau^{\ast})) > 0$ yields the claim. When $q_{\mathrm{acc}}$ can be taken equal to $q^\ast$, the first-order term is zero.}
\end{proof}

\FloatBarrier
\subsection{Proof of Theorem \ref{thm:insurance_value}}

\begin{proof}
\textnormal{The} value of insurance equals the reduction it allows in required revenue. Without insurance the necessary condition of profitability is $R \geq \mathbb{E}[S(\tau_\theta)] + \rho\gamma \text{CVaR}_\beta[S(\tau_\theta)]$, while with insurance it becomes $R \geq \mathbb{E}[S_{\text{net}}(\tau^{\ast})] + \Pi + \rho\gamma_{\text{ins}} \text{CVaR}_\beta[S_{\text{net}}(\tau^{\ast})]$.

Since $S_{\text{net}} = S - C_{\text{tot}}$, we have $\mathbb{E}[S_{\text{net}}] = \mathbb{E}[S] - \mathbb{E}[C_{\text{tot}}]$ and the value difference is:
\begin{align}
V_{\theta} &= \mathbb{E}[S(\tau_\theta)] - \mathbb{E}[S(\tau^{\ast})] + \mathbb{E}[C_{\text{tot}}(\tau^{\ast})] - \Pi\\
&\quad + \rho\gamma \text{CVaR}_\beta[S(\tau_\theta)] - \rho\gamma_{\text{ins}} \text{CVaR}_\beta[S_{\text{net}}(\tau^{\ast})]\nonumber
\end{align}

Adding and subtracting $\rho\gamma \text{CVaR}_\beta[S(\tau^{\ast})]$ and grouping terms:
\begin{align}
V_{\theta} &= \underbrace{\mathbb{E}[S(\tau_\theta) - S(\tau^{\ast})] + \rho\gamma[\text{CVaR}_\beta[S(\tau_\theta)] - \text{CVaR}_\beta[S(\tau^{\ast})]]}_{\text{Calibration value } V_{\text{calib}}^{\theta}}\\
&\quad + \underbrace{\mathbb{E}[C_{\text{tot}}(\tau^{\ast})] - \Pi + \rho[\gamma \text{CVaR}_\beta[S(\tau^{\ast})] - \gamma_{\text{ins}} \text{CVaR}_\beta[S_{\text{net}}(\tau^{\ast})]]}_{\text{Base insurance value } V_{\mathrm{base}}}\nonumber.
\end{align}

\end{proof}

\FloatBarrier
\subsection{Proof of Proposition \ref{prop:limit_behavior}}

\begin{revblock}
\begin{proof}
\textnormal{Part 1}: As $L_{\text{occ}}, L_{\text{agg}} \to 0$, we have $C_{\text{tot}} \to 0$, hence $\Pi \to 0$ and $S_{\text{net}} \to S$. Thus:
$$V_{\mathrm{base}} \to \underbrace{0}_{\text{Risk transfer} \to 0} + \underbrace{\rho(\gamma - \gamma_{\text{ins}}) \text{CVaR}_\beta[S(\tau^{\ast})]}_{\text{Capital relief only}}$$

Part 2: As $L_{\text{occ}}, L_{\text{agg}} \to \infty$, the insurer pays the whole liability component, $C_{\text{tot}} \to S_L(\tau^{\ast})$, and the retained loss $S_{\text{net}} \to s_{\mathrm{op}}$, a constant, so $\text{CVaR}_\beta[S_{\text{net}}] \to s_{\mathrm{op}}$; moreover $\text{CVaR}_\beta[S(\tau^{\ast})] = s_{\mathrm{op}} + \text{CVaR}_\beta[S_L(\tau^{\ast})]$ by translation equivariance. Substituting the premium \eqref{eq:premium},
$$V_{\mathrm{base}} \to \mathbb{E}[S_L] - (1+\mu)\mathbb{E}[S_L] - r\bigl(\text{CVaR}_\beta[S_L] - \mathbb{E}[S_L]\bigr) + \rho\gamma\bigl(s_{\mathrm{op}} + \text{CVaR}_\beta[S_L]\bigr) - \rho\gamma_{\text{ins}}\,s_{\mathrm{op}},$$
which rearranges to $(\rho\gamma - r)\,\text{CVaR}_\beta[S_L] - (\mu - r)\,\mathbb{E}[S_L] + \rho(\gamma - \gamma_{\text{ins}})\,s_{\mathrm{op}}$, all quantities at $\tau^{\ast}$.
\end{proof}

\FloatBarrier
\subsection{Proof of Corollary \ref{cor:viability}}

\begin{proof}
\textnormal{The} first claim restates the full-coverage limit of Proposition~\ref{prop:limit_behavior}(2). Writing $(\rho\gamma - r)\,\text{CVaR}_\beta[S_L] - (\mu - r)\,\mathbb{E}[S_L] = (\rho\gamma - r)(\text{CVaR}_\beta[S_L] - \mathbb{E}[S_L]) + (\rho\gamma - \mu)\,\mathbb{E}[S_L]$ shows the limit as a sum of three terms, each non-negative when $\rho\gamma \geq \max\{\mu, r\}$ (since $\text{CVaR}_\beta \geq \mathbb{E}$, $\mathbb{E}[S_L] > 0$, $\gamma \geq \gamma_{\text{ins}}$ and $s_{\mathrm{op}} \geq 0$), and positive under the stated conditions. The last form divides the condition by $\mathbb{E}[S_L] > 0$ with $s_{\mathrm{op}} = 0$ and $\rho\gamma - r > 0$.
\end{proof}
\end{revblock}

\FloatBarrier
\subsection{Proof of Proposition \ref{prop:optimal_duration_full}}

\begin{revblock}
\begin{proof}
\textnormal{The} average cost $\mathcal{A}(T) = (\mathcal{C}_a + \int_0^T e)/T$ is differentiable on $(0, \infty)$ with
\[
\mathcal{A}'(T) = \frac{T e(T) - \mathcal{C}_a - \int_0^T e(t)\,dt}{T^2} = \frac{H(T) - \mathcal{C}_a}{T^2}.
\]
For $T_2 > T_1$, $H(T_2) - H(T_1) = \int_{T_1}^{T_2}[e(T_2) - e(t)]\,dt + T_1[e(T_2) - e(T_1)] \geq 0$ because $e$ is non-decreasing, so $H$ is non-decreasing, and strictly increasing if $e$ is; $H(0) = 0$.

\emph{(i)} $\mathcal{A}$ is decreasing on $\{H < \mathcal{C}_a\}$ and increasing on $\{H > \mathcal{C}_a\}$. If $\mathcal{C}_a < H_\infty$, the level set $\{H = \mathcal{C}_a\}$ is a non-empty bounded interval (a single point when $H$ is strictly increasing), and $\mathcal{A}$ attains its minimum exactly there, which is Equation~\eqref{eq:F0_condition_full}. If $\mathcal{C}_a > H_\infty$, then $\mathcal{A}' < 0$ everywhere, and if $\mathcal{C}_a = H_\infty$, then $\mathcal{A}' \leq 0$ with equality exactly where $H = H_\infty$.

\emph{(ii)} The root of $H(T) = \mathcal{C}_a$ moves to the right as $\mathcal{C}_a$ increases because $H$ is non-decreasing.

\emph{(iii)} $H_1(T_2^{\ast}) - H_2(T_2^{\ast}) = \int_0^{T_2^{\ast}}\bigl[(e_1 - e_2)(T_2^{\ast}) - (e_1 - e_2)(t)\bigr]dt \geq 0$ since $e_1 - e_2$ is non-decreasing on $[0, T_2^{\ast}]$, so $H_1(T_2^{\ast}) \geq H_2(T_2^{\ast}) = \mathcal{C}_a$; as $H_1$ is non-decreasing, its first crossing of $\mathcal{C}_a$, the smallest minimizer $T_1^{\ast}$, is not to the right of $T_2^{\ast}$.
\end{proof}

\FloatBarrier
\subsection{Proof of Corollary \ref{cor:sqrt}}
\begin{proof}
\textnormal{With} $e(t) = st$, $H(T) = sT^2 - sT^2/2 = sT^2/2$, and $H(T^{\ast}) = \mathcal{C}_a$ gives $T^{\ast} = \sqrt{2\mathcal{C}_a/s}$.
\end{proof}

\FloatBarrier
\subsection{Proof of Corollary \ref{cor:recalibration}}
\begin{proof}
\textnormal{By} Lemma~\ref{lem:cvar_drift}, $\text{CVaR}_\beta(S(\tau^{\ast}_t, t)) > \text{CVaR}_\beta(S(\tau^{\ast}_t, 0)) \geq \text{CVaR}_\beta(S(\tau^{\ast}, 0))$, since $\tau^{\ast}_t \in (0,1)$ by Lemma~\ref{lem:convexity} and $\tau^{\ast}$ minimizes the inception risk; $e_{\mathrm{rc}} \leq e$ because $\tau^{\ast}_t$ minimizes $\text{CVaR}_\beta(S(\cdot, t))$. The last claim follows from Proposition~\ref{prop:optimal_duration_full}(iii) with $e_1 = e$ and $e_2 = e_{\mathrm{rc}}$.
\end{proof}
\end{revblock}

\FloatBarrier
\subsection{Proof of Proposition \ref{prop:optimal_zeta}}

\begin{revblock}
\begin{proof}
\textnormal{The} insured firm's total cost with interpretability level $\zeta$ is
$TC(\zeta) = \Pi(\zeta) + c(\zeta) = A(1 - \xi g(\zeta)) + c(\zeta)$, with $A = \Pi(\tau^{\ast})$, since the premium scales by $1 - \xi g(\zeta)$. $TC$ is continuous on $[0,1]$, since $c$ and $g$ are (Section~\ref{subsec:interpretability}), and differentiable on $(0,1)$. Its derivative is
$$\frac{dTC}{d\zeta} = -A\xi g'(\zeta) + c'(\zeta) = g'(\zeta)\bigl[\psi(\zeta) - A\xi\bigr],$$
with $g' > 0$. Since $\psi$ is strictly increasing, $\{\psi < A\xi\}$ is an initial interval of $(0,1)$. On it $TC$ is strictly decreasing, and on $\{\psi > A\xi\}$ it is strictly increasing, so $TC$ is unimodal and its minimizer is unique. The minimizer equals $0$ if $A\xi \leq \psi(0^+)$, equals $1$ if $A\xi \geq \psi(1^-)$, and otherwise is the unique root of $\psi(\zeta) = A\xi$, which is the first-order condition $c'(\zeta^\ast) = A\xi g'(\zeta^\ast)$. If $c$ is convex and $g$ concave, $c'$ is non-decreasing and $g'$ positive and non-increasing, and both are continuous as derivatives of differentiable convex or concave functions, so $\psi$ is continuous and non-decreasing. It is strictly increasing if $c$ is strictly convex. It is also strictly increasing if $g$ is strictly concave, provided $c' > 0$ on $(0,1)$, which holds because a non-decreasing $c'$ that vanished at some point would vanish on an initial interval, on which $c$ would be constant, contradicting that $c$ is increasing. For the power family, $\psi(\zeta) = \frac{c_0 a}{b}\zeta^{a-b}$ is strictly increasing with $\psi(0^+) = 0$ and $\psi(1^-) = c_0a/b$, and solving $\psi(\zeta) = A\xi$ gives Equation~\eqref{eq:zetastar}.
\end{proof}
\end{revblock}

\FloatBarrier
\section{Numerical Validation of Trigonometric Model}\label{app:trigonometric_validation}

\begin{revblock}
To validate the trigonometric model, we examine how well the proposed functional form fits the empirical mapping from classifier scores to true class probabilities, on clinical data and on synthetic data, and the tail-risk cost of using it under misspecification.

\paragraph{Data.} We use four clinical datasets. The Breast Cancer Wisconsin (Diagnostic) data \citepec{wolberg1993wisconsin} comprise $n = 569$ cases, 37\% malignant, described by features of fine-needle aspirates of suspicious masses. The Mammographic Mass data \citepec{elter2007prediction} comprise $n = 830$ complete cases, 49\% malignant, described by the BI-RADS assessment, age, and the shape, margin and density of the mass, and are the closest analogue of the diagnostic setting. The Pima Indians Diabetes data \citepec{smith1988adap} comprise $n = 768$ cases, 35\% positive. The Thyroid disease data \citepec{quinlan1986thyroid}, in their neural-network version, comprise $n = 3{,}772$ cases, 7.5\% positive, where a positive case is a hyperfunction or a subnormal function. The Wisconsin and Mammographic Mass data are taken from the UCI repository \citepec{kelly2023uci} and the Pima and Thyroid data from OpenML (datasets 37 and 40497) \citepec{vanschoren2013openml}. To probe class imbalance we also subsample the positives of the first three to a prevalence of 15\% and of the last to 5\% and 1\%. The synthetic datasets use scikit-learn's \texttt{make\_classification} with $n = 5{,}000$, class-0 shares $\alpha \in \{0.01, 0.05, 0.15, 0.25, 0.40, 0.50, 0.60, 0.75, 0.85, 0.95, 0.99\}$ and the four complexity configurations of Table~\ref{tab:dataset_configs}.

\begin{table}[htbp]
\centering
\tiny
\caption{Complexity configurations of the synthetic datasets.}
\label{tab:dataset_configs}
\begin{tabular}{lccccc}
\toprule
Complexity & Features & Informative & Clusters/Class & Class Sep. & Label Noise \\
\midrule
Linear & 20 & 18 & 1 & 2.0 & 0.01 \\
Moderate & 20 & 15 & 2 & 1.0 & 0.05 \\
Complex & 20 & 10 & 3 & 0.5 & 0.10 \\
High-dim & 100 & 75 & 2 & 1.0 & 0.05 \\
\bottomrule
\end{tabular}
\end{table}

\paragraph{Protocol.} For each dataset we train six classifiers in scikit-learn \citepec{pedregosa2011scikit}, namely logistic regression with L2 and L1 penalties \citepec{hastie2009elements}, random forests with unlimited and with maximum depth 20 \citepec{breiman2001random}, and gradient boosting with default and with maximum depth 10 \citepec{friedman2001greedy}, in repeated stratified five-fold cross-validation (four repeats) and keep only out-of-fold predicted probabilities. Since the model assumes a uniform score, scores are rank-transformed, $\overline{\gamma} = \text{rank}/n$, and the empirical calibration curve is the fraction of positives in ten uniform bins of $\overline{\gamma}$. The parameter $k$ is fitted by least squares, with $\alpha$ fixed to the class-0 share, on the curve of the first two repeats and evaluated on the curve of the last two repeats, which re-partition the same cases and which we refer to as the held-out curve. The fit is measured by $R^2 = 1 - \sum_i(y_i - \hat{y}_i)^2/\sum_i(y_i - \bar{y})^2$, where $y_i$ are the empirical fractions and $\hat{y}_i$ the model predictions. The equal-variance binormal map of Section~\ref{sec:binormal} (parameter $d$) and an isotonic regression \citepec{zadrozny2002transforming} are fitted and scored in the same way. We also record the empirical AUC of the out-of-fold scores, and the value of $k$ it implies when the AUC lies within the range attained by the model. Finally, taking the isotonic map as the truth and the diagnostic cost structure of Section~\ref{sec:casestudy}, we compute the CVaR-optimal threshold under each map and measure the penalty of using the trigonometric or the binormal threshold by $\text{CVaR}_\beta(S(\tau^{\ast}_{\text{fit}}))/\text{CVaR}_\beta(S(\tau^{\ast}_{\text{iso}})) - 1$.

\begin{table}[htbp]
\centering
\tiny
\setlength{\tabcolsep}{2pt}
\caption{Fit of the calibration maps and misspecification cost of the fitted thresholds.}
\label{tab:trigonometric_validation}
\begin{tabular}{lrrrrrrrrr}
\toprule
 & & \multicolumn{3}{c}{\textbf{Trigonometric $R^2$}} & \textbf{Binormal} & \textbf{Isotonic} & \textbf{Gap in} & \multicolumn{2}{c}{\textbf{CVaR penalty}} \\
\cmidrule(lr){3-5} \cmidrule(lr){9-10}
\textbf{Category} & \textbf{Fits} & \textbf{med.} & \textbf{IQR} & \textbf{$>0.8$} & \textbf{$R^2$ med.} & \textbf{$R^2$ med.} & \textbf{$\tau^{\ast}$ med.} & \textbf{trig.\ med./max} & \textbf{bin.\ med.} \\ \midrule
\textbf{All clinical} & 54 & 0.882 & 0.229 & 72\% & 0.986 & 0.976 & 0.283 & 53\% / 9,801\% & 16\% \\
\quad Wisconsin & 6 & 0.879 & 0.024 & 100\% & 0.999 & 0.998 & 0.288 & 69\% / 222\% & 10\% \\
\quad Wisconsin, 15\% & 6 & 0.872 & 0.010 & 100\% & 0.998 & 0.957 & 0.406 & 52\% / 140\% & 1\% \\
\quad Mammographic Mass & 6 & 0.969 & 0.011 & 100\% & 0.975 & 0.989 & 0.066 & 11\% / 34\% & 11\% \\
\quad Mammographic Mass, 15\% & 6 & 0.716 & 0.175 & 33\% & 0.926 & 0.979 & 0.382 & 53\% / 54\% & 19\% \\
\quad Pima & 6 & 0.962 & 0.017 & 100\% & 0.986 & 0.982 & 0.039 & 2\% / 12\% & 3\% \\
\quad Pima, 15\% & 6 & 0.887 & 0.051 & 83\% & 0.970 & 0.949 & 0.075 & 12\% / 32\% & 9\% \\
\quad Thyroid & 6 & 0.912 & 0.031 & 100\% & 0.996 & 0.901 & 0.458 & 2,056\% / 9,801\% & 422\% \\
\quad Thyroid, 5\% & 6 & 0.272 & 0.582 & 17\% & 0.981 & 0.816 & 0.360 & 1,760\% / 7,922\% & 139\% \\
\quad Thyroid, 1\% & 6 & 0.029 & 0.526 & 17\% & 0.573 & 0.088 & 0.017 & 165\% / 246\% & 107\% \\
\quad Logistic (L2) & 9 & 0.878 & 0.081 & 78\% & 0.991 & 0.945 & 0.281 & 54\% / 166\% & 17\% \\
\quad Logistic (L1) & 9 & 0.880 & 0.073 & 78\% & 0.994 & 0.983 & 0.285 & 97\% / 242\% & 31\% \\
\quad Random forest & 9 & 0.870 & 0.269 & 67\% & 0.971 & 0.975 & 0.288 & 54\% / 9,801\% & 7\% \\
\quad Random forest (depth 20) & 9 & 0.870 & 0.268 & 67\% & 0.969 & 0.975 & 0.288 & 54\% / 9,742\% & 7\% \\
\quad Gradient boosting & 9 & 0.884 & 0.212 & 67\% & 0.988 & 0.979 & 0.274 & 58\% / 2,615\% & 11\% \\
\quad Gradient boosting (depth 10) & 9 & 0.894 & 0.061 & 78\% & 0.984 & 0.990 & 0.122 & 12\% / 1,653\% & 17\% \\
\quad prevalence $<$ 5\% & 6 & 0.029 & 0.526 & 17\% & 0.573 & 0.088 & 0.017 & 165\% / 246\% & 107\% \\
\quad 5--20\% & 30 & 0.875 & 0.223 & 67\% & 0.988 & 0.956 & 0.365 & 56\% / 9,801\% & 19\% \\
\quad 20--45\% & 12 & 0.885 & 0.090 & 100\% & 0.993 & 0.990 & 0.079 & 8\% / 222\% & 3\% \\
\quad 45--100\% & 6 & 0.969 & 0.011 & 100\% & 0.975 & 0.989 & 0.066 & 11\% / 34\% & 11\% \\
\textbf{All synthetic} & 264 & 0.878 & 0.868 & 63\% & 0.989 & 0.985 & 0.077 & 17\% / 223\% & 6\% \\
\quad Logistic (L2) & 44 & 0.837 & 0.996 & 57\% & 0.973 & 0.930 & 0.065 & 16\% / 223\% & 6\% \\
\quad Logistic (L1) & 44 & 0.836 & 1.000 & 57\% & 0.973 & 0.928 & 0.064 & 19\% / 223\% & 6\% \\
\quad Random forest & 44 & 0.913 & 0.523 & 64\% & 0.990 & 0.989 & 0.088 & 18\% / 222\% & 8\% \\
\quad Random forest (depth 20) & 44 & 0.912 & 0.549 & 64\% & 0.990 & 0.984 & 0.090 & 15\% / 223\% & 9\% \\
\quad Gradient boosting & 44 & 0.916 & 0.382 & 70\% & 0.992 & 0.993 & 0.053 & 13\% / 209\% & 5\% \\
\quad Gradient boosting (depth 10) & 44 & 0.915 & 0.491 & 68\% & 0.991 & 0.994 & 0.094 & 17\% / 162\% & 6\% \\
\quad prevalence $<$ 5\% & 18 & $-$0.276 & 0.535 & 0\% & 0.632 & 0.517 & 0.515 & 9\% / 37\% & 3\% \\
\quad 5--20\% & 54 & 0.882 & 0.690 & 61\% & 0.970 & 0.896 & 0.226 & 20\% / 222\% & 12\% \\
\quad 20--45\% & 48 & 0.967 & 0.119 & 88\% & 0.997 & 0.998 & 0.112 & 14\% / 222\% & 14\% \\
\quad 45--100\% & 144 & 0.889 & 0.871 & 64\% & 0.987 & 0.990 & 0.018 & 19\% / 223\% & 3\% \\
\midrule
\textbf{Overall} & 318 & 0.881 & 0.727 & 65\% & 0.988 & 0.981 & 0.091 & 21\% / 9,801\% & 8\% \\
\bottomrule
\end{tabular}
\par\smallskip{\raggedright\textit{Notes.} Held-out $R^2$ on the rank scale. Fits counts dataset--classifier pairs and IQR is the interquartile range. The gap in $\tau^{\ast}$ is the median absolute difference between the optimal thresholds under the fitted and the isotonic map, and the CVaR penalty is the cost of using the fitted threshold, both with the isotonic map as the truth and the diagnostic cost structure.\par}
\end{table}

\paragraph{Results.} Table~\ref{tab:trigonometric_validation} reports medians and interquartile ranges by dataset, classifier and realized prevalence. Within the range of discrimination attained by the model and at moderate prevalence, the trigonometric map fits closely. The median held-out $R^2$ is 0.97 on the Mammographic Mass data, whose classifiers reach AUCs of 0.85--0.90 as in our case study, 0.96 on Pima, and 0.88--0.97 on the synthetic datasets at moderate prevalence. The fit deteriorates outside this range. When the classifier's AUC exceeds the model's maximum of about 0.90 (0.89--0.91, depending on $\alpha$, as for the Wisconsin and Thyroid data), $k$ reaches $\pi/2$ and $R^2$ remains at 0.88--0.91, against 0.996--0.999 for the binormal map. At prevalences of 5\% or less, the trigonometric $R^2$ falls below 0.3, because the map equals $1 - \alpha$ at the $\alpha$-quantile (Lemma~\ref{lem:boundary}), whereas the binormal $R^2$ remains above 0.57. Misspecification costs follow the same pattern. With the isotonic map as the truth, the trigonometric threshold raises tail risk by a median of 11\% on the Mammographic Mass data and 2\% on Pima, but by 52--69\% on the Wisconsin data and the 15\%-prevalence Mammographic Mass subsample, and by more on the nearly separable Thyroid data. The binormal threshold raises it by a median of 16\% over all clinical fits. The 15\%-prevalence subsample of the Mammographic Mass data, with about 500 cases, fits less well ($R^2$ of 0.72), and Section~\ref{sec:robustness_ec} checks the diagnostic setting at this prevalence under the binormal model as well. Where the AUC lies within the range attained by the model, the value of $k$ implied by the AUC (Section~\ref{sec:trigonomodel}) is close to the least-squares estimate (medians of 1.18 and 1.33 over the four classifiers in range on the Mammographic Mass data).

These results delimit the trigonometric model. It describes classifier calibration adequately when the AUC lies between about 0.83 and 0.90 and the prevalence is moderate, as in the diagnostic setting of Section~\ref{sec:casestudy}, and the binormal model should be used outside this range, as we do for screening. By Remark~\ref{rem:scope}, the structural results apply to both models, and Section~\ref{sec:binormal} gives the closed forms for the binormal model. Figure~\ref{fig:validation_anchors} shows the fitted calibration curves and the implied and empirical ROC curves for the two clinical anchors.

\begin{figure}[htbp]
    \centering
    \includegraphics[width=0.9\linewidth]{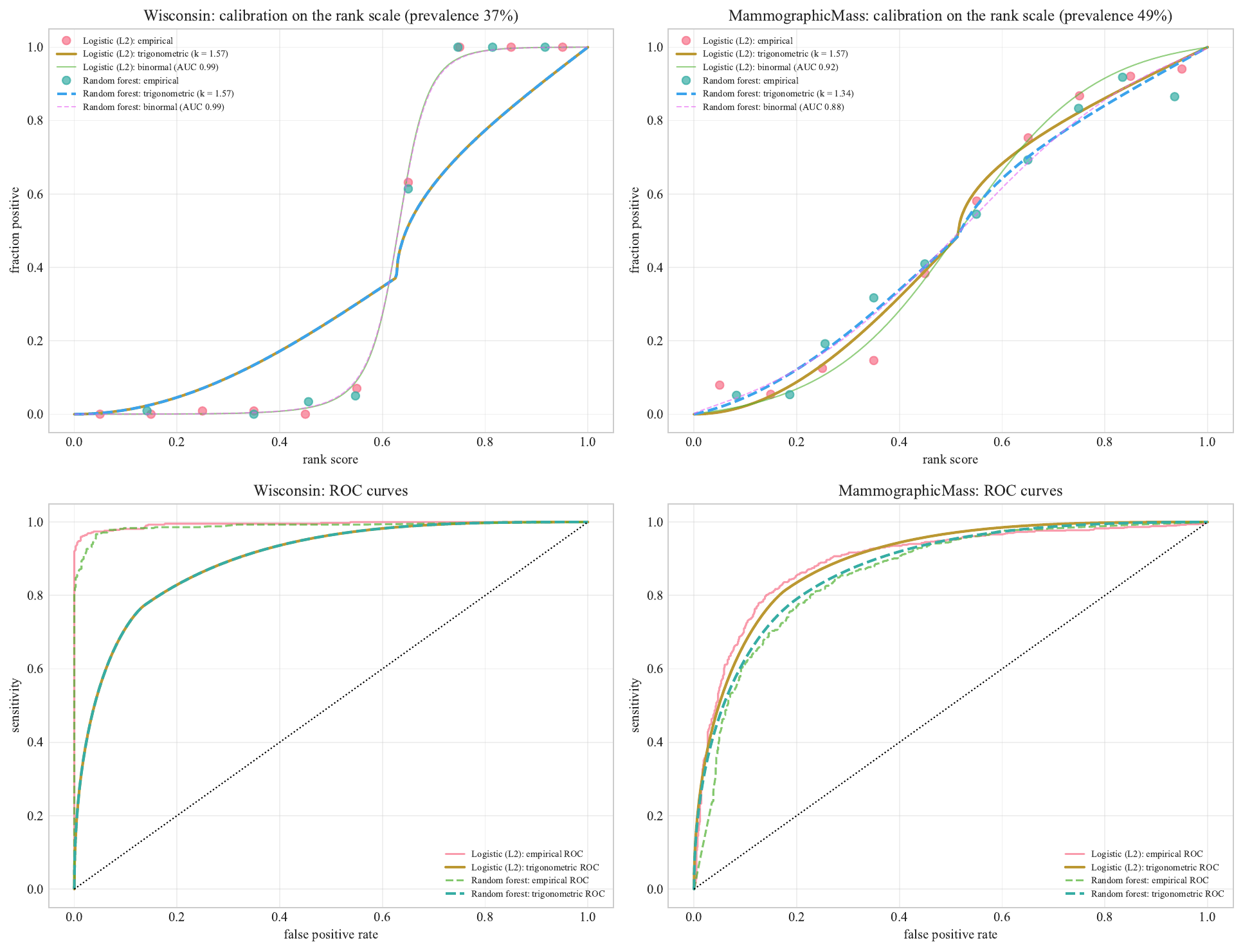}
    \caption{Calibration curves on the rank scale (top) and ROC curves (bottom) for the Wisconsin and Mammographic Mass data, empirical and fitted, for logistic regression and \rev{random forest}.}
    \label{fig:validation_anchors}
\end{figure}
\end{revblock}

\FloatBarrier
\section{Case Study Supplementary Information}\label{sec:scenariolosses}

\begin{revblock}
This section documents the inputs of the case study and their sources (Section~\ref{sec:provenance}), the construction of the scenario losses and of the insurer's payments (Section~\ref{sec:scenario_construction}), the robustness of the calibration results (Section~\ref{sec:robustness_ec}), the sensitivity of insurance value (Section~\ref{sec:insurance_sensitivity_ec}) and of the renewal and oversight results (Section~\ref{sec:renewal_oversight_ec}). As in Section~\ref{sec:casestudy}, every result is a mean over 100 independent draws of the scenarios, and the tables report the 2.5--97.5\% range across draws in parentheses. All computations are reproducible from the code accompanying the paper.

\FloatBarrier
\subsection{Inputs and Provenance}\label{sec:provenance}

Table~\ref{tab:provenance} lists every input with its value, unit and source. The setting is the United States, and all monetary amounts are expressed in 2025 US dollars. Medical costs (work-up, biopsy, radiologist fees) are inflated with the CPI-U medical care index, and malpractice payments with the CPI-U all-items index. Both use the annual averages published by the Bureau of Labor Statistics (for 2025, the average of the eleven months published, October being unavailable). Where a published figure is used as a modeling choice rather than as an estimate, the table says so, and Section~\ref{sec:robustness_ec} varies it over the range suggested by the evidence.

\begin{table}[htbp]
\centering
\tiny
\caption{Inputs of the case study and their sources, in 2025 US dollars.}
\label{tab:provenance}
\begin{tabular}{@{}>{\raggedright\arraybackslash}p{3.0cm}>{\raggedright\arraybackslash}p{1.4cm}>{\raggedright\arraybackslash}p{3.0cm}>{\raggedright\arraybackslash}p{8.2cm}@{}}
\toprule \textbf{Input} & \textbf{Setting} & \textbf{Value} & \textbf{Source} \\ \midrule
Cancer prevalence & screening & 0.8\% (share of exams) & Modeling choice; \citealpec{lehman2017national} (cancer detection rate 5.1 + missed 0.8 per 1,000); robustness 0.4--1.2\% \\
Cancer prevalence & diagnostic & 15\% (share of exams) & Modeling choice representative of an enriched diagnostic cohort; \citealpec{sprague2017national} for context \\
AI discrimination & screening & AUC 0.85, binormal & \citealpec{mckinney2020international}; \citealpec{chen2023performance}; performance in new populations below development studies (\citealpec{yu2022external}) \\
AI discrimination & diagnostic & AUC 0.90, trigonometric & \citealpec{lotter2021robust}; \citealpec{rodriguez2019stand} \\
Annual volume & screening & 50,000 (exams/yr) & Large hospital network or regional screening service \\
Annual volume & diagnostic & 6,000 (exams/yr) & Separate enriched diagnostic cohort \\
Operational cost of a false positive & screening & \$1,191 (2025 USD) & \citealpec{ong2015}: \$852 health-plan expenditure within 12 months of a false-positive mammogram (2011--13 data, taken as 2012 USD); CPI-U medical care 414.924 (2012) to 580.102 (2025), factor 1.398 \\
Operational cost of a false positive & diagnostic & \$2,516 (2025 USD) & \citealpec{vlahiotis2018}: \$1,938 mean payment per breast biopsy (2015 USD, MarketScan); CPI-U medical care 446.752 (2015) to 580.102 (2025), factor 1.298 \\
Missed-cancer claim size (base) & screening & \$250,000 (per paid claim) & Modeling choice; PIAA average \$438,047 (2002, via \citealpec{dick2009}); \citealpec{studdert2006claims}; \citealpec{lee2020breast} \\
Missed-cancer claim size (base) & diagnostic & \$500,000 (per paid claim) & Modeling choice; same sources \\
False-positive claim size (base) & both & \$15,000 / \$25,000 (per paid claim) & Modeling choice (no published source) \\
Paid-claim probability, missed cancer & both & 2\% & \citealpec{localio1991}; \citealpec{studdert2000} (1.5--3\% file) $\times$ 38--73\% paid (\citealpec{harvey2016}; \citealpec{studdert2006claims}); \citealpec{vanbreestsmallenburg2012} \\
Paid-claim probability, false positive & both & 0.1\% & Modeling choice; lawsuits over unnecessary work-up are uncommon (\citealpec{elmore2005radiology}) \\
Scenario tiers & both & 4 tiers; shares 50/30/15/5\% and 40/30/25/5\% & Elicited liability regimes; FN-heavy multipliers calibrated to mean/median 2.4 (\citealpec{studdert2006claims}; \citealpec{jena2011}) and 7.6\% of claims above 3.03$\times$ the mean (\citealpec{schaffer2017}) \\
Coverage limits & both & \$1M per occurrence / \$3M aggregate & Most common US physician limits (\citealpec{ama2025}) \\
Premium & both & $(1+\mu)\,\mathbb{E}[C] + r\,(\text{CVaR}_\beta[C] - \mathbb{E}[C])$ & Cost-of-capital method (Solvency II risk margin); $r = 1 + \mu$ gives the tail-priced premium \\
Premium loading $\mu$ & both & 0.15 & Implied loading over expected payouts 20--24\%, against 22--32\% in the US medical professional liability market (\citealpec{naic2026}; \citealpec{burns2024}) \\
Insurer cost of tail capital $r$ & both & 6\% & Solvency II cost-of-capital rate (\citealpec{solvency2015}, Art. 39); US P\&C cost of equity 6.11\% (\citealpec{damodaran2026}) \\
Firm cost of equity capital $\rho$ & both & 8\% & \citealpec{damodaran2026}: total market 8.02\%; hospitals 7.52\%; healthcare IT 8.89\% \\
Capital multiplier $\gamma$ & both & 4.0 & Modeling choice; top of the market-risk VaR multiplier (3.00--4.00, \citealpec{cfr217}) \\
Capital multiplier $\gamma_{\text{ins}}$ & both & 3.2 & Basel II caps insurance recognition at 20\% of the operational-risk charge (\citealpec{basel2006international, migueis2020}) \\
Oversight cost scale $c_0$ & screening & \$1,746,711 (2025 USD/yr) & \citealpec{cms2025}: 77067-26, 1.08 RVU $\times$ \$32.3465 = \$34.93 per read $\times$ 50,000 \\
Oversight cost scale $c_0$ & diagnostic & \$271,711 (2025 USD/yr) & \citealpec{cms2025}: 77066-26, 1.40 RVU $\times$ \$32.3465 = \$45.29 per read $\times$ 6,000 \\
Oversight curvature and gain & both & $a = 3$, $b = 1.5$, $\xi = 0.5$ & Modeling choice; robustness over $a$, $b$, $c_0$, $\xi$ \\
Drift rates & both & $\delta = 0.10, 0.60$ per year & \citealpec{singh2025experiences} \\
Renewal cost & both & 5\% of baseline CVaR & Modeling choice; robustness 2--10\% \\
Risk level $\beta$ & both & 0.95 & Modeling choice \\
Scenarios & both & $J = 1{,}000$ per draw; 100 independent draws (seeds 0--99) & Results are means over the draws, with the 2.5th--97.5th percentile range across draws in parentheses \\
\bottomrule
\end{tabular}
\end{table}

\FloatBarrier
\subsection{Scenario Construction}\label{sec:scenario_construction}

\paragraph{Cost components.} A false positive incurs an operational work-up cost $c_{\mathrm{op}}$ borne by the firm in every case, and a liability cost when it leads to a paid claim. A missed cancer incurs a liability cost only. With $\phi_{\mathrm{FP}}$ and $\phi_{\mathrm{FN}}$ the probabilities that an error leads to a paid claim and $\mathrm{sev}_{K,j}$, $\mathrm{sev}_{L,j}$ the claim sizes in scenario $j$, the per-error costs are the same for every instance of a scenario, $K_{ji} = \kappa_j$ and $L_{ji} = \ell_j$ in the notation of Equation~\eqref{eq:loss_scenario}, with
\begin{equation}\label{eq:cost_components}
\kappa_j = c_{\mathrm{op}} + \phi_{\mathrm{FP}}\,\mathrm{sev}_{K,j}, \qquad \ell_j = \phi_{\mathrm{FN}}\,\mathrm{sev}_{L,j},
\end{equation}
so that the aggregate costs are $K_j = N\kappa_j$, $L_j = N\ell_j$ and the annual loss of scenario $j$ at threshold $\tau$ is $S_j(\tau) = N\,[\mathbb{P}_{\mathrm{FP}}(\tau)\kappa_j + \mathbb{P}_{\mathrm{FN}}(\tau)\ell_j]$, with $N$ the annual volume and $\mathbb{P}_{\mathrm{FP}}$, $\mathbb{P}_{\mathrm{FN}}$ the joint error probabilities of the performance model (Lemmas~\ref{lem:fn_case}--\ref{lem:fp_case} for the trigonometric model and Section~\ref{sec:binormal} for the binormal model). The tail set consists of the $50$ worst of the $J = 1{,}000$ scenarios at $\beta = 0.95$. The insurer's payment in scenario $j$ evaluates Equation~\eqref{eq:insurer_payment} on the liability component with the expected numbers of paid claims, $n_{\mathrm{FP}} = N\,\mathbb{P}_{\mathrm{FP}}(\tau^{\ast})\,\phi_{\mathrm{FP}}$ and $n_{\mathrm{FN}} = N\,\mathbb{P}_{\mathrm{FN}}(\tau^{\ast})\,\phi_{\mathrm{FN}}$:
\begin{equation}\label{eq:payout_scenario}
C_j = \min\bigl\{L_{\text{agg}},\ n_{\mathrm{FN}}\min\{\mathrm{sev}_{L,j}, L_{\text{occ}}\} + n_{\mathrm{FP}}\min\{\mathrm{sev}_{K,j}, L_{\text{occ}}\}\bigr\},
\end{equation}
and the retained loss is $S_j(\tau^{\ast}) - C_j$. Evaluating the capped payment at the expected numbers of paid claims is a deterministic approximation. Random claim counts would lower the expected payment where the aggregate limit can bind, as in screening, and would raise the tail of the payment when that tail is taken over claim counts as well as scenarios.

\paragraph{Scenario tiers.} The claim sizes $\mathrm{sev}_{K,j}$ and $\mathrm{sev}_{L,j}$ are the base claim sizes of Table~\ref{tab:cost_distributions} times scenario-level multipliers. Scenarios are grouped in four tiers, which represent elicited liability regimes rather than historical frequencies, with shares of 50/30/15/5\% in screening and 40/30/25/5\% in the diagnostic setting. The standard tier represents ordinary conditions, and the elevated, high and extreme tiers represent common shocks that affect all claims of a period, such as the legal climate, litigation waves, or correlated errors of a model version. Within the elevated, high and extreme tiers, a scenario is either false-positive-heavy (litigation over unnecessary interventions) or false-negative-heavy (delayed diagnoses of aggressive cancers), with the probabilities in the table. The multipliers of each option are drawn uniformly within their ranges. Because a tier multiplier scales all of a scenario's claims together, the tiers induce the heterogeneity of tail losses that the CVaR analysis prices.

\paragraph{Calibration of the tail.} The shares and the false-positive multipliers are elicited values. The false-negative multipliers of the high and extreme tiers were rescaled, by one factor per tier, so that the distribution of a paid missed-cancer claim implied by the tiers matches two published, scale-free statistics of US paid malpractice claims. The first is a mean-to-median ratio of 2.4 (2.35 in \citeec{studdert2006claims}, 2.46 in \citeec{jena2011}). The second is a share of 7.6\% of claims above 3.03 times the mean, the proportion of paid claims exceeding \$1 million against a mean of \$329{,}565 in the National Practitioner Data Bank, 1992--2014 \citepec{schaffer2017}. The calibrated table reproduces both statistics (Table~\ref{tab:cost_distributions}). The implied coefficient of variation (1.9 in screening, 1.1 in diagnostic) remains below the 2.4 of litigated breast cancer cases \citepec{lee2020breast}. Two caveats apply. First, the published statistics are per claim, whereas a tier multiplier scales all of a year's claims. Second, with 2.6 (screening) and 0.3 (diagnostic) expected paid missed-cancer claims per year, a year's liability is dominated by one or two claims, so the distribution of annual losses is coarser than that of claim sizes. Section~\ref{sec:robustness_ec} varies the tail across the range of the published targets and back to the uncalibrated multipliers.

\begin{table}[htbp]
\centering
\tiny
\caption{Scenario tiers.}
\label{tab:cost_distributions}
\begin{tabular}{lll}
\toprule
\textbf{Component} & \textbf{Screening Scenario} & \textbf{Diagnostic Scenario} \\
\midrule
\multicolumn{3}{l}{\textit{False positive claim size}} \\
Base claim size & \$15,000 & \$25,000 \\
Scenario distribution & & \\
\quad Standard scenarios & 50\%, 0.8--1.2$\times$ & 40\%, 0.9--1.1$\times$ \\
\quad Elevated risk & 30\%, 1.5--2$\times$ (50\%) & 30\%, 1.5--2.5$\times$ (40\%) \\
\quad & or 0.7--0.9$\times$ (50\%) & or 0.6--0.8$\times$ (60\%) \\
\quad High risk & 15\%, 2.5--4$\times$ (40\%) & 25\%, 3--5$\times$ (30\%) \\
\quad & or 0.5--0.7$\times$ (60\%) & or 0.4--0.6$\times$ (70\%) \\
\quad Extreme tail & 5\%, 4--6$\times$ (30\%) & 5\%, 6--10$\times$ (20\%) \\
\quad & or 0.3--0.5$\times$ (70\%) & or 0.2--0.4$\times$ (80\%) \\
\midrule
\multicolumn{3}{l}{\textit{False negative claim size}} \\
Base claim size & \$250,000 & \$500,000 \\
Scenario distribution & & \\
\quad Standard scenarios & 50\%, 0.9--1.1$\times$ & 40\%, 0.9--1.1$\times$ \\
\quad Elevated risk & 30\%, 0.8--1$\times$ (50\%) & 30\%, 0.7--0.9$\times$ (40\%) \\
\quad & or 1.2--1.8$\times$ (50\%) & or 1.5--2.5$\times$ (60\%) \\
\quad High risk & 15\%, 0.7--1$\times$ (40\%) & 25\%, 0.6--1$\times$ (30\%) \\
\quad & or 5.8--8.7$\times$ (60\%) & or 4.25--8.5$\times$ (70\%) \\
\quad Extreme tail & 5\%, 0.8--1.2$\times$ (30\%) & 5\%, 0.5--1$\times$ (20\%) \\
\quad & or 16.24--32.48$\times$ (70\%) & or 8.25--15.47$\times$ (80\%) \\
\midrule
\multicolumn{3}{l}{\textit{Implied dispersion of false-negative claim sizes}} \\
Mean / median & 2.40 & 2.40 \\
Share of claims above 3.03$\times$ the mean & 7.6\% & 7.6\% \\
Coefficient of variation & 1.90 & 1.13 \\
Mean claim size & \$607,775 & \$1,256,763 \\
\midrule
\multicolumn{3}{l}{\textit{Key Parameters}} \\
Annual patient volume & 50,000 & 6,000 \\
Cancer prevalence & 0.8\% & 15\% \\
Work-up cost per false positive & \$1,191 & \$2,516 \\
Paid-claim probability (FP / FN) & 0.1\% / 2\% & 0.1\% / 2\% \\
Average K/L ratio (per error) & 0.100 (0.093--0.107) & 0.101 (0.097--0.108) \\
Tail K/L ratio (worst 5\%, at $\tau^{\ast}$) & 0.012 (0.011--0.013) & 0.023 (0.021--0.024) \\
\bottomrule
\end{tabular}
\par\smallskip{\raggedright\textit{Notes.} Shares of the $J = 1{,}000$ scenarios by tier, with the probabilities of the two options within a tier in parentheses, and multipliers of the base claim sizes. The false-positive and false-negative multipliers of an option are drawn independently and uniformly within their ranges. The calibration targets are a mean/median ratio of 2.4 \citepec{studdert2006claims, jena2011} and 7.6\% of paid claims above 3.03$\times$ the mean \citepec{schaffer2017}.\par}
\end{table}

\FloatBarrier
\subsection{Robustness of the Calibration Results}\label{sec:robustness_ec}

\paragraph{The screening operating point.} US screening practice recalls 11.6\% of women with a sensitivity of 0.87 \citepec{lehman2017national}. The risk-aware policy has a higher recall rate (15.9\%) and a lower sensitivity (0.67), because a model with $\text{AUC} = 0.85$ reaches a sensitivity of only 0.60 at the recall rate of current practice. The policy minimizes the firm's financial loss, in which a missed cancer enters through a 2\% probability of a paid claim, whereas clinical practice values detection for its own sake. Consistent with this distinction, malpractice concerns are not associated with radiologists' recall rates \citepec{elmore2005radiology}. The liability-driven optimum therefore need not coincide with clinical standards.

\paragraph{One-at-a-time variations.} Table~\ref{tab:robustness} varies one input at a time over the range suggested by its evidence. The screening prevalence takes 0.4\% and 1.2\%, around the 0.8\% base, and the diagnostic prevalence 10\% and 20\%. The work-up cost takes the lower bound of \citeec{tosteson2008cost}, \$330 per false-positive screen and \$723 per core biopsy in 2005 Medicare rates, inflated. The claim sizes take half and twice their base. The missed-cancer paid-claim probability takes 0.5\% and the stress values of 20\% (screening) and 35\% (diagnostic), and the false-positive one takes zero and the stress values of 2\% and 3\%. The tail takes the lightest and heaviest published targets (mean/median 2.35 with 6.2\% above the threshold, and 2.46 with 8.0\%, the range of \citeec{schaffer2017} over 1992--2014) and the uncalibrated multipliers. Every row is a mean over the 100 scenario draws with its 2.5--97.5\% range, which also measures the stability of the optimum under scenario sampling (see Lemma~\ref{lem:continuity}). Two further rows per setting perturb every scenario's costs independently and uniformly by up to $\pm 10\%$ and $\pm 25\%$, and a further row reports the other performance model at the same AUC as a misspecification check.

\paragraph{Regularities.} Three regularities hold across all rows. First, the direction of the results is unchanged. The risk-aware threshold has a lower tail risk than the accuracy and expected-loss thresholds in every variation of the inputs and in every draw. The mean penalties are at least 18\% (accuracy) and 11\% (expected loss), with one exception. Under the misspecified screening model the expected-loss penalty is negligible, because both thresholds then lie at the prevalence. Second, the level of the risk-aware threshold is governed by the ratio of missed-cancer liability to work-up cost. Within the evidence-based ranges the screening policy recalls between 3\% and 30\% of examinations, whereas the stress values of the paid-claim probabilities raise the recall to 71\% (screening) and 93\% (diagnostic). The case study therefore uses the evidence-based rates. Third, the optimum is stable. Across the 100 scenario draws the share flagged ranges within 14--18\% in screening and 70--72\% in the diagnostic setting. Independent cost perturbations of $\pm 25\%$ lower it on average, to 15.2\% (13.1--17.2\%) and 65.4\% (64.1--66.7\%), and leave the base insurance value within 0.3 points of its value at the unperturbed costs.

\paragraph{Misspecification.} Under the trigonometric model the screening threshold falls in its second regime and flags almost exactly the prevalence (0.8\%) at the base inputs and whenever the tail cost share exceeds the prevalence, illustrating the limitation discussed in Section~\ref{sec:casesetup}. The binormal model in the diagnostic setting flags 65\% against 71\%, with the same ordering of policies.

\begin{table}[htbp]
\centering
\tiny
\setlength{\tabcolsep}{3pt}
\caption{Robustness of the calibration results to one-at-a-time input variations.}
\label{tab:robustness}
\setlength{\tabcolsep}{1pt}
\begin{tabular}{@{}>{\raggedright\arraybackslash}p{1.5cm}rrrrrrrr@{}}
\toprule
\textbf{Variation} & \textbf{Flagged} & \textbf{Sens.} & \textbf{PPV} & \textbf{Pen.\ acc.} & \textbf{Pen.\ EL} & \textbf{$V$ acc.} & \textbf{$V$ EL} & \textbf{$V$ CVaR} \\
\midrule
\multicolumn{9}{l}{\textit{Screening}} \\ \midrule
Base inputs & 15.9\,(14.3--17.7) & 67\,(65--70) & 3\,(3--4) & 77\,(70--86) & 51\,(46--57) & 5.7\,(4.9--6.6) & $-$4.2\,($-$4.6--$-$3.8) & 8.3\,(8.1--8.6) \\
Prevalence 0.4\% & 7.0\,(6.1--8.0) & 49\,(46--52) & 3\,(3--3) & 39\,(36--44) & 30\,(27--34) & 0.8\,(0.3--1.4) & $-$2.7\,($-$3.5--$-$1.8) & 8.8\,(8.5--9.1) \\
Prevalence 1.2\% & 23.7\,(21.7--26.0) & 77\,(75--79) & 4\,(4--4) & 113\,(103--125) & 66\,(60--73) & 13.9\,(11.8--16.0) & $-$3.7\,($-$3.8--$-$3.6) & 8.1\,(7.8--8.2) \\
Work-up cost, lower bound & 29.8\,(27.5--32.4) & 82\,(80--84) & 2\,(2--2) & 149\,(135--164) & 79\,(71--86) & 24.1\,(20.9--27.4) & $-$2.9\,($-$3.3--$-$2.4) & 7.8\,(7.6--8.1) \\
FN claim size $\times$0.5 & 7.2\,(6.3--8.2) & 49\,(46--52) & 5\,(5--6) & 39\,(35--44) & 30\,(27--34) & 1.8\,(1.3--2.4) & $-$1.5\,($-$2.3--$-$0.7) & 10.0\,(9.8--10.3) \\
FN claim size $\times$2 & 29.8\,(27.5--32.4) & 82\,(80--84) & 2\,(2--2) & 149\,(135--164) & 78\,(71--86) & 23.8\,(20.5--27.2) & $-$3.4\,($-$3.9--$-$2.8) & 7.0\,(6.9--7.1) \\
FP claim size $\times$0.5 & 15.9\,(14.4--17.8) & 67\,(65--70) & 3\,(3--4) & 78\,(71--86) & 51\,(46--57) & 5.9\,(5.1--6.8) & $-$4.1\,($-$4.4--$-$3.6) & 8.3\,(8.1--8.6) \\
FP claim size $\times$2 & 15.8\,(14.3--17.6) & 67\,(65--70) & 3\,(3--4) & 77\,(70--85) & 51\,(46--57) & 5.2\,(4.4--6.1) & $-$4.6\,($-$4.9--$-$4.1) & 8.3\,(8.0--8.6) \\
FN claim prob. 0.5\% & 2.7\,(2.3--3.2) & 31\,(29--34) & 9\,(8--10) & 18\,(16--21) & 16\,(14--18) & 1.4\,(0.5--2.3) & 0.6\,($-$0.4--1.5) & 9.1\,(8.8--9.5) \\
FN claim prob. 20\%, stress & 71.2\,(68.9--73.4) & 98\,(97--98) & 1\,(1--1) & 670\,(609--743) & 137\,(125--147) & 231\,(208--256) & 13.9\,(11.1--16.8) & 6.9\,(6.8--7.0) \\
FP claim prob. 0 & 16.0\,(14.4--17.8) & 67\,(65--70) & 3\,(3--4) & 78\,(71--86) & 51\,(46--56) & 6.2\,(5.4--7.1) & $-$3.9\,($-$4.2--$-$3.5) & 8.3\,(8.1--8.6) \\
FP claim prob. 2\%, stress & 10.8\,(9.6--12.6) & 58\,(55--62) & 4\,(4--5) & 66\,(61--73) & 48\,(44--52) & 6.8\,(4.8--8.8) & $-$0.2\,($-$2.2--1.4) & 7.9\,(7.8--8.0) \\
Tails, uncalibrated & 2.9\,(2.5--3.2) & 32\,(30--34) & 9\,(8--9) & 19\,(17--21) & 11\,(10--13) & 9.1\,(8.5--10.0) & 5.6\,(4.8--6.5) & 11.6\,(11.3--11.9) \\
Tails, lightest targets & 15.7\,(14.1--17.6) & 67\,(65--70) & 3\,(3--4) & 77\,(69--85) & 51\,(46--57) & 5.4\,(4.7--6.3) & $-$4.2\,($-$4.6--$-$3.8) & 8.4\,(8.1--8.7) \\
Tails, heaviest targets & 16.5\,(14.9--18.4) & 68\,(66--71) & 3\,(3--4) & 80\,(73--89) & 53\,(48--58) & 6.0\,(5.1--7.0) & $-$4.4\,($-$4.6--$-$3.9) & 8.2\,(8.0--8.5) \\
Model, trigonometric & 0.8\,(0.8--0.8) & 54\,(54--54) & 54\,(54--54) & 25\,(24--25) & 1\,(0--1) & 19.1\,(18.8--19.5) & 9.1\,(8.8--9.4) & 9.0\,(8.7--9.2) \\
Costs $\pm 10\%$ & 15.6 (13.9--17.4) & & & & & & & 8.4 (8.1--8.6)\textsuperscript{a} \\
Costs $\pm 25\%$ & 15.2 (13.1--17.2) & & & & & & & 8.4 (8.1--8.7)\textsuperscript{a} \\
\midrule
\multicolumn{9}{l}{\textit{Diagnostic}} \\ \midrule
Base inputs & 71.0\,(69.9--72.0) & 98\,(98--99) & 21\,(21--21) & 297\,(275--320) & 74\,(69--82) & 105\,(95--114) & 8.8\,(7.9--10.0) & 6.6\,(6.5--6.6) \\
Prevalence 10\% & 62.1\,(60.7--63.4) & 97\,(97--97) & 16\,(15--16) & 186\,(171--202) & 106\,(99--117) & 47.9\,(41.4--53.3) & 10.9\,(9.6--12.8) & 6.8\,(6.7--6.8) \\
Prevalence 20\% & 76.2\,(75.4--76.9) & 99\,(99--99) & 26\,(26--26) & 401\,(373--432) & 60\,(56--65) & 161\,(148--173) & 8.0\,(7.3--9.2) & 6.5\,(6.5--6.5) \\
Work-up cost, lower bound & 77.5\,(75.9--78.9) & 99\,(99--99) & 19\,(19--20) & 597\,(555--643) & 50\,(46--55) & 267\,(248--284) & 9.7\,(8.7--10.9) & 6.4\,(6.3--6.5) \\
FN claim size $\times$0.5 & 58.4\,(56.9--59.8) & 96\,(95--96) & 25\,(24--25) & 129\,(118--141) & 87\,(78--96) & 22.2\,(17.9--25.7) & 2.0\,($-$1.1--4.6) & 7.7\,(7.5--7.9) \\
FN claim size $\times$2 & 79.3\,(78.5--80.2) & 99\,(99--99) & 19\,(19--19) & 625\,(582--671) & 49\,(46--54) & 280\,(260--298) & 8.3\,(7.4--9.2) & 6.2\,(6.2--6.3) \\
FP claim size $\times$0.5 & 71.0\,(70.0--72.0) & 99\,(98--99) & 21\,(21--21) & 297\,(275--321) & 74\,(68--81) & 106\,(96--115) & 8.9\,(8.0--10.1) & 6.7\,(6.7--6.8) \\
FP claim size $\times$2 & 69.5\,(68.1--70.9) & 98\,(98--98) & 21\,(21--22) & 295\,(273--318) & 75\,(70--83) & 105\,(95--113) & 10.2\,(9.0--11.4) & 6.4\,(6.4--6.5) \\
FN claim prob. 0.5\% & 40.3\,(38.2--42.4) & 88\,(87--89) & 33\,(32--34) & 46\,(40--51) & 24\,(20--28) & $-$8.8\,($-$9.8--$-$8.1) & $-$18.1\,($-$18.9--$-$17.3) & 7.4\,(7.2--7.5) \\
FN claim prob. 35\%, stress & 92.5\,(92.0--92.9) & 100\,(100--100) & 16\,(16--16) & 5,433\,(5,066--5,824) & 14\,(12--15) & 2,899\,(2,724--3,061) & 7.9\,(7.6--8.1) & 6.5\,(6.4--6.5) \\
FP claim prob. 0 & 71.0\,(70.0--72.0) & 99\,(98--99) & 21\,(21--21) & 298\,(276--321) & 73\,(68--80) & 107\,(97--115) & 8.9\,(8.1--10.2) & 6.8\,(6.7--6.8) \\
FP claim prob. 3\%, stress & 48.8\,(46.5--51.1) & 92\,(91--93) & 28\,(27--30) & 165\,(144--189) & 59\,(52--71) & 63.7\,(55.2--71.6) & 19.0\,(17.3--21.3) & 7.7\,(7.6--7.8) \\
Tails, uncalibrated & 69.7\,(68.3--70.9) & 98\,(98--98) & 21\,(21--22) & 273\,(245--303) & 91\,(84--99) & 87.6\,(76.7--98.0) & 11.0\,(9.4--13.0) & 6.7\,(6.6--6.7) \\
Tails, lightest targets & 71.3\,(70.2--72.4) & 99\,(98--99) & 21\,(20--21) & 304\,(279--331) & 80\,(75--88) & 105\,(95--115) & 9.7\,(8.8--11.2) & 6.6\,(6.5--6.6) \\
Tails, heaviest targets & 71.6\,(70.6--72.6) & 99\,(98--99) & 21\,(20--21) & 311\,(288--335) & 75\,(70--82) & 111\,(101--120) & 9.1\,(8.2--10.3) & 6.6\,(6.5--6.6) \\
Model, binormal & 64.8\,(63.5--66.1) & 98\,(98--98) & 23\,(22--23) & 444\,(416--473) & 53\,(50--58) & 191\,(178--201) & 3.1\,(2.7--3.5) & 6.8\,(6.8--6.9) \\
Costs $\pm 10\%$ & 68.7 (67.7--69.8) & & & & & & & 6.7 (6.7--6.8)\textsuperscript{a} \\
Costs $\pm 25\%$ & 65.4 (64.1--66.7) & & & & & & & 6.9 (6.8--6.9)\textsuperscript{a} \\
\bottomrule
\end{tabular}
\par\smallskip{\raggedright\textit{Notes.} Share flagged, sensitivity and positive predictive value (PPV) at $\tau^{\ast}$, risk penalties of the accuracy and expected-loss thresholds, and total insurance value by firm type relative to $\text{CVaR}_\beta[S(\tau^{\ast})]$, all in \%. The rows Costs $\pm 10\%$ and $\pm 25\%$ perturb every scenario's costs independently and uniformly by up to that amount, once per draw. \textsuperscript{a}Base value $V_{\mathrm{base}}$, common to the three firm types.\par}
\end{table}

\FloatBarrier
\subsection{Sensitivity of Insurance Value}\label{sec:insurance_sensitivity_ec}

\paragraph{Premium principle and coverage limits.} Under the tail-priced premium $r = 1 + \mu$ (Section~\ref{sec:insurance_motivation}), the market limits would carry loadings of 188\% (screening) and 120\% (diagnostic) over expected payouts, against 24\% and 20\% under the cost-of-capital premium, and the base value would fall from 8.3\% and 6.6\% to 0.3\% and 3.8\%. Most of the value of coverage rests on the insurer pricing the tail at its cost of capital rather than at the loading. The limiting values of Proposition~\ref{prop:limit_behavior}, evaluated at the inputs of Section~\ref{sec:valueinsurance_case}, bracket the base value, from the pure capital relief of 6.4\% in both settings with vanishing coverage to 17.2\% and 9.0\% with full coverage. The tail-to-average ratio of the insured loss at the same inputs, $\chi = 7.4$ (screening) and $3.4$ (diagnostic), satisfies the condition of Corollary~\ref{cor:viability}, since $\rho\gamma = 0.32 > \max\{\mu, r\}$.

\paragraph{Sensitivity to the insurance parameters.} Table~\ref{tab:insurance_sensitivity} reports the components of Theorem~\ref{thm:insurance_value} under $\pm 25\%$ variations of $\mu$, $\rho$, $\gamma$, and $\gamma_{\text{ins}}$, across the capital relief $1 - \gamma_{\text{ins}}/\gamma$ from none to one half, and across the insurer's cost of tail capital $r$ from the 4.75\% that Solvency II adopts from 2027 to the tail-priced premium $r = 1 + \mu$. The base value is linear in $\gamma_{\text{ins}}$ and most sensitive to the relief and to $\rho\gamma$. It ranges from 2.7\% to 16.8\% in screening and from 0.4\% to 15.9\% in the diagnostic setting as the relief goes from none to one half. It changes by about $\pm 2.4$ points and $\pm 1.8$ points for $\pm 25\%$ in $\rho$ or $\gamma$, against $\pm 0.2$ points for $\mu$ and about 0.4 points for $r$ within the cost-of-capital range. The calibration value of the accuracy-maximizing firm in the diagnostic setting exceeds 74\% in every row. The sign of the total value of each firm type is the same in every row, with two exceptions. The screening accuracy-maximizing firm turns negative when $\rho$ or $\gamma$ falls by 25\% or under the tail-priced premium, and the screening expected-loss firm turns positive when $\rho$ or $\gamma$ rises by 25\% or the relief reaches 40\% or more.

\begin{table}[htbp]
\centering
\tiny
\setlength{\tabcolsep}{1.5pt}
\caption{Sensitivity of insurance value.}
\label{tab:insurance_sensitivity}
\begin{tabular}{@{}lrrrrrrr@{}}
\toprule
\textbf{Parameter} & \textbf{Value} & \textbf{$V_{\mathrm{base}}$} & \textbf{$V_{\mathrm{calib}}$ acc.} & \textbf{$V_{\mathrm{calib}}$ EL} & \textbf{$V$ acc.} & \textbf{$V$ EL} & \textbf{$V$ CVaR} \\
\midrule
\multicolumn{8}{l}{\textit{Screening}} \\ \midrule
$\mu$ ($\times0.75$) & 0.1125 & 8.5\,(8.2--8.8) & $-$2.6\,($-$3.7--$-$1.5) & $-$12.6\,($-$12.7--$-$12.4) & 5.9\,(5.1--6.8) & $-$4.1\,($-$4.4--$-$3.6) & 8.5\,(8.2--8.8) \\
$\mu$ ($\times1$) & 0.15 & 8.3\,(8.1--8.6) & $-$2.6\,($-$3.7--$-$1.5) & $-$12.6\,($-$12.7--$-$12.4) & 5.7\,(4.9--6.6) & $-$4.2\,($-$4.6--$-$3.8) & 8.3\,(8.1--8.6) \\
$\mu$ ($\times1.25$) & 0.1875 & 8.2\,(7.9--8.4) & $-$2.6\,($-$3.7--$-$1.5) & $-$12.6\,($-$12.7--$-$12.4) & 5.5\,(4.7--6.4) & $-$4.4\,($-$4.7--$-$4.0) & 8.2\,(7.9--8.4) \\
$\rho$ ($\times0.75$) & 0.06 & 6.0\,(5.8--6.1) & $-$8.8\,($-$9.5--$-$8.3) & $-$16.7\,($-$17.1--$-$16.1) & $-$2.9\,($-$3.5--$-$2.4) & $-$10.7\,($-$11.4--$-$9.9) & 6.0\,(5.8--6.1) \\
$\rho$ ($\times1$) & 0.08 & 8.3\,(8.1--8.6) & $-$2.6\,($-$3.7--$-$1.5) & $-$12.6\,($-$12.7--$-$12.4) & 5.7\,(4.9--6.6) & $-$4.2\,($-$4.6--$-$3.8) & 8.3\,(8.1--8.6) \\
$\rho$ ($\times1.25$) & 0.1 & 10.7\,(10.3--11.1) & 3.6\,(1.9--5.2) & $-$8.5\,($-$8.8--$-$8.1) & 14.3\,(13.0--15.7) & 2.2\,(2.2--2.4) & 10.7\,(10.3--11.1) \\
$\gamma$ ($\times0.75$) & 3 & 6.0\,(5.8--6.1) & $-$8.8\,($-$9.5--$-$8.3) & $-$16.7\,($-$17.1--$-$16.1) & $-$2.9\,($-$3.5--$-$2.4) & $-$10.7\,($-$11.4--$-$9.9) & 6.0\,(5.8--6.1) \\
$\gamma$ ($\times1$) & 4 & 8.3\,(8.1--8.6) & $-$2.6\,($-$3.7--$-$1.5) & $-$12.6\,($-$12.7--$-$12.4) & 5.7\,(4.9--6.6) & $-$4.2\,($-$4.6--$-$3.8) & 8.3\,(8.1--8.6) \\
$\gamma$ ($\times1.25$) & 5 & 10.7\,(10.3--11.1) & 3.6\,(1.9--5.2) & $-$8.5\,($-$8.8--$-$8.1) & 14.3\,(13.0--15.7) & 2.2\,(2.2--2.4) & 10.7\,(10.3--11.1) \\
$\gamma_{\text{ins}}$ ($\times0.75$) & 2.4 & 14.0\,(13.8--14.1) & $-$2.6\,($-$3.7--$-$1.5) & $-$12.6\,($-$12.7--$-$12.4) & 11.3\,(10.4--12.3) & 1.4\,(1.2--1.7) & 14.0\,(13.8--14.1) \\
$\gamma_{\text{ins}}$ ($\times1$) & 3.2 & 8.3\,(8.1--8.6) & $-$2.6\,($-$3.7--$-$1.5) & $-$12.6\,($-$12.7--$-$12.4) & 5.7\,(4.9--6.6) & $-$4.2\,($-$4.6--$-$3.8) & 8.3\,(8.1--8.6) \\
$\gamma_{\text{ins}}$ ($\times1.25$) & 4 & 2.7\,(2.3--3.1) & $-$2.6\,($-$3.7--$-$1.5) & $-$12.6\,($-$12.7--$-$12.4) & 0.1\,($-$0.6--0.9) & $-$9.9\,($-$10.3--$-$9.3) & 2.7\,(2.3--3.1) \\
\cmidrule(l){1-8}
$\gamma_{\text{ins}}/\gamma$ & 0.5 & 16.8\,(16.7--16.9) & & & 14.1\,(13.1--15.2) & 4.2\,(4.0--4.5) & 16.8\,(16.7--16.9) \\
$\gamma_{\text{ins}}/\gamma$ & 0.6 & 14.0\,(13.8--14.1) & & & 11.3\,(10.4--12.3) & 1.4\,(1.2--1.7) & 14.0\,(13.8--14.1) \\
$\gamma_{\text{ins}}/\gamma$ & 0.7 & 11.1\,(10.9--11.4) & & & 8.5\,(7.6--9.4) & $-$1.4\,($-$1.7--$-$1.0) & 11.1\,(10.9--11.4) \\
$\gamma_{\text{ins}}/\gamma$ & 0.8 & 8.3\,(8.1--8.6) & & & 5.7\,(4.9--6.6) & $-$4.2\,($-$4.6--$-$3.8) & 8.3\,(8.1--8.6) \\
$\gamma_{\text{ins}}/\gamma$ & 0.9 & 5.5\,(5.2--5.8) & & & 2.9\,(2.2--3.7) & $-$7.1\,($-$7.4--$-$6.5) & 5.5\,(5.2--5.8) \\
$\gamma_{\text{ins}}/\gamma$ & 1 & 2.7\,(2.3--3.1) & & & 0.1\,($-$0.6--0.9) & $-$9.9\,($-$10.3--$-$9.3) & 2.7\,(2.3--3.1) \\
\cmidrule(l){1-8}
$r$ (loading 22\%) & 0.0475 & 8.4\,(8.1--8.7) & & & 5.8\,(5.0--6.7) & $-$4.1\,($-$4.5--$-$3.7) & 8.4\,(8.1--8.7) \\
$r$ (loading 24\%) & 0.06 & 8.3\,(8.1--8.6) & & & 5.7\,(4.9--6.6) & $-$4.2\,($-$4.6--$-$3.8) & 8.3\,(8.1--8.6) \\
$r$ (loading 27\%) & 0.08 & 8.2\,(7.9--8.4) & & & 5.6\,(4.7--6.5) & $-$4.4\,($-$4.7--$-$3.9) & 8.2\,(7.9--8.4) \\
$r$ (loading 30\%) & 0.1 & 8.0\,(7.8--8.3) & & & 5.4\,(4.5--6.3) & $-$4.5\,($-$4.8--$-$4.1) & 8.0\,(7.8--8.3) \\
$r = 1 + \mu$ (loading 188\%) & 1.15 & 0.3\,($-$0.6--1.2) & & & $-$2.3\,($-$4.4--$-$0.4) & $-$12.3\,($-$13.2--$-$11.4) & 0.3\,($-$0.6--1.2) \\
\midrule
\multicolumn{8}{l}{\textit{Diagnostic}} \\ \midrule
$\mu$ ($\times0.75$) & 0.1125 & 6.7\,(6.6--6.7) & 99\,(89--107) & 2.2\,(1.3--3.5) & 105\,(95--114) & 8.9\,(8.0--10.1) & 6.7\,(6.6--6.7) \\
$\mu$ ($\times1$) & 0.15 & 6.6\,(6.5--6.6) & 99\,(89--107) & 2.2\,(1.3--3.5) & 105\,(95--114) & 8.8\,(7.9--10.0) & 6.6\,(6.5--6.6) \\
$\mu$ ($\times1.25$) & 0.1875 & 6.5\,(6.4--6.5) & 99\,(89--107) & 2.2\,(1.3--3.5) & 105\,(95--114) & 8.7\,(7.8--9.9) & 6.5\,(6.4--6.5) \\
$\rho$ ($\times0.75$) & 0.06 & 4.8\,(4.8--4.8) & 74.9\,(66.3--81.7) & $-$3.8\,($-$4.3--$-$3.0) & 79.7\,(71.1--86.5) & 1.0\,(0.5--1.8) & 4.8\,(4.8--4.8) \\
$\rho$ ($\times1$) & 0.08 & 6.6\,(6.5--6.6) & 99\,(89--107) & 2.2\,(1.3--3.5) & 105\,(95--114) & 8.8\,(7.9--10.0) & 6.6\,(6.5--6.6) \\
$\rho$ ($\times1.25$) & 0.1 & 8.4\,(8.3--8.4) & 122\,(111--133) & 8.1\,(6.8--10.0) & 131\,(119--141) & 16.5\,(15.2--18.3) & 8.4\,(8.3--8.4) \\
$\gamma$ ($\times0.75$) & 3 & 4.8\,(4.8--4.8) & 74.9\,(66.3--81.7) & $-$3.8\,($-$4.3--$-$3.0) & 79.7\,(71.1--86.5) & 1.0\,(0.5--1.8) & 4.8\,(4.8--4.8) \\
$\gamma$ ($\times1$) & 4 & 6.6\,(6.5--6.6) & 99\,(89--107) & 2.2\,(1.3--3.5) & 105\,(95--114) & 8.8\,(7.9--10.0) & 6.6\,(6.5--6.6) \\
$\gamma$ ($\times1.25$) & 5 & 8.4\,(8.3--8.4) & 122\,(111--133) & 8.1\,(6.8--10.0) & 131\,(119--141) & 16.5\,(15.2--18.3) & 8.4\,(8.3--8.4) \\
$\gamma_{\text{ins}}$ ($\times0.75$) & 2.4 & 12.8\,(12.8--12.8) & 99\,(89--107) & 2.2\,(1.3--3.5) & 111\,(101--120) & 15.0\,(14.1--16.2) & 12.8\,(12.8--12.8) \\
$\gamma_{\text{ins}}$ ($\times1$) & 3.2 & 6.6\,(6.5--6.6) & 99\,(89--107) & 2.2\,(1.3--3.5) & 105\,(95--114) & 8.8\,(7.9--10.0) & 6.6\,(6.5--6.6) \\
$\gamma_{\text{ins}}$ ($\times1.25$) & 4 & 0.4\,(0.3--0.4) & 99\,(89--107) & 2.2\,(1.3--3.5) & 99\,(89--108) & 2.6\,(1.7--3.8) & 0.4\,(0.3--0.4) \\
\cmidrule(l){1-8}
$\gamma_{\text{ins}}/\gamma$ & 0.5 & 15.9\,(15.9--15.9) & & & 115\,(104--123) & 18.1\,(17.2--19.4) & 15.9\,(15.9--15.9) \\
$\gamma_{\text{ins}}/\gamma$ & 0.6 & 12.8\,(12.8--12.8) & & & 111\,(101--120) & 15.0\,(14.1--16.2) & 12.8\,(12.8--12.8) \\
$\gamma_{\text{ins}}/\gamma$ & 0.7 & 9.7\,(9.7--9.7) & & & 108\,(98--117) & 11.9\,(11.0--13.1) & 9.7\,(9.7--9.7) \\
$\gamma_{\text{ins}}/\gamma$ & 0.8 & 6.6\,(6.5--6.6) & & & 105\,(95--114) & 8.8\,(7.9--10.0) & 6.6\,(6.5--6.6) \\
$\gamma_{\text{ins}}/\gamma$ & 0.9 & 3.5\,(3.4--3.5) & & & 102\,(92--111) & 5.7\,(4.8--6.9) & 3.5\,(3.4--3.5) \\
$\gamma_{\text{ins}}/\gamma$ & 1 & 0.4\,(0.3--0.4) & & & 99\,(89--108) & 2.6\,(1.7--3.8) & 0.4\,(0.3--0.4) \\
\cmidrule(l){1-8}
$r$ (loading 19\%) & 0.0475 & 6.6\,(6.6--6.7) & & & 105\,(95--114) & 8.8\,(7.9--10.0) & 6.6\,(6.6--6.7) \\
$r$ (loading 20\%) & 0.06 & 6.6\,(6.5--6.6) & & & 105\,(95--114) & 8.8\,(7.9--10.0) & 6.6\,(6.5--6.6) \\
$r$ (loading 22\%) & 0.08 & 6.5\,(6.5--6.6) & & & 105\,(95--114) & 8.7\,(7.8--10.0) & 6.5\,(6.5--6.6) \\
$r$ (loading 24\%) & 0.1 & 6.5\,(6.4--6.5) & & & 105\,(95--114) & 8.7\,(7.8--9.9) & 6.5\,(6.4--6.5) \\
$r = 1 + \mu$ (loading 120\%) & 1.15 & 3.8\,(3.4--4.1) & & & 102\,(92--111) & 5.9\,(4.9--7.3) & 3.8\,(3.4--4.1) \\
\bottomrule
\end{tabular}
\par\smallskip{\raggedright\textit{Notes.} Components of Theorem~\ref{thm:insurance_value} relative to $\text{CVaR}_\beta[S(\tau^{\ast})]$, in \%. Within each setting, the top block varies $\mu$, $\rho$, $\gamma$ and $\gamma_{\text{ins}}$ by $\pm 25\%$, with $\gamma$ varied at fixed $\gamma_{\text{ins}}/\gamma = 0.8$. The middle block varies the capital relief through $\gamma_{\text{ins}}/\gamma$, and the bottom block varies the insurer's cost of tail capital $r$ up to the tail-priced premium $r = 1 + \mu$, with the implied loading over expected payouts. The calibration components are left blank where they do not depend on the varied parameter.\par}
\end{table}

\subsection{Renewal and Oversight}\label{sec:renewal_oversight_ec}

\paragraph{Renewal.} Table~\ref{tab:renewal_oversight} reports the optimal renewal time for renewal costs of 2\%, 5\% and 10\% of the baseline CVaR, with the excess risk charged at the firm's capital cost $\lambda = \rho\gamma_{\text{ins}}$, for both drift rates and both policies, and the optimal oversight level across the family of cost and benefit functions. In screening, renewal times grow roughly with the square root of the renewal cost, as Corollary~\ref{cor:sqrt} suggests where the excess risk grows nearly linearly. In the diagnostic setting with recalibration they grow faster, because the recalibrated excess risk flattens. Recalibration lengthens the renewal time in the diagnostic setting and leaves it within a quarter of a month of the no-action value in screening. \rev{The two settings are not comparable in the severity of drift, since after 36 months of fast drift the AUC of the screening model falls to 0.57, whereas the AUC of the diagnostic model cannot fall below about 0.83, the lower bound of the trigonometric family.} Under fast drift in the diagnostic setting, the excess risk plateaus and no finite renewal is optimal at a renewal cost of 10\% without recalibration and of 5\% or more with recalibration (the case $\mathcal{C}_a > H_\infty$ of Proposition~\ref{prop:optimal_duration_full}, a consequence of the lower bound on the trigonometric model's discrimination).

\paragraph{Oversight.} At the base specification, the optimal oversight levels yield net benefits of \$63{,}300 (screening) and \$27{,}200 (diagnostic) per year, 4.8\% and 7.9\% of the premium (Figure~\ref{fig:interpretability_premium}). In the oversight family, $\zeta^\ast$ increases with the cost exponent $a$ at the base specification (a steeper cost near full review makes moderate review relatively cheap) and is interior in every screening case and in all but one diagnostic case (the corner $\zeta^\ast = 1$ arises at the lowest cost scale with $a = 2$, $b = 1.5$ and $\xi = 0.75$, and three neighboring specifications reach it in 8\% of the draws). If the firm counts, in addition to the premium, the reduction of its retained losses and of its capital charge, the benefit base is 12 and 33 times the premium, and the closed form of Equation~\eqref{eq:zetastar} yields $\zeta^\ast = 1$ in every draw of both settings.

\paragraph{Computation.} Algorithm~\ref{alg:bisection} reaches the optimal threshold in 40 iterations for $\varepsilon = 10^{-12}$, in less than 0.01 seconds per setting on a standard personal computer, and only about fifty distinct tail sets occur along $\tau \in [0,1]$ in each setting (on average 53 in screening and 56 in the diagnostic setting over the 100 draws, and never more than 62), against the bound $J(J-1)+1 = 999{,}001$ of Lemma~\ref{lem:tailsets}.

\begin{table}[htbp]
\centering
\tiny
\setlength{\tabcolsep}{2pt}
\caption{Renewal and oversight.}
\label{tab:renewal_oversight}
\begin{tabular}{llrrr}
\toprule
\textbf{Setting} & \textbf{Drift, policy} & \textbf{$\mathcal{C}_a = 2\%$} & \textbf{5\%} & \textbf{10\%} \\ \midrule
screening & $\delta = 0.10$, no action & 15.8 (15.6--15.9) & 25.4 (25.1--25.7) & 36.8 (36.2--37.4) \\
screening & $\delta = 0.10$, recalibrate & 15.9 (15.7--16.0) & 25.6 (25.3--25.8) & 37.0 (36.6--37.5) \\
screening & $\delta = 0.60$, no action & 6.8 (6.6--6.9) & 11.5 (11.2--11.8) & 18.3 (17.5--19.0) \\
screening & $\delta = 0.60$, recalibrate & 6.8 (6.7--6.9) & 11.5 (11.3--11.8) & 18.2 (17.5--19.1) \\
diagnostic & $\delta = 0.10$, no action & 18.3 (18.3--18.3) & 32.0 (32.0--32.1) & 51.6 (51.5--51.7) \\
diagnostic & $\delta = 0.10$, recalibrate & 22.1 (21.8--22.4) & 46.5 (45.3--47.7) & 106.7 (100.5--115.1) \\
diagnostic & $\delta = 0.60$, no action & 9.9 (9.9--9.9) & 25.4 (25.4--25.5) & -- \\
diagnostic & $\delta = 0.60$, recalibrate & 25.0 (22.7--28.3) & -- & -- \\
\midrule
\multicolumn{5}{l}{\textit{Oversight: $\zeta^\ast$ by $(a, b)$ at the base scale and gain, with its range and corner share over the family}} \\ \midrule
\textbf{Setting} & \textbf{$a$} & \textbf{$b = 0.5$ / 1 / 1.5} & \textbf{Range of $\zeta^\ast$} & \textbf{Corners} \\ \midrule
screening & 2 & 0.21 (0.20--0.22) / 0.19 (0.18--0.20) / 0.08 (0.07--0.09) & 0.01--0.73 & 0\% \\
screening & 3 & 0.33 (0.32--0.34) / 0.36 (0.35--0.37) / 0.33 (0.32--0.34) & 0.13--0.69 & 0\% \\
screening & 4 & 0.42 (0.41--0.43) / 0.46 (0.45--0.47) / 0.46 (0.45--0.47) & 0.26--0.71 & 0\% \\
diagnostic & 2 & 0.29 (0.28--0.31) / 0.32 (0.29--0.34) / 0.22 (0.20--0.26) & 0.01--1.00 & 4\% \\
diagnostic & 3 & 0.41 (0.40--0.42) / 0.46 (0.44--0.47) / 0.46 (0.44--0.48) & 0.18--0.96 & 0\% \\
diagnostic & 4 & 0.48 (0.47--0.49) / 0.54 (0.53--0.55) / 0.56 (0.55--0.58) & 0.32--0.87 & 0\% \\
\bottomrule
\end{tabular}
\par\smallskip{\raggedright\textit{Notes.} The top block gives the optimal renewal time $T^{\ast}$ in months for renewal costs of 2\%, 5\% and 10\% of the baseline CVaR, where a dash denotes no finite optimum. The bottom block gives the optimal oversight level $\zeta^\ast$ by cost exponent $a$ and benefit exponent $b$ at the base specification, with its range and share of corner solutions over the full family.\par}
\end{table}

\begin{figure}[htbp]
    \centering
    \includegraphics[width=0.75\linewidth]{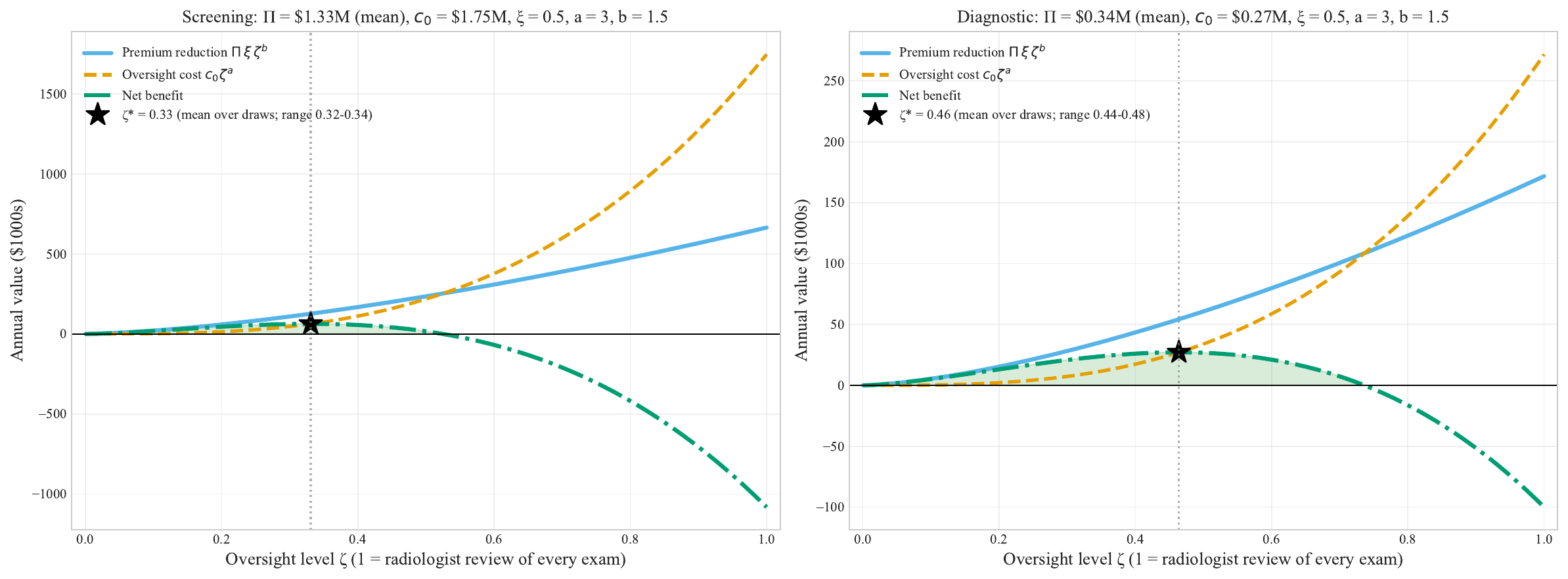}
    \caption{Human oversight and premium reduction. Premium reduction, oversight cost and net benefit as functions of the oversight level $\zeta$, with the optimum of Proposition~\ref{prop:optimal_zeta} and its 2.5--97.5\% range.}
    \label{fig:interpretability_premium}
\end{figure}
\end{revblock}

\FloatBarrier
\section{Supplementary Figures}\label{sec:suppfigures}

\FloatBarrier
\newpage
\clearpage
\section{Notation Summary}\label{tab:notationsummary}

\begin{table}[htbp]
\centering
\tiny
\caption{Notation.}
    \label{tab:notation}
\begin{tabular}{@{}p{0.2\textwidth}p{0.75\textwidth}@{}}
\toprule
\textbf{Symbol} & \textbf{Description} \\
\toprule
\multicolumn{2}{@{}l}{\textit{Binary Classification Model Parameters}} \\ \midrule
$\alpha$ & Proportion of class 0 instances in the population (class imbalance) \\
$k$ & Classifier discrimination parameter, governing model quality\\
$\overline{\gamma}$ & \rev{Rank score of an instance (model score transformed to a uniform marginal)} \\
$\tau$ & Classification threshold used to separate class 0 vs. class 1 \\
$\tau^{\ast}$ & Risk-aware threshold that minimizes tail risk (CVaR-optimal) \\
$\tau_{\mathrm{acc}}$ & Accuracy-maximizing threshold \\
$\tau_{\text{EL}}^{\rev{\ast}}$ & Expected-loss minimizing threshold \\
$f(\tau)$ & Auxiliary function used in trigonometric model \\
$\mathbb{P}(Y=1 \mid \overline{\gamma})$ & Conditional probability of class 1 given predicted score $\overline{\gamma}$ \\ \midrule
\multicolumn{2}{@{}l}{\textit{Error Probabilities}} \\ \midrule
$\mathbb{P}_{\text{FN}}(\tau)$ & False negative rate at threshold $\tau$ \\
$\mathbb{P}_{\text{FP}}(\tau)$ & False positive rate at threshold $\tau$ \\
$\text{Acc}(\tau)$ & Accuracy of classifier at threshold $\tau$ \\\midrule
\multicolumn{2}{@{}l}{\textit{Scenario-Based Loss Model}} \\ \midrule
$N$ & Number of instances (e.g., patients, transactions) \\
$J$ & Number of risk scenarios (representing contextual heterogeneity) \\
$K_{ji}$ & Cost incurred by false positive on instance $i$ under scenario $j$ \\
$L_{ji}$ & Cost incurred by false negative on instance $i$ under scenario $j$ \\
$S_j(\tau)$ & Total economic loss in scenario $j$ at threshold $\tau$ \\
$\mathcal{J}_\beta$ & \rev{Tail set, the $m = (1-\beta)J$ scenarios with the largest losses} \\ \midrule
\multicolumn{2}{@{}l}{\textit{Risk Measures}} \\ \midrule
$\text{CVaR}_\beta(\cdot)$ & Conditional Value-at-Risk at confidence level $\beta$ \\
$\text{VaR}_\beta(\cdot)$ & Value-at-Risk at confidence level $\beta$ \\
$\Delta_{\text{CVaR}}$ & Relative increase in tail risk (CVaR penalty) due to suboptimal threshold \\
$\Delta_\tau$ & Efficiency gap; deviation between optimal and benchmark thresholds \\ \midrule
\multicolumn{2}{@{}l}{\textit{Insurance Contract Parameters}} \\ \midrule
$L_{\text{occ}}$ & Per-occurrence loss limit in insurance contract \\
$L_{\text{agg}}$ & Aggregate annual loss limit in insurance contract \\
$C_{\text{tot}}(\tau, L_{\text{occ}}, L_{\text{agg}})$ & Total insurer payment under threshold $\tau$ and contract limits \\
$\Pi$ & \rev{Insurance premium (cost-of-capital method, Equation~\eqref{eq:premium})} \\
$\mu$ & Premium loading factor for insurer profit and admin cost \\
$\gamma, \gamma_{\text{ins}}$ & Capital multipliers for uninsured vs. insured exposures \\
$\rho$ & \rev{Firm's cost of capital} \\
$D$ & \rev{Number of decisions in the policy period} \\
$V_\theta$ & Total insurance value for firm type $\theta$ \\
$V_{\text{base}}$ & Value from risk transfer and capital relief \\
$V_{\text{calib}}^{\theta}$ & Value from calibration improvement (for suboptimally configured firms) \\
\rev{$\mathcal{Q}_\beta$, $q^\ast$} & \rev{Risk envelope of CVaR; worst-case scenario weights at $\tau^{\ast}$} \\
$p^\ast$ & Tail-weighted share of false positive costs, $K_{q^\ast}/(K_{q^\ast}+L_{q^\ast})$ \\
$r$ & Insurer's cost of tail capital in the premium \\
$s_{\mathrm{op}}$, $S_L$ & Operational (uninsured) and liability (insured) components of the loss \\
$e(t)$, $H(T)$ & Excess-risk cost rate under drift; renewal condition function \\
\rev{$\psi(\zeta)$} & Marginal oversight cost per unit of marginal risk reduction, $c'/g'$ \\
\rev{$\chi$} & \rev{Tail-to-average ratio of the insured loss, $\text{CVaR}_\beta[S_L]/\mathbb{E}[S_L]$} \\
\rev{$\phi$, $\phi_{\mathrm{FP}}$, $\phi_{\mathrm{FN}}$} & \rev{Claim-emergence rate; probabilities that a false positive or a missed case leads to a paid claim} \\
\rev{$p_{\min}$, $p_{\max}$} & \rev{Lower and upper bounds on the derivative of the calibration map} \\ \midrule
\multicolumn{2}{@{}l}{\textit{Extensions and Other Parameters}} \\ \midrule
$k(t)$ & Time-varying model quality under drift \\
$\delta$ & Drift rate (rate of performance decay) \\
$\mathcal{C}_a$ & Administrative renewal cost for contracts \\
\rev{$T$, $T^{\ast}$} & \rev{Renewal time and its optimum} \\
\rev{$\mathcal{A}(T)$, $\lambda$} & \rev{Long-run average cost of renewing every $T$ periods; capital charge on retained risk, $\rho\gamma_{\text{ins}}$} \\
$\zeta$ & Interpretability level of AI system (0 = black box, 1 = full transparency) \\
$c(\zeta)$ & Cost function for achieving interpretability level $\zeta$\rev{, with scale $c_0 = c(1)$ and exponents $a$ (cost) and $b$ (benefit) in the power family} \\
$\xi$ & Human-AI error reduction potential (relative performance gap) \\
$g(\zeta)$ & Fraction of errors humans can catch as a function of interpretability \\ \bottomrule
\end{tabular}
\end{table}

\FloatBarrier
\bibliographystyleec{informs2014}
\bibliographyec{ref}

\end{document}